# Information-Theoretic Methods for Planning and Learning in Partially Observable Markov Decision Processes

Thesis submitted for the degree of

"Doctor of Philosophy"

by

Roy Fox

# Acknowledgements

First and foremost, I would like to express my deep gratitude to my advisor, Prof. Naftali Tishby, for instilling in me a passion for science, for sharing with me his tremendous insight, for empowering me to ask the important questions, and for supporting me in my first steps as an independent researcher.

During my exchange program at Columbia University, I was hosted by Prof. Larry Abbott and Prof. Liam Paninski, and I greatly appreciate their kind help and generous support.

Many thanks to my collaborators, Ari Pakman, Michal Moshkovitz, Josh Merel, Prof. Liam Paninski and Prof. Tony Jebara, for our pleasant, professional and productive discussions.

I am privileged to have had so many fascinating discussions and exchanges of truly exciting ideas with my cherished friends and colleagues, Prof. Daniel Polani, Noga Zaslavsky, Pedro Ortega, David Pfau, Ari Pakman, Stas Tiomkin, Nori Jacoby, Michal Moshkovitz, Nadav Amir and Hadar Levi.

I would like to thank my advisory committee members, Prof. Amir Globerson and Prof. Shie Mannor, for their advice and guidance.

I am forever grateful to my beloved parents and brother, Miri, Amit and Ken, for their continuing support in all my endeavors.

Lastly, my very special thanks belong to my dear partner, Noga Zaslavsky, whose ideas, insight, advice, collaboration and support have been making this journey possible and worthwhile.

# Abstract

We model the interaction of an intelligent agent with its environment as a Partially Observable Markov Decision Process (POMDP), where the joint dynamics of the internal state of the agent and the external state of the world are subject to extrinsic and intrinsic constraints. Extrinsic constraints of partial observability and partial controllability specify how the agent's input observation depends on the world state and how the latter depends on the agent's output action. The agent also incurs an extrinsic cost, based on the world states reached and the actions taken in them.

Bounded agents are also limited by intrinsic constraints on their ability to process information that is available in their sensors and memory and choose actions and memory updates. In this dissertation, we model these constraints as information-rate constraints on communication channels connecting these various internal components of the agent.

The simplest is to first consider reactive (memoryless) agents, with a channel connecting their sensors to their actuators. The problem of optimizing such an agent, under a constraint on the information rate between the input and the output, is a sequential rate-distortion problem. The marginal distribution of the observation can be computed by a forward inference process, whereas the expected cost-to-go of an action can be computed by a backward control process. Given this source distribution and this effective distortion, respectively, each step can be optimized by solving a rate-distortion problem that trades off the extrinsic cost with the intrinsic information rate.

Retentive (memory-utilizing) agents can be reduced to reactive agents by interpreting the state of the memory component as part of the external world state. The memory reader can then be thought of as another sensor and the memory writer as another actuator and they are limited by the same informational constraint between inputs and outputs.

In this dissertation we make four major contributions detailed below and many smaller contributions detailed in each section.

First, we formulate the problem of optimizing the agent under both extrinsic and intrinsic constraints and develop the main tools for solving it. This optimization problem is highly non-convex, with many local optima. Its difficulty is mostly due to the coupling of the forward inference process and the backward control process. The inference policy and the control policy can be optimal given each other but still jointly suboptimal as a pair. For example, if some information is not attended to it cannot be used and if it is not used it should optimally not be attended to.

Second, we identify another reason for the challenging convergence properties of the optimization algorithm, which is the bifurcation structure of the update operator near phase transitions. We show that the update operator may undergo period doubling, after which the optimal policy is periodic and the optimal stationary policy is unstable. Any algorithm for planning in such domains must therefore allow for periodic policies, which may themselves be subject to an informational constraint on the clock signal.

Third, we study the special case of linear-Gaussian dynamics and quadratic cost (LQG), where the optimal solution has a particularly simple and solvable form. Under informational constraints, the forward and the backward processes are not separable. However, we show that they do have a more explicitly solvable structure; namely, a sequential semidefinite program. This also allows us to analyze the structure of the retentive solution under the reduction to the reactive setting.

Fourth, we explore the learning task, where the model of the world dy-namics is unknown and sample-based updates are used instead. We focus on fully observable domains and measure the informational cost with the KL divergence, so that the problem can be solved with a backward-only algorithm. We suggest a schedule for the tradeoff coefficient, such that more emphasis is put on reducing the extrinsic cost and less on the simplicity of the solution, as uncertainty is gradually removed from the value function through learning. This leads to improved performance over existing reinforcement learning algorithms.

# Contents

# Chapter 1

# Introduction

In this chapter we introduce the conceptual framework that is the basis for the results presented in this thesis. Although reinforcement learning and information theory have both been studied intensively for many decades, some aspects of our approach to these fields are novel. The preliminaries are included here in somewhat non-standard notation and are accompanied by several new organizing principles and insights.

## 1.1 Partially Observable Markov Decision Processes

A Partially Observable Markov Decision Process (POMDP) is a dynamical system, with outputs that partially reveal the state of the system, and inputs that partially control the state dynamics. This richly expressive model has numerous and diverse applications, from autonomous vehicles to ad displays [1]. POMDPs, and particularly the reinforcement learning paradigm for optimizing and learning them, have therefore enjoyed increasing attention from the research community in recent years.

### 1.1.1 Setting

**Dynamics**

A discrete-time dynamical system has a time-dependent state, and possibly stochastic dynamics that determine the distribution of each next state given each current state. A closed system has no input, and the dynamics are simply a Markov chain of states $\{s_t\}$, induced by the conditional probability distribution $p(s_{t+1}|s_t)$ of the state transition. In open systems, which we consider in this thesis, an input control signal $a_t \in \mathcal{A}$, also called an action, can affect the dynamics of the state $w_t \in \mathcal{W}$, which are now given by the distribution $p(w_{t+1}|w_t, a_t)$.

The system also emits an output signal $o_t \in \mathcal{O}$, also called an observation, based on its state. The observation dynamics are given by the distribution $\sigma(o_t|w_t)$. In the special case of a fully-observable Markov Decision Process, the observation space contains the state space, and $\sigma(o_t|w_t) = \delta_{o_t = w_t}$.

The control signal is generated by an agent, based on past observations, according to some policy. A history-based policy is given by the distribution $\pi(a_t|o_{\leqslant t})$, where $o_{\leqslant t}$ denotes the observable history; i.e., the sequence of observations up to time $t$. Jointly with its environment, also called the world, the agent forms a larger dynamical system, which induces a stochastic process over $\{w_t, o_t, a_t\}$.

Generic history-based policies are hard to optimize, implement, and even represent, since the space of observable histories grows exponentially in size with the length of the history. Instead, the agent is equipped with some memory $m_t \in \mathcal{M}$, which summarizes the observable history, and on which the future actions are based. In addition to the control policy $\pi(a_t|m_t)$, the agent now consists of an inference policy $q(m_{t+1}|m_t, o_{t+1})$, for updating the memory state using the new observation. This induces a stochastic process over $\{w_t, o_t, m_t, a_t\}$.

**Extrinsic Constraints**

The world dynamics can be considered extrinsic limitations on how the agent can interact with the world state. Without these limitations, the agent could precisely observe the current state $w_t$ of the world, and completely determine its next state $w_{t+1}$. In POMDPs, the observability is partial, in that the only information about $w_t$ that the agent can use is that given by $o_{\leqslant t}$. Dually, controllability is also partial, in that the only future trajectory of states that the agent can effect are those induced by $a_{\geqslant t}$. Put another way, the state dynamics $p$ limits how the agent can control the world

$$\mathbb{P}_\pi(w_{t+1}|w_t, m_t) = \mathbb{E}_{a_t \sim \pi(\cdot|m_t)}[p(w_{t+1}|w_t, a_t)],$$

so that the distribution of $w_{t+1}$, given $w_t$ and $m_t$, is a selected mixture $\pi$ of the fixed distributions $p$. Similarly, the observation dynamics $\sigma$ limits how the memory state can adapt to the new world state

$$\mathbb{P}_q(m_{t+1}|m_t, w_{t+1}) = \mathbb{E}_{o_{t+1} \sim \sigma(\cdot|w_{t+1})}[q(m_{t+1}|m_t, o_{t+1})],$$

so that the distribution of $m_{t+1}$, given $m_t$ and $w_{t+1}$, is a fixed mixture $\sigma$ of the selected distributions $q$.

Another limitation on the policy of the agent, which is often viewed as the target of the policy optimization, is to achieve low values of the expectation of some cost (or equivalently, high values of an expected reward). Without loss of generality, the cost is taken to be a function $c(w_t, a_t)$ of the world state and the action. On a long timescale, expected cost accumulates at a linear rate, and we are concerned with that asymptotic rate

$$V_{\pi,q} = \limsup_{T \to \infty} \frac{1}{T} \sum_{t=0}^{T-1} \mathbb{E}[c(w_t, a_t)]. \tag{1}$$

**Stationary Processes**

Let

$$\mathbb{P}_{\pi,q}(w_{t+1}, m_{t+1}|w_t, m_t) = \mathbb{P}_{\pi}(w_{t+1}|w_t, m_t)\, \mathbb{P}_q(m_{t+1}|m_t, w_{t+1}).$$

For a marginal distribution $\bar{p}(s_t)$ over the joint state $s_t = (w_t, m_t)$ of the world and the agent, the forward operator

$$P_{\pi,q} : \bar{p} \mapsto \mathbb{E}_{s_t \sim \bar{p}}[\mathbb{P}_{\pi,q}(\cdot|s_t)],$$

induces a Markov process on the joint state. The limit (1) is clearly related to fixed points of $P_{\pi,q}$, called stationary distributions of the process: if at any point the process reaches a stationary distribution $\bar{p}$, it remains in that distribution, and

$$V_{\pi,q} = \mathbb{E}_{\substack{(w_t,m_t)\sim\bar{p} \\ a_t\sim\pi(\cdot|m_t)}}[c(w_t, a_t)]. \qquad (2)$$

However, the process does not always have a stationary distribution, and when it does it may not be unique, with the one actually reached depending on the initial distribution of $s_0 = (w_0, m_0)$.

To formulate the conditions under which there exists a unique stationary distribution, we require some results from ergodicity theory, an extensively researched field which we address here only in a nutshell.

We say that the joint state $s$ communicates with $s'$, and denote $s \to_{\pi,q} s'$, if $s'$ is reached from $s$ with positive probability after some finite time $t(s, s')$

$$P_{\pi,q}^t[\delta_s](s') = \mathbb{P}_{\pi,q}(s_t = s'|s_0 = s) > 0.$$

Consider the equivalence classes of the equivalence relation $s \leftrightarrow_{\pi,q} s'$ (i.e. $s$ and $s'$ communicate with each other), and the partial order $\to_{\pi,q}$ induced on the set of these communicating classes.

We say that a communicating class is closed if the probability of leaving it is 0. A closed communicating class has a unique stationary distribution over its member states. Even if the process has period $T$, and the marginal joint distributions follow a limit cycle $\bar{p}_0, \ldots, \bar{p}_{T-1}$, the limit average (1) is the expected cost (2), with respect to the stationary distribution $\bar{p} = \frac{1}{T} \sum_{t=0}^{T-1} \bar{p}_t$.

If the process has multiple closed communicating classes, the convex combination of their stationary distributions is also stationary, and thus the value of the policy depends on the initial state distribution. On the other hand, the process may have no closed communicating classes, or more generally the total probability of reaching any closed communicating class from the initial state may be less than 1. If a closed communicating class is not reached, the process goes through an infinite sequence of distinct communicating classes, which of course excludes stationarity. This is only possible when the state space is infinite, in the case where $\mathcal{W}$ or $\mathcal{M}$ is infinite.

In reinforcement learning, it is common to assume ergodic processes, consisting in particular of a single closed communicating class. However, it should be recalled that the process is induced jointly by the world dynamics and the agent policy, and some policies are not ergodic. Some of the complications this creates are explored in Section 2.2; however, further implications are beyond the scope of this dissertation. Here we restrict the discussion to policies that are well-behaved, in that the process they induce reaches a single closed communicating class with probability 1.

#### Finite-Horizon Processes

In many reinforcement learning domains the process terminates upon reaching a terminating state. A terminating state can be modeled as persisting with probability 1 and cost 0, making it a closed communicating class, and the value of any well-behaved policy 0. For the comparison of policies in this episodic setting to be meaningful, we consider the total life-long expected

cost

$$V_{\pi,q} = \sum_{t=0}^{\infty} \mathbb{E}[c(w_t, a_t)],$$

rather than the average cost.

A special case of this finite-horizon setting is the popular discounted setting, although it is often mistakenly considered to have an infinite horizon [2]. In this setting, each transition has a fixed probability $0 < 1 - \gamma \leq 1$ of terminating, regardless of the current state or action. The horizon $T_f$ is distributed geometrically with parameter $1 - \gamma$, and we have

$$\begin{aligned}
V_{\pi,q} &= \sum_{T=1}^{\infty} \mathbb{P}(T_f = T) \sum_{t=0}^{T_f - 1} \mathbb{E}[c(w_t, a_t)] \\
&= \sum_{T=1}^{\infty} (1 - \gamma)\gamma^{T-1} \sum_{t=0}^{T-1} \mathbb{E}[c(w_t, a_t)] \\
&= \sum_{t=0}^{\infty} \mathbb{E}[c(w_t, a_t)|T_f > t] \sum_{T=t+1}^{\infty} (1 - \gamma)\gamma^{T-1} \\
&= \sum_{t=0}^{\infty} \gamma^t \, \mathbb{E}[c(w_t, a_t)|T_f > t].
\end{aligned}$$

It may appear in this expression that the horizon is infinite and the costs are discounted exponentially by $\gamma$. However the contribution of later time steps to the total cost becomes negligible on the effective horizon, which is in the order of the expected termination time $\frac{1}{1-\gamma}$.

Another special case is the fixed finite horizon $T$

$$V_{\pi,q} = \sum_{t=0}^{T-1} \mathbb{E}[c(w_t, a_t)].$$

This can be modeled by keeping track of the time index as a part of the state $w'_t = (t, w_t)$, and terminating when $t = T$. Policies in this setting are often

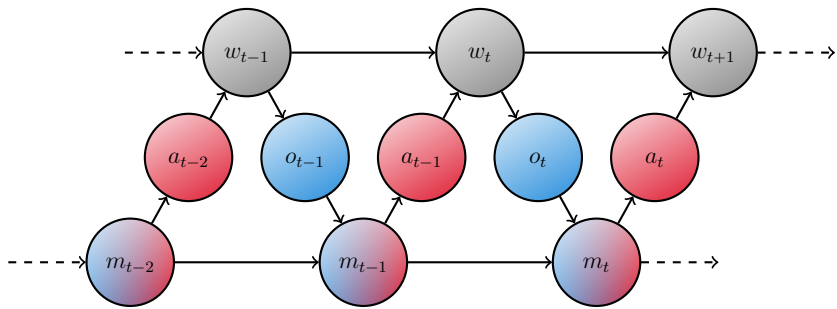

Figure 1.1: Bayesian network of a POMDP

time-dependent, as can be the dynamics or the cost.

Throughout this thesis, we will use many of these different horizon settings. In Section 2.1 we consider a finite horizon, in Section 2.2 a periodic infinite horizon, in Chapter 3 an infinite horizon, and in Section 4.1 a discounted horizon.

We have thus seen two common limitations on the agent policy; namely, the extrinsic limitations of partial observability and partial controllability, and the desire to incur low costs. Another set of possible limitations results from the scarcity of intrinsic resources required by the agent for computation and communication. Such conditions of bounded rationality or intrinsic motivation are the central theme of this dissertation, and will be introduced in Section 1.2.2.

### 1.1.2 Structure

**Symmetries**

The structure of the process defined in the previous section can be summarized in the Bayesian network in Figure 1.1. This structure has several symmetries that are insightful to explore, and which yield useful dualities [3].

First, there is a horizontal (up/down) symmetry between the agent and the world that maps the inputs and outputs of one component to those of the other; that is, actions are mapped to observations and vice versa. This symmetry underlines the fundamental distinction between the agent and the world: the dynamics $p$ and $\sigma$ of one remain fixed, as those of the other, $q$ and $\pi$, are optimized. This suggests that in a closed system, the component whose dynamics are adaptive on shorter timescales can be thought of as an agent with respect to the rest of the system.

Second, there is a vertical (left/right) symmetry between the past and the future, that again maps the inputs to the outputs and vice versa. This symmetry underlines the role of causality in the process: while the outputs of $w_t$, namely $o_t$ and $w_{t+1}$, are independent given the state $s_t = (w_t, m_t)$

$$o_t \perp w_{t+1} \mid w_t, m_t,$$

the inputs of $w_{t+1}$, namely $w_t$ and $a_t$, may be dependent given $s'_t = (m_t, w_{t+1})$

$$a_t \not\perp w_t \mid w_{t+1}, m_t.$$

To illustrate this, consider a light switch $w$ that is either on or off. A noisy observation $o$ of the state of the switch only affects the future through its perception by an agent $m$. On the other hand, given the next switch state $w'$, a change $w \neq w'$ is much more likely if the agent touches the switch than if it does not, so the actual action $a$ carries information on $w$ even beyond the mere intention $m$. Other similar causal asymmetries exist as well.

Nevertheless, this imperfect symmetry gives rise to an important duality between inference and control [4]. Inference is inherently a forward process, that is performed by computing the forward dynamics of the process while marginalizing and conditioning probabilities. Control is inherently a backward process, where earlier and earlier actions are selected based on their previously-computed future consequences. The two processes and the interchange between them are central in reinforcement learning, and the duality between them is thus highly insightful.

A third symmetry in the infinite-horizon setting is time shifting, which is the basis for the stationary analysis of this setting.

**Reactive Agents**

An interesting restriction of the solution space is to only consider reactive (memoryless) agents. That is, when observing $o_t$ and deciding on an action $a_t$, we disallow any access to previously inferred statistics of the observable history, and restrict the policy to be of the form $\pi(a_t|o_t)$.

Requiring the agent policy to be reactive may have a considerable impact on its optimal value, since the optimal reactive policy can be arbitrarily worse than the optimal retentive (memory-utilizing) policy. Nevertheless, there are many reasons to consider reactive policies. First, in many real cases there are optimal or near-optimal reactive policies. Fully-observable MDPs are one important class of such cases, but others exist as well.

Second, even when reactive policies are not near-optimal, they may be preferred for the simplicity of optimizing and implementing them. For example, reactive policies were recently successfully employed to play Atari games, with only the 4 most recent screen frames available as observation in each step [5]. Third, a good treatment of reactive policies can be a springboard for general policies.

Last, but not least, in a certain sense, no generality is lost by restricting attention to reactive policies, because the general case of planning with retentive policies can be reduced to the problem of planning with reactive policies in an extended POMDP.

The extended POMDP is defined over the joint world-agent state space $\tilde{\mathcal{W}} = \mathcal{M} \times \mathcal{W}$. The observation space is $\tilde{\mathcal{O}} = \mathcal{M} \times \mathcal{O}$, and similarly the action space is $\tilde{\mathcal{A}} = \mathcal{M} \times \mathcal{A}$. The extended state dynamics are

$$\tilde{p}((m_t, w_{t+1})|(m_{t-1}, w_t), (m'_t, a_t)) = \delta_{m_t = m'_t} p(w_{t+1}|w_t, a_t),$$

the observation dynamics are

$$\tilde{\sigma}((m'_{t-1}, o_t)|(m_{t-1}, w_t)) = \delta_{m'_{t-1}=m_{t-1}} \sigma(o_t|w_t),$$

and the cost is

$$\tilde{c}((m_{t-1}, w_t), (m_t, a_t)) = c(w_t, a_t).$$

That is, the memory component of the world state is fully observable, fully controllable, and does not affect the cost.

To complete the reduction, we need to translate the solution reactive policy in the extended POMDP, $\tilde{\pi}((m_t, a_t)|(m_{t-1}, o_t))$, back into a retentive policy in the original POMDP. This policy does not generally have the property that $a_t$ is independent of $(m_{t-1}, o_t)$ given $m_t$, as it should according to our original notation, but this condition can usually be relaxed without practical implications (see Section 3.2). Alternatively, the retentive policy can have memory space $\mathcal{M} \times \mathcal{A}$, and

$$q((m_t, a_t)|(m_{t-1}, a_{t-1}), o_t) = \tilde{\pi}((m_t, a_t)|(m_{t-1}, o_t))$$
$$\pi(a'_t|(m_t, a_t)) = \delta_{a'_t=a_t}.$$

### 1.1.3 Methods

**Optimization Problem**

We are interested in optimizing the value of the policy $(\pi, q)$

$$V_{\pi,q} = \limsup_{T \to \infty} \frac{1}{T} \sum_{t=0}^{T-1} \mathbb{E}[c(w_t, a_t)] = \mathbb{E}_{\substack{(w_t, m_t) \sim \bar{p} \\ a_t \sim \pi(\cdot|m_t)}}[c(w_t, a_t)]$$

under a constraint on the observability and controllability allowed by the world dynamics $(p, \sigma)$, so that $\bar{p}$ is a fixed point of the forward recursion

$$\bar{p}(w_{t+1}, m_{t+1}) = \mathbb{E}_{\substack{(w_t, m_t) \sim \bar{p} \\ a_t \sim \pi(\cdot|m_t) \\ o_{t+1} \sim \sigma(\cdot|w_{t+1})}} [p(w_{t+1}|w_t, a_t) q(m_{t+1}|m_t, o_{t+1})]. \quad (3)$$

We can formulate the optimization target as the Lagrangian

$$\begin{aligned}
\mathcal{L}_{\bar{p}, \pi, q, \nu} &= \mathbb{E}_{\substack{(w_t, m_t) \sim \bar{p} \\ a_t \sim \pi(\cdot|m_t)}} [c(w_t, a_t)] \\
&\quad + \nu \cdot \left( \mathbb{E}_{\substack{(w_t, m_t) \sim \bar{p} \\ a_t \sim \pi(\cdot|m_t) \\ o_{t+1} \sim \sigma(\cdot|w_{t+1})}} [p(\cdot|w_t, a_t) \otimes q(\cdot|m_t, o_{t+1})] - \bar{p} \right) \\
&= \mathbb{E}_{(w_t, m_t) \sim \bar{p}} \Bigg[ \mathbb{E}_{a_t \sim \pi(\cdot|m_t)} \bigg[ c(w_t, a_t) \\
&\quad + \mathbb{E}_{\substack{w_{t+1} \sim p(\cdot|w_t, a_t) \\ o_{t+1} \sim \sigma(\cdot|w_{t+1}) \\ m_{t+1} \sim q(\cdot|m_t, o_{t+1})}} [\nu(w_{t+1}, m_{t+1})] \bigg] - \nu(w_t, m_t) \Bigg],
\end{aligned}$$

where $\nu(w_t, m_t)$ is the Lagrange multiplier that corresponds to the constraint of the forward recursion (3). In the infinite-horizon setting, we also add the constraint that $\bar{p}$ is a normalized probability distribution, $\mathbb{E}_{\bar{p}}[1] = 1$, with multiplier $\lambda$.

Although this optimization problem is highly non-convex, and many local optima exist, we chose a parameterization under which the Lagrangian is linear separately in each parameter. This enables us to easily find the gradient with respect to each parameter, and completely optimize over it with the other parameters fixed.

**Backward Operator**

The gradient with respect to $\bar{p}$ is

$$\partial_{\bar{p}(w_t, m_t)} \mathcal{L}_{\bar{p}, \pi, q, \nu} = \mathbb{E}[c(w_t, a_t) + \nu(w_{t+1}, m_{t+1}) | w_t, m_t] - \nu(w_t, m_t) - \lambda.$$

A necessary condition for the solution to be optimal is that the gradient be 0. By taking an expectation on both sides with respect to a stationary distribution $\bar{p}(w_t, m_t)$, $\lambda$ must be the target expected cost $\mathbb{E}[c(w_t, a_t)]$, and

$$\nu(w_t, m_t) = \mathbb{E}_{\substack{a_t \sim \pi(\cdot|m_t) \\ w_{t+1} \sim p(\cdot|w_t, a_t) \\ o_{t+1} \sim \sigma(\cdot|w_{t+1}) \\ m_{t+1} \sim q(\cdot|m_t, o_{t+1})}} \left[ c(w_t, a_t) + \nu(w_{t+1}, m_{t+1}) \right] - \lambda. \quad (4)$$

As often is the case with Lagrange multiplier, $\lambda$ and $\nu$ can be interpreted as quantities of interest in the problem. Intriguingly, the Karush-Kuhn-Tucker optimality conditions require $\lambda$ to be our target expected cost.

As for $\nu$, (4) is a backward recursion for the value of being in joint state $(w_t, m_t)$. It computes the cost-to-go for trajectories starting at that state, by accumulating backwards the cost, in a dynamic-programming scheme. Although it was introduced as a Lagrange multiplier for the forward recursion, $\nu$ turns out to be the joint-state value function, also called the cost-to-go.

The duality between inference and control that was previously mentioned in Section 1.1.2 manifests here in the primal-dual sense of optimization theory: the backward recursion (4) determines the Lagrange multiplier of the forward constraint (3). Since values and costs often appear as dual to log-probabilities [4], it is fitting that the mean cost $\lambda$ is subtracted as an equalizer in (4), somewhat resembling in form a normalization of probabilities.

In the finite-horizon setting, we exclude the terminating state from our notation, and the stationary distribution is guaranteed by the dynamics (3) to be 0 in non-terminating states. $\bar{p}$ is no longer normalized, and the Lagrange multiplier $\lambda$ is omitted to get the ordinary Bellman equation [2].

Note that, throughout this discussion, we assumed that the constraint $\bar{p} \geq 0$ is inactive, otherwise it must also be included in the Lagrangian.

**Optimal Policy**

As for the agent policy, we have that optimally $\pi$ deterministically selects the action that optimizes the future value

$$a_t^*(m_t) = \arg\min_{a_t} \mathbb{E}[c(w_t, a_t) + \nu(w_{t+1}, m_{t+1})|m_t] \qquad (5)$$

The expectation is with respect to the posterior distribution given the agent's memory state

$$\mathbb{P}_{\bar{p}}(w_t|m_t) = \frac{\bar{p}(w_t, m_t)}{\mathbb{E}_{(w_t', m_t') \sim \bar{p}}[\delta_{m_t' = m_t}]}.$$

This is the objective belief that the agent should have about what the state of the world $w_t$ may be, when the agent itself is in state $m_t$. To be able to perform this computation, the agent must be able to interpret $m_t$ as representing this belief. An inference policy that induces a subjective belief $b_{m_t}(w_t)$ which equals the objective belief $\mathbb{P}_{\bar{p}}(w_t|m_t)$, is called *objectively consistent*.

Similarly, the optimal $q$ deterministically selects the next memory state that optimizes the future value starting at the inference half-step

$$m_{t+1}^*(m_t, o_{t+1}) = \arg\min_{m_{t+1}} \mathbb{E}[\nu(w_{t+1}, m_{t+1})|m_t, o_{t+1}]. \qquad (6)$$

Here the expectation is with respect to the updated posterior

$$\mathbb{P}_{\bar{p},\pi}(w_{t+1}|m_t, o_{t+1}) = \frac{\bar{p}'(m_t, w_{t+1})\sigma(o_{t+1}|w_{t+1})}{\mathbb{E}_{(m_t', w_{t+1}') \sim \bar{p}'}[\delta_{m_t' = m_t}\sigma(o_{t+1}|w_{t+1}')]},$$

where $\bar{p}'$ is the half-step phased stationary distribution

$$\bar{p}'(m_t, w_{t+1}) = \mathbb{E}_{\substack{(w_t', m_t') \sim \bar{p} \\ a_t' \sim \pi(\cdot|m_t')}}[\delta_{m_t' = m_t} p(w_{t+1}|w_t', a_t')].$$

If the memory state space is unlimited, inference is optimized by choosing a state $m_{t+1}$ that represents this updated posterior; i.e., having

$$b_{m_{t+1}}(w_{t+1}) = \mathbb{P}_{\bar{p},\pi}(w_{t+1}|m_t, o_{t+1}).$$

This is the Bayesian inference policy that is the standard in reinforcement learning [6]. It is deterministic and objectively consistent.

In summary, an analysis of the optimization problem gives us a forward recursion (3) on the marginal distributions, a backward recursion (4) on the cost-to-go function, and policy optimization equations (5), (6). These can be treated as update equations that allow the iterative update of each solution parameter given the others. Since each update brings a parameter to its optimum, the solution is monotonically improved in each iteration, and is guaranteed to converge in value, at least to a local optimum. This type of forward-backward algorithm is a central element in reinforcement learning and in this thesis.

### Model-Based vs. Sample-Based Learning

Our approach so far has been model-based, in that the world dynamics $(p, \sigma)$ are needed to compute the forward (3) and backward (4) updates, as well as the inference policy (6). Indeed, general POMDPs are usually solved using model-based methods [7] [8] [9].

When the model is known at the time of agent design, the task of optimizing its policy is called planning. When the model is unknown, the task is called learning, since the agent must learn something about the world before good behavior can be identified. The agent cannot simply exploit its partial knowledge of the world dynamics before it sufficiently explores unknown aspects of the world, because the unknown aspects could allow it to choose a much better policy. The agent must therefore trade off exploration and exploitation [2].

However, the distinction is blurred in model-based methods, because at least in principle, the unknown dynamical parameters of the world can be considered part of the unknown state of the world. Exploration in this sense is not unlike active perception [10], where the actions are selected also for their benefit in better observing the world state.

Learning and exploration are more pertinent to sample-based methods, where no model of the world is known or learned, and only the value function or the policy are learned. Unable to perform computations that involve the unknown dynamics explicitly, sample-based methods interact with an environment implementing these dynamics, and use the gathered samples to approximate the computations. This is usually done in the fully observable setting, where no inference or forward computation is needed. A prominent example is the approximation of the expectation in the backward recursion (4) with various sampling techniques [2]. This approach is employed in Section 4.1.

**Value Iteration vs. Gradient Methods**

The method presented above utilizes a specific parameterization of the problem that makes the target linear in each parameter separately. This enables the global optimization of each parameter given the others, which is then iterated until convergence, an approach called value iteration [2] [11].

Gradient methods update the parameters in a different manner [12] [13]. Whereas value iteration methods follow each coordinate of the gradient to convergence before moving on to another coordinate, gradient methods only take a small step in the direction of the gradient in each iteration. This makes these methods suitable for combining with the plethora of gradient-based parametric function learning methods developed in recent years in the optimization literature. Compared to value iteration methods, gradient methods follow a different trajectory in solution space, with implications for convergence that are a subject of ongoing research.

It is also possible to mix and match the approaches. For example, policy gradient methods often take small gradient steps with respect to the policy, but keep the forward-backward equations consistent.

### 1.1.4 Challenges

The POMDP planning problem is provably computationally hard [14], and the optimization problem is highly non-convex. Nevertheless, the problem is important enough to merit the attention it has been getting. Inspired by the fact that natural agents do regularly solve instances of the problem, the research community has come up with useful insights and increasingly effective approaches for solving it, but has also faced significant challenges.

**Memory Space Identification**

Central to these challenges is the identification of a good space of memory states. This is essentially a representation learning problem. Naively, since memory states represent belief states, we may be tempted to consider the entire space of distributions over world states. Unfortunately, this space is continuous, and its discretization requires a state space exponential in the number of world states. This explosion in the size of the state space is called the curse of dimensionality. To put this in precise terms, the number of $(|\mathcal{W}| - 1)$-dimensional simplexes with edge length $\epsilon$ needed to tile the simplex representing the distributions over $\mathcal{W}$, is $(\sqrt{2}/\epsilon)^{|\mathcal{W}|-1}$.

Not all Bayesian beliefs are reachable in a given POMDP. The Bayesian inference policy is deterministic, making the belief state a function of the observable history. The number of reachable beliefs is therefore bounded by the number of different histories, $|\mathcal{O}|^T$, which is unfortunately exponential in the horizon $T$, an explosion called the curse of history.

With a memory state space this large, it may not even be clear how to represent the solution policy, let alone compute a good one. There is therefore a crucial need to reduce the size of the memory space, and the number of representable beliefs. The set of all reachable beliefs needs to be clustered, perhaps implicitly, into these representable beliefs. This is the premise of all successful approaches to POMDP planning, such as point-based value iteration [11] and finite-state controllers [15].

A remaining challenge with existing approaches is that they employ heuristics for choosing the subset of representable beliefs, often without even making these beliefs explicit. The representable beliefs are the centroids of the belief space clustering that the inference policy implements, and there is a defined cost for the information lost in this clustering. A more direct approach can compare this increase in cost to the belief compression it allows, and trade them off, as in rate-distortion theory. In a very fundamental sense, the complexity of POMDP planning really stems from the difficulty of representing the informational landscape of beliefs — a *curse of information*.

**Forward-Backward Coupling**

This brings us to the final aspect of the challenge, which is the forward-backward nature of the algorithms involved. Computing beliefs and information costs requires marginal distributions to be found using a forward process. At the same time, value functions are computed using a backward process. When the forward and backward processes are separable, the problem becomes much easier to solve.

This is the case when observability is full, rendering the forward process trivial. In domains with linear-Gaussian dynamics and quadratic cost (LQG), the Gaussian distribution and the quadratic function make the processes separable, and the problem easily solvable, despite the partial observability and controllability. This hinges on the important property that the reachable beliefs are themselves Gaussian distributions over the world state space, with fixed covariances. The memory space can optimally be parameterized by the means of the beliefs, and these parameters inferred by linear updates from observations.

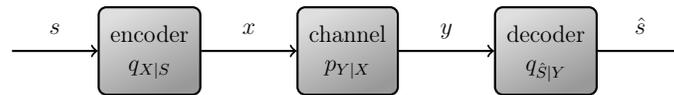

Figure 1.2: Block diagram of joint source and channel coding

In general POMDPs, the forward and backward processes are coupled, which manifests in the non-convexity of the problem, and contributes to its complexity. To illustrate this issue, consider a solution policy that neglects to infer from observations and to utilize in actions a useful piece of information about the hidden state. This solution may easily become a local optimum for optimization algorithms, since on the one hand the information cannot be used in the control policy (backward process) if it is not inferred, and on the other hand it is wasteful to commit the information to the limited memory in the inference policy (forward process) if it is never used. The control and inference policies may therefore be optimal for each other, but only locally optimal as a pair.

The coupling of the forward and the backward processes is mediated by the representation. Given the semantics of the memory states as the beliefs for which they stand, the processes become separable. This again emphasizes the importance of the challenge of finding good representations for reinforcement learning.

## 1.2 Information Theory

### 1.2.1 Rate-Distortion Theory

**Source Coding and Channel Coding**

In classic coding theory, we are concerned with the lossy transmission of a signal $s$ over a noisy memoryless channel (Figure 1.2). We are given the distribution $p_S(s)$ of the signal source, and a cost $d(s, \hat{s})$, also called distor-tion, of reconstructing $\hat{s}$ at the receiver. We are also given the cost $c(x)$ of inputting $x$ into the channel, and the probability distribution $p_{Y|X}(y|x)$ of the channel output $y$ given its input $x$. We are required to design an encoder $q_{X|S}(x|s)$ and a decoder $q_{\hat{S}|Y}(\hat{s}|y)$ that achieve a good tradeoff of the expected distortion and the expected cost; i.e., minimizes

$$\mathbb{E}_{\substack{s\sim p_S \\ x\sim q_{X|S}(\cdot|s) \\ y\sim p_{Y|X}(\cdot|x) \\ \hat{s}\sim q_{\hat{S}|Y}(\cdot|y)}} [\alpha d(s,\hat{s}) + c(x)],$$

for some tradeoff coefficient $\alpha$.

Due to the data-processing inequality [16] applied to the Markov chain $s - x - y - \hat{s}$, we have $\mathbb{I}[s;\hat{s}] \leqslant \mathbb{I}[x;y]$, where

$$\mathbb{I}[x;y] = \mathbb{E}\left[\log \frac{p_{Y|X}(y|x)}{q_Y(y)}\right]$$

is the Shannon mutual information between $x$ and $y$, and similarly for $s$ and $\hat{s}$. Since the expected distortion only depends on the joint distribution of $s$ and $\hat{s}$, and the expected cost only depends on the distribution of $x$, this suggests a separation of source coding and channel coding. In the rate-distortion problem, we determine the distribution $q_{\hat{S}|S}(\hat{s}|s)$ induced by the source coding, so that it achieves low expected distortion while also compressing the signal to keep the information rate $\mathbb{I}[s;\hat{s}]$ low. In the capacity-cost problem, we determine the distribution $q_X(x)$ induced by the channel coding, so that it achieves low expected cost while also allowing a greater information capacity $\mathbb{I}[x;y]$ on the channel.

**Source-Channel Separation**

Clearly, a solution to the joint source-channel coding problem is also feasible separately for each of the rate-distortion and capacity-cost problems. The optima of the subproblems therefore give a lower bound on the joint optimum.

Classic coding theory shows that if we allow the encoders and decoders to map a large block of inputs, as one unit, into a large block of outputs, then asymptotically for large blocks the lower bound obtained by separation is tight [16]. This separation principle allows coding theorists and practitioners to focus their efforts on one of the subproblems at a time, without having to worry about combining the solutions into a joint solution, at least in this simple setting.

When we come to apply this theory to reinforcement learning in Section 1.2.2, however, we find that encoding blocks of inputs is not an option. Our encoder and decoder are part of a controller that needs to take in a single observation and output a single action in each time step. In a large and rich system, it may be the case that each single observation or action is complex enough to treat it as a block in and of itself. More generally, however, single-letter coding is required.

A characterization of sources and channels that are matched for single-letter coding is given in [17]. In such source-channel pairs there exists a single-letter coding that achieves the lower bound of separate source and channel coding. Luckily, when designing a reinforcement learning agent, it is often possible to choose the channel that is built into the agent, so that it matches the agent's information sources. Then the problem of internal agent communication can be separated into its source-coding and channel-matching parts, and the former treated as a sequential version of the rate-distortion problem, which we discuss in Section 1.2.2. For example, in Chapter 3 we rely on the result that, if the optimal reconstruction distribution $q_{\hat{S}|S}$ is Gaussian with linear mean and fixed covariance, the linear-Gaussian channel with quadratic channel cost matches the source.

**Optimal Lossy Source Coding**

To trade off the expected distortion $\mathbb{E}[d(s, \hat{s})]$ and the information rate $\mathbb{I}[s; \hat{s}]$, we can set one of these terms as our optimization target, with a constraint that the other is not too high. The Lagrangian of this optimization problem is

$$\mathcal{F}_{q,\bar{q};\beta} = \mathbb{E}_{\substack{s \sim p \\ \hat{s} \sim q(\cdot|s)}} \left[ \tfrac{1}{\beta} \log \frac{q(\hat{s}|s)}{\bar{q}(\hat{s})} + d(s, \hat{s}) \right],$$

plus terms that constrain the distributions $q$ and $\bar{q}$ to be normalized. Here $\beta$ is a Lagrange multiplier that sets the relative marginal costs of the distortion and the information rate. This Lagrangian is also called the free energy, due to similarities to the quantity of that name in statistical physics, with $\beta$ the inverse temperature.

Note that we did not constrain the distribution $\bar{q}(\hat{s})$ over the reconstruction to be the marginal that corresponds to $p(s)$ and $q(\hat{s}|s)$. Instead, this required property will emerge as a necessary condition for a solution to be an optimum. The optimum must have gradient 0 with respect to all parameters, which implies

$$\partial_{q(\hat{s}|s)} \mathcal{F}_{q,\bar{q};\beta} = p(s) \left( \tfrac{1}{\beta} \log \frac{q(\hat{s}|s)}{\bar{q}(\hat{s})} + d(s, \hat{s}) + \tfrac{1}{\beta} + \lambda_s \right) = 0$$

$$\partial_{\bar{q}(\hat{s})} \mathcal{F}_{q,\bar{q};\beta} = \frac{\mathbb{E}_{s \sim p}[q(\hat{s}|s)]}{\beta \bar{q}(\hat{s})} + \lambda = 0,$$

and thus

$$q(\hat{s}|s) = \frac{1}{Z_\beta(s)} \bar{q}(\hat{s}) \exp(-\beta d(s, \hat{s}))$$

$$\bar{q}(\hat{s}) = \mathbb{E}_{s \sim p}[q(\hat{s}|s)].$$

Here $Z_\beta$ is a normalizing partition function.

The equations for $q$ and $\bar{q}$ provide us with a method for finding the optimal solution. We can take them as update equations, and iteratively improve a solution until it converges. This algorithm, known as Blahut-Arimoto, turns out to be an alternating projection algorithm, where each step is a projection onto a convex set, guaranteeing convergence to a global optimum [18].

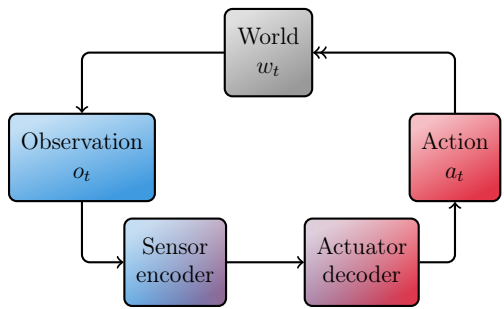

Figure 1.3: Channel from sensor to actuator of a reactive (memoryless) agent

### 1.2.2 Sequential Rate-Distortion

**Problem Formulation**

Rate-distortion theory gives us a way to model intrinsic limitations of bounded agents. Agents often operate under limited capacity of their internal storage and communication channels. Such constraints can also be used as a proxy for computational limitations caused by scarcity of information processing resources.

Intrinsic costs on information rates between the various components of an agent limit the space of policies that can feasibly be implemented. The agent may be unable to pay attention to the entire observation available in its sensors, commit the entire observable history (or a sufficient statistic of it) to memory, or specify its intended actions with perfect fidelity. Thus, in addition to the requirement to incur low extrinsic costs, and to extrinsic constraints of partial observability and partial controllability, the agent is subject to intrinsic constraints of *partial attendability* and *partial intendability*.

It is instructive to first study these constraints in reactive (memoryless) policies. Reactive agents have a sensor taking in input observations and an actuator emitting output actions, but no internal memory component. Therefore they have one memoryless channel, connecting the sensor to the actuator, as in Figure 1.3.

If the capacity of this channel is limited, we need to trade off the extrinsic expected cost $\mathbb{E}[c(w_t, a_t)]$ with the rate of information $\mathbb{I}[o_t; a_t]$ between the input observation and the output action. Similarly to the extrinsic cost, we consider the long-term average of an intrinsic cost

$$I_{\pi,\bar{\pi}} = \limsup_{T\to\infty} \frac{1}{T} \sum_{t=0}^{T-1} \mathbb{E}\left[\log \frac{\pi(a_t|o_t)}{\bar{\pi}(a_t)}\right],$$

which converges to the mutual information in the stationary distribution

$$I_\pi = \mathbb{E}_{\substack{w_t \sim \bar{p} \\ o_t \sim \sigma(\cdot|w_t) \\ a_t \sim \pi(\cdot|o_t)}} \left[\log \frac{\pi(a_t|o_t)}{\bar{\pi}(a_t)}\right],$$

with

$$\bar{\pi}(a_t) = \mathbb{E}_{\substack{w_t \sim \bar{p} \\ o_t \sim \sigma(\cdot|w_t)}} [\pi(a_t|o_t)].$$

The free energy of this sequential rate-distortion problem is

$$\mathcal{F}_{\bar{p},\pi,\bar{\pi},\nu;\beta} = \mathbb{E}_{w_t \sim \bar{p}}\Bigg[\mathbb{E}_{\substack{o_t \sim \sigma(\cdot|w_t) \\ a_t \sim \pi(\cdot|o_t)}}\Big[\tfrac{1}{\beta} \log \frac{\pi(a_t|o_t)}{\bar{\pi}(a_t)} + c(w_t, a_t) \\ + \mathbb{E}_{w_{t+1} \sim p(\cdot|w_t,a_t)}[\nu(w_{t+1})]\Big] - \nu(w_t)\Bigg].$$

**Optimality Principle**

The sequential rate-distortion Lagrangian is non-linear but convex in the policy parameters, and their global optimum given the other parameters is

$$\pi(a_t|o_t) = \frac{1}{Z_\beta(o_t)} \bar{\pi}(a_t) \exp(-\beta\, \mathbb{E}[\nu(w_{t+1})|o_t, a_t]) \qquad (7)$$

$$\bar{\pi}(a_t) = \mathbb{E}_{\substack{w_t \sim \bar{p} \\ o_t \sim \sigma(\cdot|w_t)}}[\pi(a_t|o_t)], \qquad (8)$$

where for $\pi$ the expectation is with respect to the predictive posterior

$$\mathbb{P}_\pi(w_{t+1}|o_t, a_t) = \frac{\mathbb{E}_{w_t\sim\bar{p}}[\sigma(o_t|w_t)\pi(a_t|o_t)p(w_{t+1}|w_t,a_t)]}{\mathbb{E}_{w_t\sim\bar{p}}[\sigma(o_t|w_t)\pi(a_t|o_t)]}.$$

The backward recursion is similar to the unbounded case, except that the cost-to-go now also accumulates the intrinsic informational cost

$$\nu(w_t) = \mathbb{E}_{\substack{o_t\sim\sigma(\cdot|w_t)\\a_t\sim\pi(\cdot|o_t)\\w_{t+1}\sim p(\cdot|w_t,a_t)}} \left[\tfrac{1}{\beta}\log\frac{\pi(a_t|o_t)}{\bar{\pi}(a_t)} + c(w_t,a_t) + \nu(w_{t+1})\right] - \lambda.$$

Note that these updates are inherently forward-backward, even if we assume full observability, unlike the unbounded case which is backward-only when observability is full. Both the forward and the backward processes are needed to compute the parameters for the single-step rate-distortion problem of optimizing the policy $\pi$ and its marginal $\bar{\pi}$. The forward process computes the marginal $\bar{p}$, which takes the role of the source distribution in rate-distortion theory, and is needed in the marginalization step (8) of the Blahut-Arimoto algorithm. The backward process computes the cost-to-go $\nu$, which takes the role of the distortion between the source and its reconstruction

$$d(o_t, a_t) = \mathbb{E}[\nu(w_{t+1})|o_t, a_t],$$

and is needed in the update step (7). The rate-distortion optimization of each step depends on past solutions through $\bar{p}$ and on future solutions through $\nu$.

**Extensions**

One way to avoid the complication of the forward-backward coupling is to eliminate the optimization over the parameter $\bar{\pi}$, which yields the marginal distribution, and replace it with a fixed prior $\bar{\pi}(a_t)$. More generally, we can allow this fixed prior to depend on the input as well, in the form $\rho(a_t|o_t)$. If we also restrict attention to fully observable domains, the state marginal $\bar{p}$ is no longer needed, and the entire optimization can be performed by a backward algorithm. We use a sample-based version of this approach in Section 4.1.

We can generalize to retentive (memory-utilizing) policies by considering the reduction presented in Section 1.1.2. When the agent has an internal memory component, we can think of the memory reader as another sensor, and of the memory writer as another actuator. Then we can consider the joint information rate $\mathbb{I}[m_{t-1}, o_t; m_t, a_t]$ on the channel from the sensory and memory inputs to the control and memory outputs. This is the approach taken in Section 3.2.

We cannot always assume that all sensory inputs can be encoded together by a single encoder, and the actuatory outputs decoded together. Different sensors, such as an external sensor and a memory reader, can be distributed between different components, and their inputs encoded separately by distinct encoders, and similarly for the outputs. The channels themselves can likewise be independent, each with its own capacity.

The tradeoffs involved in this more general setting are much more complicated. This is a special case of the diverse setting studied in network information theory [19], where tight bounds on the achievable rate-distortion regions are often unknown. We do not have a complete solution to this problem; however, the joint sensory-memory coding of the inputs to the inference policy can be formulated as a multiterminal source coding problem [20], where the solution exhibits an intriguing tradeoff between memory and sensing, as discussed in Section 2.1.

Finally, we mention that our formulation leads to a sequential form of the source-coding problem and is missing its channel-coding counterpart. We conjecture that some form of a sequential capacity-cost problem may be relevant to a more general setting than the one discussed in this thesis and solvable using similar methods.

## 1.3  Organization of the Dissertation

This dissertation is organized as follows. In Section 2.1 we present our approach to POMDP planning under informational constraints, and introduce the forward-backward algorithm for sequential rate-distortion. The setting in this section is restricted to passive POMDPs, where actions incur costs but do not affect the state of the world. The same algorithm can be implemented in general POMDPs, but the convergence properties are poorer. Section 2.2 explores one insightful aspect of these convergence challenges; namely, that the optimal solution can be either a limit cycle or an unstable fixed point of the update operator.

Sections 3.1 and 3.2 study the important case of a POMDP over continuous state, observation and action spaces, where the dynamics are linear with Gaussian noise, and the extrinsic cost is quadratic. This LQG setting is often better-behaved than the general setting, making it useful in practice, and allowing insights into the properties of the theory that generalize to discrete and nonlinear domains. Section 3.1 focuses on reactive (memoryless) control policies, section 3.2 applies the reduction from retentive (memory-utilizing) policies to reactive ones to study the structure of the solution in the general case, and section 3.3 contains supplementary lemmas and proofs.

In Section 4.1 we turn to the learning setting, where sample-based updates are used instead of model-based ones. Focusing on fully observable domains and Kullback-Leibler (KL) costs allows the algorithm to be backward-only, with guaranteed convergence to the global optimum. The convergence rate, as well as other desirable attributes, are shown to improve on existing reinforcement learning algorithms.

Finally, in Section 5 we discuss the contributions of this dissertation, and the results and insights obtained.

# Chapter 2

# Minimum-Information POMDP Planning

## 2.1 Bounded Planning in Passive POMDPs

Published: Roy Fox and Naftali Tishby, *Bounded Planning in Passive POMDPs*, In Proceedings of the 29th International Conference on Machine Learning (ICML), 2012.

# Bounded Planning in Passive POMDPs

**Roy Fox**  ROYF@CS.HUJI.AC.IL
**Naftali Tishby**  TISHBY@CS.HUJI.AC.IL
School of Computer Science and Engineering
The Hebrew University
Jerusalem 91904, Israel

## Abstract

In Passive POMDPs actions do not affect the world state, but still incur costs. When the agent is bounded by information-processing constraints, it can only keep an approximation of the belief. We present a variational principle for the problem of maintaining the information which is most useful for minimizing the cost, and introduce an efficient and simple algorithm for finding an optimum.

## 1. Introduction

### 1.1. Passive POMDPs Planning

Planning in Partially Observable Markov Decision Processes (POMDPs) is an important task in reinforcement learning, which models an agent's interaction with its environment as a discrete-time stochastic process. The environment goes through a sequence of *world states* $W_1, \ldots, W_n$ in a finite domain $\mathcal{W}$. These states are hidden from the agent except for an observation $O_t$ in a finite domain $\mathcal{O}$, distributed by $\sigma(O_t|W_t)$.

In the standard POMDP, the agent then chooses an action, which affects the next world state and incurs a cost. Here we consider *Passive POMDPs*, in which the action affects the cost, but not the world state. We assume that the world itself is a Markov Chain, with states governed by a time-independent transition probability function $p(W_t|W_{t-1})$ and an initial distribution $P_1(W_1)$.

The agent maintains an internal memory state $M_t$ in a finite domain $\mathcal{M}$. In each step the memory state is updated from the previous memory state and the current observation, according to a memory-state transition function $q_t(M_t|M_{t-1}, O_t)$ which serves as an *inference policy*. Figure 1 summarizes the stochastic process.

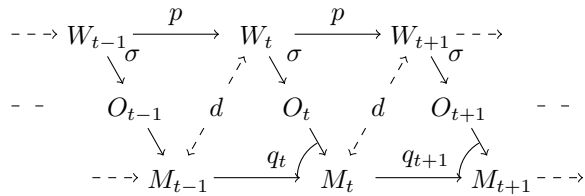

*Figure 1.* Structure of the Bayes network model of Passive POMDP planning

The agent's goal is to minimize the average expected cost of its actions. In this paper we take the agent's memory state to represent the action, and define a cost function $d : \mathcal{W} \times \mathcal{M} \to \mathbb{R}$ on the world and memory states. The planning task is then to minimize

$$\frac{1}{n} \sum_{t=1}^{n} \mathop{\mathrm{E}}_{W_t, M_t} d(W_t, M_t)$$

given the model parameters $P_1$, $p$, $\sigma$ and $d$.

A Passive POMDP can be viewed as an HMM in which inference quality is measured by a cost function. Examples of Passive POMDPs include various gambling scenarios, such as the stock exchange or horse racing, where the betting does not affect the world state. In some settings, the reward depends directly on the amount of information that the agent has on the world state (Kelly gambling, see Cover & Thomas, 2006).

When the agent is unbounded it has a simple deterministic optimal inference policy. It can maintain a *belief* $B_t(W_t|O_{(t)})$, which is the posterior probability of the world state $W_t$ given the observable history $O_{(t)} = O_1, \ldots, O_t$. The belief is a minimal sufficient statistic of $O_{(t)}$ for $W_t$, and therefore keeps all the relevant information. It can be computed sequentially by a forward algorithm, starting with $B_1(W_1|O_1) \propto P_1(W_1)\sigma(O_1|W_1)$, and at each step updating

$$B_t(W_t|O_{(t)})$$
$$\propto \sum_{w_{t-1}} B_{t-1}(w_{t-1}|O_{(t-1)}) p(W_t|w_{t-1}) \sigma(O_t|W_t),$$

normalized to be a probability vector.

## 1.2. Information Constraints

The sufficiency of the exact belief allows the agent to minimize the external cost, but it incurs significant internal costs. The amount of information which the agent needs to keep in memory can be large, and even each observation can be more than the agent can grasp. Anyway, not all of this information is equally useful in reducing external costs.

In general, the agent's information-processing capacity may be bounded in two ways:

1. The capacity of the agent's memory may limit its information rate between $M_{t-1}$ and $M_t$, to $R_M$.
2. The capacity of the channel from the agent's sensors to its memory may limit the rate at which the agent is able to process the observation $O_t$ while it is available, to $R_S$.

The requirement that the agent keeps sufficient statistics and exact beliefs is unrealistic. Rather, the agent's memory $M_t$ must be a statistic of $O_{(t)}$ which is not sufficient, but is still "good" in the sense that it keeps the external cost low. We also want it to be "minimal" for that level of quality, in terms of information-processing rates, so that the agent keeps only information which is useful enough. For each step individually, this is exactly the notion captured by rate-distortion theory, and here we have a sequential extension of it.

The main results of this paper are the formulation of the setting described above, and the introduction of an efficient and simple algorithm to solve it. We prove that the algorithm converges to a local optimum, and demonstrate in simulations the tradeoff of memory and sensing intrinsic to this setting. The application of our results to previously studied problems, and a comparison to existing algorithms, are left for future work.

This paper is organized as follows. In section 2 we formulate out setting in information-theoretical terms. In section 3 we solve the problem for one step by finding a variational principle and an efficient optimization algorithm. In section 4 we analyze the complete sequential problem and introduce an algorithm to solve it. In section 5 we show two simulations of our solution.

## 1.3. Related Work

Unconstrained planning in Passive POMDPs is easily done by maintaining the exact belief, and choosing each action to minimize the subjective expected cost. Planning in general POMDPs is harder, in one aspect due to the size of the belief space. Many algorithms plan efficiently but approximately by focusing on a subset of this space.

Several works do so by optimizing a finite-state controller of a given size (Poupart & Boutilier, 2003; Amato et al., 2010). The belief represented by each state of the controller is then the posterior probability of the world state given that memory state. A different approach is to explicitly select a subset of beliefs, and use them to guide the iterations (Pineau et al., 2003). Another is to reduce the dimension of the belief space to its principle components (Roy & Gordon, 2002).

In this paper we present the novel setting of planning in Passive POMDPs which is constrained by information capacities. This setting allows treatment of reinforcement learning in an information-theoretic framework. It may also provide a principled method for belief approximation in general POMDPs. With a fixed action policy the POMDP becomes a Passive POMDP, and a bounded inference policy can be computed. This reduces the belief space, which in turn guides the action planning. This decoupling is similar to Chrisman (1992), and will be explored in future work.

Some research treats POMDPs where the cost is the $D_{KL}$ between the distributions of the next world state when it is controlled and uncontrolled (Todorov, 2006; Kappen et al., 2009). This has interesting analogies to our setting. Our information-rate constraints define, in effect, components of the cost which are the $D_{KL}$ between the distribution of the next *memory* state and its marginals (see section 3.1). Tishby & Polani (2011) combine similar information-rate constraints of perception and action together. Future work will explore and exploit this symmetry in the special case where the memory information rate is unconstrained.

## 2. Preliminaries

Assume that the model parameters $P_1$, $p$, $\sigma$ and $d$ are given. The agent strives to find an inference policy $q_{(n)}$ such that the average expected cost satisfies

$$\frac{1}{n} \sum_{t=1}^{n} \mathop{\mathrm{E}}_{W_t, M_t} d(W_t, M_t) \leq D.$$

for the minimal $D$ possible. However, the agent operates under capacity constraints on the channels from $M_{t-1}$ and $O_t$ to $M_t$. The external cost $d$ parallels the distortion in rate-distortion theory, while the internal costs are information rates. The agent actually needs to minimize a combination of these costs.

Note that the agent will generally have some information on the next observation even before seeing it, i.e. $M_{t-1}$ and $O_t$ will not be independent. The agent therefore has some freedom in choosing what part of the information common to $M_{t-1}$ and $O_t$ it remembers, and what part it forgets and observes anew.

The average information rate in both channels combined cannot exceed their total capacity, that is

$$\frac{1}{n}\sum_{t=1}^{n} I(M_{t-1}, O_t; M_t) \leq R_M + R_S.$$

In addition, in each step the portion of the above information that is absent from $O_t$ may only be passed on the memory channel, and so

$$\frac{1}{n}\sum_{t=1}^{n} I(M_{t-1}; M_t | O_t) \leq R_M.$$

Similarly, information absent from $M_{t-1}$ is subject to the sensory channel capacity

$$\frac{1}{n}\sum_{t=1}^{n} I(O_t; M_t | M_{t-1}) \leq R_S.$$

The distortion constraint and the three information-rate constraints together form the problem of inference-planning in Passive POMDPs (*Problem 1*).

The emergence of three information-rate constraints for two channels is similar in spirit to multiterminal source coding (Berger, 1977). In their terminology, the agent needs to implement in each $M_t$ a lossy coding of the correlated sources $M_{t-1}$ and $O_t$, under capacity constraints, so as to minimize an average expected distortion. The main difference is that here we chose to allow the encoding not to be distributed, in keeping with the ability of memory to interact with perception in biological agents (Laeng & Endestad, 2012).

## 3. One-Step Optimization

### 3.1. Variational Principle

Before we consider the long-term planning required of the agent in Problem 1, we focus on the choice of $q_n$ in the final step, given the other transitions, that is, given the joint distribution of $M_{n-1}$, $W_n$ and $O_n$. We define the *joint belief* $\theta_n(M_{n-1}, W_n)$ to be the joint distribution of $M_{n-1}$ and $W_n$, and have

$$\Pr_{\theta_n}(M_{n-1}, W_n, O_n) = \theta_n(M_{n-1}, W_n)\sigma(O_n|W_n).$$

We are interested in the rate-distortion region which includes all points $(R_M, R_S, D)$ which are *achievable*, that is, for which there exists some $q_n(M_n|M_{n-1}, O_n)$ with

$$\mathcal{D}_{\theta_n}(q_n) \stackrel{def}{=} \operatorname*{E}_{W_n, M_n} d(W_n, M_n) \leq D$$

$$\mathcal{I}_{C,\theta_n}(q_n) \stackrel{def}{=} I(M_{n-1}, O_n; M_n) \leq R_M + R_S$$

$$\mathcal{I}_{M,\theta_n}(q_n) \stackrel{def}{=} I(M_{n-1}; M_n | O_n) \leq R_M$$

$$\mathcal{I}_{S,\theta_n}(q_n) \stackrel{def}{=} I(O_n; M_n | M_{n-1}) \leq R_S.$$

For any information-rate pair $(R_M, R_S)$, the minimal achievable $D$ lies on the boundary $\mathcal{D}^*_{\theta_n}(R_M, R_S)$ of the rate-distortion region. When $\theta_n$ and $q_n$ are clear from context, we refer to these quantities as $\mathcal{D}$, $\mathcal{I}_C$, $\mathcal{I}_M$, $\mathcal{I}_S$ and $\mathcal{D}^*$. We find $\mathcal{D}^*(R_M, R_S)$ by minimizing the expected distortion under information-rate constraints. The minimum exists because all our formulas are continuous, and the solution space for $q_n$ is closed.

Let $\bar{q}_n(M_n|M_{n-1})$, $\bar{q}_n(M_n|O_n)$ and $\bar{q}_n(M_n)$ be the marginals of $q_n(M_n|M_{n-1}, O_n)$. We expand the terms of the problem using these conditional probability distributions, to have

$$\min_{q_n, \bar{q}_n} \operatorname*{E}_{M_{n-1}, W_n, O_n} \sum_{m_n} q_n(m_n|M_{n-1}, O_n) d(W_n, m_n)$$

$$\operatorname*{E}_{M_{n-1}, O_n} \mathrm{D}_{\mathrm{KL}}(q_n(M_n|M_{n-1}, O_n); \bar{q}_n(M_n)) \leq R_M + R_S$$

$$\operatorname*{E}_{M_{n-1}, O_n} \mathrm{D}_{\mathrm{KL}}(q_n(M_n|M_{n-1}, O_n); \bar{q}_n(M_n|O_n)) \leq R_M$$

$$\operatorname*{E}_{M_{n-1}, O_n} \mathrm{D}_{\mathrm{KL}}(q_n(M_n|M_{n-1}, O_n); \bar{q}_n(M_n|M_{n-1})) \leq R_S$$

under normalization constraints.[1] We may waive the constraints of non-negative probabilities, which will essentially never be active as we shall see later. Also note that we optimize over $q_n$ and $\bar{q}_n$ as distinct parameters. This is justified by theorem 1 which states that, at the optimum, $\bar{q}_n$ are indeed the marginals of $q_n$.

Let the Lagrange multipliers for the constraints be $\gamma_C$, $\gamma_M$ and $\gamma_S$, and their sum $\gamma = \gamma_C + \gamma_M + \gamma_S$. Leaving aside terms of $\log q_n$, the pointwise terms in the Lagrangian will be

$$G(d, \bar{q}_n, M_{n-1}, W_n, O_n, M_n)$$

$$= d(W_n, M_n) - \gamma_C \log \bar{q}_n(M_n)$$

$$- \gamma_M \log \bar{q}_n(M_n|O_n) - \gamma_S \log \bar{q}_n(M_n|M_{n-1}).$$

In the following analysis, several expectations of this function will be useful:

- $G_{\theta_n}(d, \bar{q}_n, M_{n-1}, O_n, M_n)$
  $$= \operatorname*{E}_{W_n|M_{n-1}, O_n} G(d, \bar{q}_n, M_{n-1}, W_n, O_n, M_n),$$

- $G_{q_n}(d, \bar{q}_n, M_{n-1}, W_n)$
  $$= \operatorname*{E}_{O_n, M_n|M_{n-1}, W_n} G(d, \bar{q}_n, M_{n-1}, W_n, O_n, M_n),$$

- $G_{\theta_n, q_n}(d, \bar{q}_n)$
  $$= \operatorname*{E}_{M_{n-1}, W_n, O_n, M_n} G(d, \bar{q}_n, M_{n-1}, W_n, O_n, M_n)$$
  $$= \mathcal{D}_{\theta_n}(q_n) + \gamma_C H(\bar{q}_n(M_n))$$
  $$+ \gamma_M H(\bar{q}_n(M_n|O_n)) + \gamma_S H(\bar{q}_n(M_n|M_{n-1})),$$

---

[1] The information-rate constraints result from the $n$-step Problem 1 by fixing the first $n-1$ steps, if we consider that only two of the constraints are actually used in any instance (see corollary 3).

where $H$ is the entropy function. The Lagrangian of the problem, up to normalization terms and additive constants, can now be written as

$$\mathcal{L}_1(q_n, \bar{q}_n; \theta_n, \gamma_C, \gamma_M, \gamma_S) = G_{\theta_n, q_n}(d, \bar{q}_n) - \gamma H(q_n).$$

### 3.2. Properties of the One-Step Lagrangian

**Theorem 1.** *For any fixed $\theta_n$, $\mathcal{L}_1$ is convex in $q_n$ and $\bar{q}_n$. $\mathcal{L}_1$ is strictly convex in parameters which are conditional on $m_{n-1}$ and $o_n$ with $\Pr_{\theta_n}(m_{n-1}, o_n) > 0$, and at the minimum these satisfy*

$$q_n(M_n | M_{n-1}, O_n) \quad (1)$$
$$= \frac{\exp(-\gamma^{-1} G_{\theta_n}(d, \bar{q}_n, M_{n-1}, O_n, M_n))}{Z_n(M_{n-1}, O_n)},$$

*where $Z_n$ is a normalizing partition function, and*

$$\bar{q}_n(M_n) = \sum_{m_{n-1}, o_n} \Pr_{\theta_n}(m_{n-1}, o_n) q_n(M_n | m_{n-1}, o_n)$$
$$\bar{q}_n(M_n | O_n) = \sum_{m_{n-1}} \Pr_{\theta_n}(m_{n-1} | O_n) q_n(M_n | m_{n-1}, O_n)$$
$$\bar{q}_n(M_n | M_{n-1}) = \sum_{o_n} \Pr_{\theta_n}(o_n | M_{n-1}) q_n(M_n | M_{n-1}, o_n). \quad (2)$$

*Proof.* For any fixed $\theta_n$, $\mathcal{L}_1$ is convex since all its terms are convex. Non-zero terms only involve $m_{n-1}$ and $o_n$ with $\Pr_{\theta_n}(m_{n-1}, o_n) > 0$. Focusing on these parameters, the distortion terms are linear, and the information terms strictly convex. The unique feasible extremum of $\mathcal{L}_1$ is then the global minimum. Differentiating by each parameter gives equations 1 and 2. □

If follows from theorem 1 that complementary slackness conditions are sufficient for optimality. Table 1 shows these conditions, the information rates $(R_M, R_S)$ where the solution meets the boundary, and a subgradient of the boundary at that point. For example, if the minimum of $\mathcal{L}_1$ with $\gamma_M = \gamma_S = 0$ satisfies $\mathcal{I}_C \geq \mathcal{I}_M + \mathcal{I}_S$, then for any information-rate pair in the interval $[(\mathcal{I}_C - \mathcal{I}_S, \mathcal{I}_S), (\mathcal{I}_M, \mathcal{I}_C - \mathcal{I}_M)]$ the minimal achievable distortion is $\mathcal{D}$ and $(-\gamma_C, -\gamma_C)$ is a subgradient of the boundary.

**Theorem 2.** *For any joint belief $\theta_n$, the boundary $\mathcal{D}^*_{\theta_n}(R_M, R_S)$ of the rate-distortion region is continuous and convex. Any point $(R_M, R_S, D)$ on the boundary at which $(-\alpha_M, -\alpha_S)$ is a subgradient, is achieved by minimizing $\mathcal{L}_1$ for multipliers*

$$(\gamma_C, \gamma_M, \gamma_S) = \begin{cases} (0, \alpha_M, \alpha_S) & \text{if } \mathcal{I}_C \leq \mathcal{I}_M + \mathcal{I}_S \\ (\alpha_M, 0, \alpha_S - \alpha_M) & \text{if } \mathcal{I}_C \geq \mathcal{I}_M + \mathcal{I}_S \\ & \text{and } \alpha_M \leq \alpha_S \\ (\alpha_S, \alpha_M - \alpha_S, 0) & \text{if } \mathcal{I}_C \geq \mathcal{I}_M + \mathcal{I}_S \\ & \text{and } \alpha_M \geq \alpha_S \end{cases}$$

Table 1. Achievability of the rate-distortion boundary by a minimizer of $\mathcal{L}_1$; If the shown *Conditions* are met by the multipliers and the minimum of $\mathcal{L}_1$, then $\mathcal{D}$ is the minimal distortion for the shown *Rates*, and the shown *Subgradient* is a subgradient of $\mathcal{D}^*$ at that point

| Conditions | Rates | Subgradient |
|---|---|---|
| $\gamma_C = 0$ $\mathcal{I}_C \leq \mathcal{I}_M + \mathcal{I}_S$ | $(\mathcal{I}_M, \mathcal{I}_S)$ | $(-\gamma_M, -\gamma_S)$ |
| $\gamma_M = 0$ $\mathcal{I}_C \geq \mathcal{I}_M + \mathcal{I}_S$ | $(\mathcal{I}_C - \mathcal{I}_S, \mathcal{I}_S)$ | $(-\gamma_C, -\gamma_C - \gamma_S)$ |
| $\gamma_S = 0$ $\mathcal{I}_C \geq \mathcal{I}_M + \mathcal{I}_S$ | $(\mathcal{I}_M, \mathcal{I}_C - \mathcal{I}_M)$ | $(-\gamma_C - \gamma_M, -\gamma_C)$ |
| $\gamma_M = \gamma_S = 0$ $\mathcal{I}_C \geq \mathcal{I}_M + \mathcal{I}_S$ | $[(\mathcal{I}_C - \mathcal{I}_S, \mathcal{I}_S),$ $(\mathcal{I}_M, \mathcal{I}_C - \mathcal{I}_M)]$ | $(-\gamma_C, -\gamma_C)$ |

*Proof.* Let transitions $q_n$ and $q'_n$ achieve the rate-distortion boundary at $(R_M, R_S, D)$ and $(R'_M, R'_S, D')$, respectively, and let $0 \leq \lambda \leq 1$. Then by equations 2 and the convexity of the Kullback-Leibler divergence, the transition $\lambda q_n + (1 - \lambda) q'_n$ (over-)achieves the rate-distortion constraints $\lambda(R_M, R_S, D) + (1 - \lambda)(R'_M, R'_S, D')$. The rate-distortion region is therefore convex, and so is its boundary. The boundary is continuous by the continuity of the problem.

For a positive information-rate pair $(R_M, R_S)$, having $M_n$ independent of $M_{n-1}$ and $O_n$ makes all information-rate constraints inactive. This satisfies the Slater condition, and the multipliers detailed in the theorem are then the Karush-Kuhn-Tucker multipliers necessary for $q_n$ to be optimal. □

**Corollary 3.** *Let $\mathcal{D}^*_C$, $\mathcal{D}^*_M$ and $\mathcal{D}^*_S$ be the boundaries of the rate-distortion regions obtained by keeping each two of the three information-rate constraints. Then $\mathcal{D}^*$ is their maximum.*

### 3.3. Optimization Algorithm

An algorithm which alternatingly minimizes $\mathcal{L}_1$ over each parameter with the others fixed, in the style of Blahut-Arimoto (Cover & Thomas, 2006), will allow us to find the minimum.

**Theorem 4.** *Algorithm 1 converges[2] monotonically to the global minimum of $\mathcal{L}_1$.*

*Proof.* $\mathcal{L}_1$ is non-increasing in each iteration and is bounded from below, which guarantees its monotonic convergence. That is

$$\mathcal{L}_1(q_n^r, \bar{q}_n^r) - \mathcal{L}_1(q_n^{r+1}, \bar{q}_n^r) \xrightarrow[r \to \infty]{} 0.$$

---

[2] For the sake of clarity, here and in the rest of this paper strict convexity, uniqueness of minimum and convergence should all be taken with respect to events and transitions of positive probability, as justified by theorem 1.

**Algorithm 1** Last-Step Optimization
**Input:** $P_1, p, \sigma, d, \gamma_C, \gamma_M, \gamma_S, \theta_n$
**Output:** optimal $q_n$
   $r \leftarrow 0$
   Initialize some suggestion for $q_n^r$
   **repeat**
      Compute the marginals $\bar{q}_n^r$ of $q_n^r$ (eq. 2)
      Compute a new value for $q_n^{r+1}$ from $\bar{q}_n^r$ (eq. 1)
      $r \leftarrow r + 1$
   **until** $q_n^r$ converges

But $q_n^{r+1}$ is the unique minimum of the continuous Lagrangian. This implies that $q_n^r$ also converges to a solution $q_n^*$ with marginals $\bar{q}_n^*$. By the continuity of the Lagrangian's derivatives, they are all 0 at this solution. □

## 4. Sequential Rate-Distortion

### 4.1. Variational Principle

Returning to the entire process of Problem 1, the sequence of joint beliefs $\theta_{(2,n)} = \theta_2, \ldots, \theta_n$ depends recursively on $\theta_1$ and the policy $q_{(n)}$. For each $1 \leq t < n$

$$\theta_{t+1}(M_t, W_{t+1}) \quad (3)$$
$$= \sum_{m_{t-1}, w_t} \theta_t(m_{t-1}, w_t) \Pr_{q_t}(M_t, W_{t+1} | m_{t-1}, w_t),$$

with $\theta_1$ given as the independent distribution of $M_0$ and $W_1$.

Adding the constraints of equation 3 with multipliers $\nu_{t,m_t,w_{t+1}}$, the Lagrangian of Problem 1 is

$$\mathcal{L}_n(q_{(n)}, \bar{q}_{(n)}, \theta_{(2,n)}) = \frac{1}{n} \sum_{t=1}^{n} \mathcal{L}_1(q_t, \bar{q}_t; \theta_t, \gamma_C, \gamma_M, \gamma_S)$$

$$- \frac{1}{n} \sum_{t=1}^{n-1} \sum_{m_t, w_{t+1}} \nu_{t,m_t,w_{t+1}} \left( \theta_{t+1}(m_t, w_{t+1}) \right.$$

$$\left. - \sum_{m_{t-1}, w_t} \theta_t(m_{t-1}, w_t) \Pr_{q_t}(m_t, w_{t+1} | m_{t-1}, w_t) \right)$$

up to normalization terms and additive constants.

Solving $\mathcal{L}_n$ is much more difficult than $\mathcal{L}_1$. $\mathcal{L}_n$ is not convex, and each step may affect all future steps. Intuitively, remembering some feature of the sample in one step is less rewarding if this information is discarded in a future step, and vice versa. This leads to $\mathcal{L}_n$ having many local minima.

### 4.2. Local Optimization Algorithm

Nevertheless, Problem 1 still has some structure which can be insightful to explore. In particular, it has some interesting similarities to the standard POMDP planning problem. Differentiating $\mathcal{L}_n$ by $q_t$ we now get

$$q_t(M_t | M_{t-1}, O_t) \quad (4)$$
$$= \frac{\exp(-\gamma^{-1} G_{\theta_t}(d_t^{\vec{\nu}_t}, \bar{q}_t, M_{t-1}, O_t, M_t))}{Z_t(M_{t-1}, O_t)},$$

with

$$d_t^{\vec{\nu}_t}(W_t, M_t) = d(W_t, M_t) + \operatorname*{E}_{W_{t+1}|W_t} \nu_{t,M_t,W_{t+1}},$$

where $\vec{\nu}_n = 0$. $q_t$ now depends on the future through the multiplier vector $\vec{\nu}_t$. Note how the expectation of $\nu_{t,M_t,W_{t+1}}$ given $W_t$ plays a parallel role to that of $d(W_t, M_t)$.

$\mathcal{L}_n$ is linear in each $\theta_t$, and at the optimum must in fact be constant in every non-trivial component of $\theta_t$. This gives us a recursive formula for computing $\vec{\nu}_{t-1}$ from $\vec{\nu}_t$, $q_t$ and $\bar{q}_t$. For $1 < t \leq n$, and whenever $0 < \theta_t(M_{t-1}, W_t) < 1$, we have

$$\nu_{t-1,M_{t-1},W_t} = G_{q_t}(d_t^{\vec{\nu}_t}, \bar{q}_t, M_{t-1}, W_t) \quad (5)$$

$$- \gamma \operatorname*{E}_{O_t|W_t} H(q_t(M_t | M_{t-1}, O_t)) + \lambda_{t,W_t}.$$

Note that equation 5 is a linear backward recursion for $\vec{\nu}_t$. The multipliers $\vec{\lambda}_t$ come from the constraints that $\theta_t$ is a probability distribution function. It has no consequence, however, since it is independent of $M_{t-1}$, and is normalized out when $\vec{\nu}_{t-1}$ is used to compute $q_{t-1}$ in equation 4.

At this point, we can introduce the following generalization of algorithm 1, which finds the optimal transition $q_t$, given the joint belief $\theta_t$ and the policy suffix $q_{(t+1,n)} = q_{t+1}, \ldots, q_n$.

**Algorithm 2** One-Step Optimization
**Input:** $P_1, p, \sigma, d, \gamma_C, \gamma_M, \gamma_S, \theta_t, q_{(t+1,n)}$
**Output:** optimal $q_t$
   $r \leftarrow 0$
   Initialize some suggestion for $q_t^r$
   **repeat**
      Compute $\theta_{(t+1,n)}^r$ from $\theta_t$ and $q_{(t,n-1)}^r$ (eq. 3)
      Compute the marginals $\bar{q}_{(t,n)}^r$ of $q_{(t,n)}^r$ (eq. 2)
      Compute $\vec{\nu}_{(t,n-1)}^r$ recursively backward (eq. 5)
      Compute $q_t^{r+1}$ from $\theta_t^r$, $\bar{q}_t^r$ and $\vec{\nu}_t^r$ (eq. 4)
      $r \leftarrow r + 1$
   **until** $q_t^r$ converges

This is a forward-backward algorithm. In each iteration we compute $\theta_{(t+1,n)} = \theta_{t+1}, \ldots, \theta_n$ recursively forward, and then $\vec{\nu}_{(t,n-1)} = \vec{\nu}_t, \ldots, \vec{\nu}_{n-1}$ recursively backward. The algorithm is guaranteed to converge monotonically to an optimal solution, since $\mathcal{L}_n$ is still strictly convex in each $q_t$ separately. In fact, all our theorems and proofs regarding algorithm 1 carry over to this generalization.

### 4.3. Joint-Belief MDP

Expanding the recursion of $\vec{\nu}_t$ in equation 5 to a closed form, and disregarding $\vec{\lambda}_t$, we find that for $1 < t \leq n$ and consistent parameters[3]

$$\mathcal{L}_{n-t+1}(q_{(t,n)}; \theta_t) \tag{6}$$
$$= \frac{1}{n-t+1} \sum_{m_{t-1}, w_t} \theta_t(m_{t-1}, w_t) \nu_{t-1, m_{t-1}, w_t}.$$

If we extend the recursion by another step to define $\vec{\nu}_0$, we get that our minimization target is

$$\mathcal{L}_n(q_{(n)}; \theta_1) = \frac{1}{n} \operatorname*{E}_{M_0, W_1} \nu_{0, M_0, W_1}.$$

The minimization

$$V_t(\theta_t) = \min_{q_{(t,n)}} \operatorname*{E}_{M_{t-1}, W_t} \nu_{t-1, M_{t-1}, W_t}$$

can be looked at as the *cost-to-go* given the joint belief $\theta_t$ before step $t$. Importantly, the recursive formula 5, when minimized over $q_{(t,n)}$, is a Bellman equation. It contains a recursive term

$$\operatorname*{E}_{M_t, W_{t+1} | M_{t-1}, W_t} \nu_{t, M_t, W_{t+1}},$$

which is the expected future cost, and other terms which are the expected immediate costs, internal and external, of implementing $q_t$ in step $t$.

This suggests viewing our problem as a joint-belief MDP. Here the states are the joint beliefs $\theta_t$, the actions are $q_t$, and the next state always follows deterministically according to equation 3. This determinism allows us to use a time-dependent policy $q_{(n)}$, rather than a state-dependent one, and will prove useful in finding a solution.

The belief space of a standard POMDP can be looked at as the state space of a belief MDP, with the same actions and observations, and a linear transition function. If memory states are approximate beliefs, then our model is more like a further abstraction, where the MDP state space is the set of distributions over the belief space. Table 2 summarizes the main differences between this joint-belief MDP and the belief-MDP representation of discrete-action finite-horizon POMDPs.

---

[3]When the Lagrangian is written in terms of the policy and the initial joint belief, the other parameters are taken to be consistent with them.

*Table 2.* Differences in belief-MDP representation of POMDPs and Bounded Passive POMDPs

|  | POMDP | Bounded Passive POMDP |
|---|---|---|
| State space | beliefs, $\Delta(\mathcal{W})$ | joint beliefs, $(\Delta(\mathcal{M}))^{\mathcal{W}}$ |
| Action space | same as POMDP discrete | memory-state transitions continuous |
| State transition | stochastic linear in belief | deterministic linear in joint belief |
| Policy cost | external cost linear in belief | internal+external cost $D_{KL}$+linear in joint belief |
| Value function | piecewise-linear concave in belief | continuous concave in joint belief |

One important difference is in the structure of the value function. The expected cost $\mathcal{L}_{n-t+1}$ of a fixed policy suffix $q_{(t,n)}$ consists of some linear terms of expected distortion, but also some strictly convex terms. The latter all take the form of a Kullback-Leibler divergence between $q_{t'}$, for some $t' \geq t$, and a marginal $\bar{q}_{t'}$, the latter depending on $\theta_t$ through equations 2 and the recursion 3.

That this cost is not linear makes the representation of the value function a challenge, but a greater difficulty is the size of the policy space, which is finite in discrete-action finite-horizon POMDPs, but continuous here. Minimizing over it does not yield a piecewise-linear function of the joint belief, although it is still continuous, and the convex mixing of policies shows that it is still concave[4]. It is unclear how to finitely represent the resulting value function in our case.

### 4.4. Bounded Planning Algorithm

Perhaps surprisingly, the determinism of the joint-belief MDP allows us to define a local criterion for optimality. Together with iterations of algorithm 2 which make local improvements, this will guarantee convergence to a local optimum.

Our algorithm is a simple forward-backward algorithm, with a building block (algorithm 2) which is itself forward-backward. In each iteration we compute recursively forward the joint beliefs $\theta_{(n)}$ for the current policy $q_{(n)}$. Then we compute recursively backward a new policy $q'_{(n)}$, by finding in each step $t$ a policy suffix which is locally optimal for $\theta_t$. The criterion for optimality is that in each step we can use either $q'_{(t+1,n)}$ from the previous step or $q_{(t+1,n)}$ from the previous iteration, and whichever leads to a lower cost is chosen.

**Theorem 5.** *Algorithm 3 converges monotonically to a limit cost $\mathcal{L}^*$. For any $\epsilon \geq 0$, any $q^r_{(n)}$ which costs within $\epsilon$ of $\mathcal{L}^*$ is also within $\epsilon$ of a local minimum of*

---

[4]If rewards are used instead of costs, the value function is convex.

**Algorithm 3** Passive POMDP Bounded Planning

**Input:** $\theta_1, p, \sigma, d, \gamma_C, \gamma_M, \gamma_S, n$
**Output:** locally optimal $q_{(n)}$
   $r \leftarrow 0$
   Initialize some suggestion for $q^r_{(n)}$
   **repeat**
      $\theta^r_1 \leftarrow \theta_1$
      Compute $\theta^r_{(2,n)}$ from $\theta^r_1$ and $q^r_{(n-1)}$ (eq. 3)
      **for** $t \leftarrow n$ **to** 1 **do**
         $q^{r+1,t}_{(t,n)} \leftarrow \arg\min_{q_{(t,n)}} \mathcal{L}_{n-t+1}(q_{(t,n)}; \theta^r_t)$
         s.t. $q_{(t+1,n)} \in \left\{q^{r+1,t+1}_{(t+1,n)}, q^r_{(t+1,n)}\right\}$ (alg. 2)
      **end for**
      $q^{r+1}_{(n)} \leftarrow q^{r+1,1}_{(n)}$
      $r \leftarrow r + 1$
   **until** $\mathcal{L}_n(q^r_{(n)}; \theta_1)$ converges

*the bounded-inference-planning problem (section 2), in the sense that for any $1 \leq t \leq n$, the global minimum given $q^r_{(t-1)}$ and $q^r_{(t+1,n)}$ is at most $\epsilon$ better than $q^r_{(n)}$.*

*Proof.* In iteration $r$, $q^r_{(n)}$ from the previous iteration is feasible for $q^{r+1}_{(n)}$. Therefore the cost of $q^r_{(n)}$ is non-increasing in $r$ and converges monotonically to a limit $\mathcal{L}^*$.

Let $q^r_{(n)}$ be within some $\epsilon > 0$ of $\mathcal{L}^*$. Fix any $1 \leq t \leq n$, and let $q^*_t$ achieve the global optimum given $q^r_{(t-1)}$ and $q^r_{(t+1,n)}$. Then

$$\mathcal{L}_n(q^r_{(n)}; \theta_1) - \epsilon \leq \mathcal{L}_n(q^{r+1}_{(n)}; \theta_1)$$

$$\stackrel{(a)}{\leq} \mathcal{L}_n((q^r_{(t-1)}, q^{r+1,t}_{(t,n)}); \theta_1)$$

$$\stackrel{(b)}{\leq} \mathcal{L}_n((q^r_{(t-1)}, q^*_t, q^r_{(t+1,n)}); \theta_1),$$

where

(a) follows recursively from $(q^r_{t'}, q^{r+1,t'+1}_{(t'+1,n)})$ being feasible for $\theta^r_{t'}$ in iteration $r$, for each $1 \leq t' < t$, and

(b) follows from $(q^*_t, q^r_{(t+1,n)})$ being feasible for $\theta^r_t$ in iteration $r$.   □

Where algorithm 3 runs algorithm 2, it can initialize $q_t$ to $q^r_t$ from the previous iteration. This may speed up each iteration, particularly when the algorithm has nearly converged. In addition, when running algorithm 3 with different sets of multipliers, it converges much faster if each run is initialized with the previous result. Empirically, this also leads to much better local minima if the runs are sorted in order of decreasing multipliers.

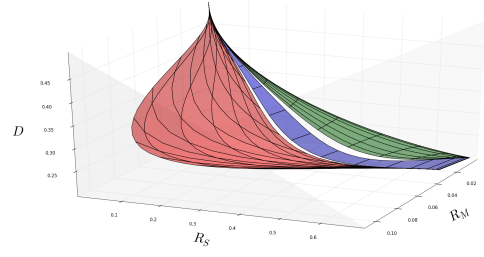

*Figure 2.* Boundary of the rate-distortion region for the sequential symmetric channel simulation
The parts from left to right: $\gamma_M = 0$; $\gamma_M = \gamma_S = 0$; $\gamma_S = 0$

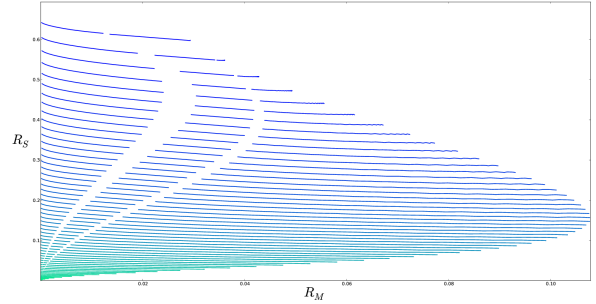

*Figure 3.* Contour map of the rate-distortion boundary for the sequential symmetric channel simulation

## 5. Simulations

### 5.1. Symmetric Channel

Figure 2 shows the boundary of the rate-distortion region for the 30-step sequential symmetric channel problem. The domains $\mathcal{W}$, $\mathcal{O}$ and $\mathcal{M}$ are all binary. The agent observes the state correctly with probability 0.8. The state remains the same for the next step independently with probability 0.8. The distortion is the delta function.

The boundary consists of three parts as in corollary 3. They have $\gamma_M = 0$ (left), $\gamma_M = \gamma_S = 0$ (middle) and $\gamma_S = 0$ (right). Empirically, taking $\gamma_C = 0$ is never feasible, as no optimal solution ever has $\mathcal{I}_C \leq \mathcal{I}_M + \mathcal{I}_S$.

To clarify this further, figure 3 shows a colored contour map of the boundary. The lower the distortion, the higher the required information rates. The tradeoff between memory and perception is illustrated by the negative slope of the contours.

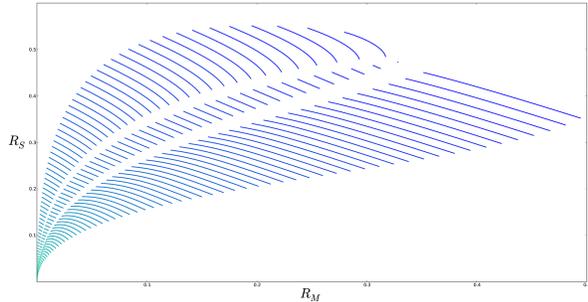

Figure 4. Contour map of the rate-distortion boundary for the Kelly gambling simulation

### 5.2. Kelly Gambling

Three horses are running in 10 races. Each horse has a fitness rating $f_i \in \{1, 2, 3\}$, and the winning horse is determined by softmax, i.e. horse $i$ wins with probability proportional to $\exp(f_i)$. Between the races, the fitness of each horse may independently grow by 1, with probability 0.1 if it is not maxed out, or drop by 1, with probability 0.1 if it is not depleted. Each horse keeps its fitness with the remaining probability.

The only observations are side races performed before each race: 2 random horses compete (with softmax) and the identities of the winner and the loser are announced. The memory state is a model of the world, consisting of the presumed fitness $\hat{f}_i$ of each horse. The log-optimal proportional gambling strategy is used (Kelly gambling, see Cover & Thomas, 2006), betting on horse $i$ a fraction of the wealth proportional to $\exp(\hat{f}_i)$. Each bet is double-or-nothing, and the distortion is the expected log return on the portfolio.

Figure 4 shows the contour map, which is not convex in this instance.

## 6. Conclusion

We have presented the problem of planning in Passive POMDPs with information-rate constraints. This problem takes the form of a sequential version of rate-distortion theory, and accordingly we were able to provide algorithms which globally optimize each step individually. Unfortunately, the full problem is not convex, and we expect that it has very hard instance sets.

Nevertheless, typical instances with some locality in their transitions and observations are expected to be easier. We have introduced an efficient and simple algorithm for finding a local minimum, and have used it to illustrate the problem with two simulations. In doing so, we have demonstrated the emergence of a memory-perception tradeoff in the problem.

Our work has been motivated by the problem of planning in general POMDPs, which may benefit from belief approximation which is principled by information theory. The application of our current results to this problem is left for future work.

## 7. Acknowledgement

This project is supported in part by the MSEE DARPA Project and by the Gatsby Charitable Foundation.

## 2.2 Optimal Selective Attention in Reactive Agents

Unpublished: Roy Fox and Naftali Tishby, *Optimal Selective Attention in Reactive Agents*, Technical Report, 2015.

# Optimal Selective Attention in Reactive Agents

**Roy Fox**     **Naftali Tishby**

School of Computer Science and Engineering
The Hebrew University

## Abstract

In POMDPs, information about the hidden state, delivered through observations, is both valuable to the agent, allowing it to base its actions on better informed internal states, and a "curse", exploding the size and diversity of the internal state space. One attempt to deal with this is to focus on reactive policies, that only base their actions on the most recent observation. However, even reactive policies can be demanding on resources, and agents need to pay selective attention to only some of the information available to them in observations. In this report we present the minimum-information principle for selective attention in reactive agents. We further motivate this approach by reducing the general problem of optimal control in POMDPs, to reactive control with complex observations. Lastly, we explore a newly discovered phenomenon of this optimization process — period doubling bifurcations. This necessitates periodic policies, and raises many more questions regarding stability, periodicity and chaos in optimal control.

## 1   Introduction

For an intelligent agent interacting with its environment, information is valuable. By observing and retaining information about its environment, the agent can form beliefs and make predictions. It represents these beliefs in an internal state, on which it can then base its actions.

If information about some event in the world is unavailable to the agent, through the lack of observability or attention, its internal state is independent of that event, and so are its actions, potentially incurring otherwise avoidable costs. The same is true if the information is only partially available, limiting the extent to which the agent's actions can depend on the state of the world.

However, information is also a "curse". Retaining much information about the world requires the agent to have a large and rich internal state space, representing diverse beliefs. This leads to complex policies for inference and control, which are computationally hard both to find and to apply. Designed agents should not be — and evolved agents are unlikely to be — more complex than is sufficient for them to perform well.

The "curse of dimensionality" [3] is the challenge of representing in the internal state space the entire belief space — the space of probability distributions over world states. The volume of this simplex is exponential in the number of world states, and approximate methods [19] [17] [1] [12] are required to explore and represent policies over this space.

The "curse of history" [14] results from representing only reachable Bayesian beliefs — posteriors of the world state given each possible observable history. The Bayesian belief is a sufficient statistic of the observable history for the world state, keeping all available information about it. Unfortunately, the size of this space can be exponential in the length of the history.

This realization immediately suggests the idea of truncating the observable history by forgetting older observations. Taken to the extreme, this leads to reactive agents [10] [20], in which each internal state can only take into account the most recent observation, discarding the previous internal state. The internal state space of reactive agents needs not be larger than the observation space, which removes the curse of history in domains where the set of observations is not too large.

**Definition 1.** *A* reactive agent *bases its actions only on the most recent observation. In contrast, a* retentive agent *can base its actions on a memory state, which is updated with each observation, and thus sum-* *marizes the entire observable history.*

A drawback of this approach is that, since the history is no longer grounded in a known initial belief, a new challenge arises of identifying which beliefs these internal states represent. This challenge generally requires forward-backward algorithms [6], as opposed to fully observable Markov Decision Processes which are solvable by backward (dynamic programming) algorithms [3].

In addition, the original difficulty remains in domains where the observation space is still too large, such as the one presented in Section 3. In this sense, the curse of history is a special case of the following principle, which we might call the "curse of information".

An agent's input — its sensors, and its memory when available — usually contains too much information for the agent to process. For the agent to encode all of this information in its new internal state, an internal state space is required that is too large to be manageable and utilized by feasible policies. As a matter of practicality, an agent must have selective attention. A retentive agent must also have selective retention [6], which is beyond the scope of this paper.

**Definition 2.** *A reactive agent (similarly, a retentive agent), is said to have* selective attention *(resp.* selective retention*) if its internal state has less information about the world state than its observation (resp. observable history) does.*

Reactive policies have been explored before in [10], with some of their challenges noted in [20]. A policy-gradient algorithm for finding such policies was presented in [7], which has the nice property of avoiding the forward-backward coupling. However, the local optimum it finds is not guaranteed to be a fixed point of the value recursion.

Information considerations in dynamical systems were presented in [24]. Algorithms were later introduced for trading off value and information in fully observable Markov Decision Processes [18] and in partially observable ones where actions have no external effect [6].

This paper offers three novel contributions, in each of the following sections.

Section 3 shows that reactive policies are as expressive as retentive policies, under proper redefinition of the model. This motivates our focus on reactive agents, at the same time that it demands a more principled cure for the curse of information than simply discarding the memory.

Section 4 provides such a principle, namely the minimum-information principle. We present the prin-ciple and formalize it, discuss its relation to source coding, and give an algorithm for its numeric solution.

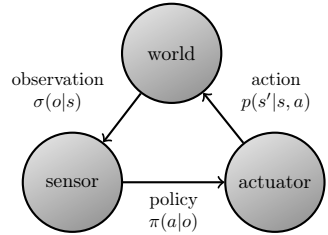

Figure 1: Schematic model of a reactive agent interacting with its environment

Section 5 demonstrates a newly discovered phenomenon in optimal control, namely the occurrence of bifurcations when attention is traded off with external cost. This is illustrated using two examples.

We conclude with a short discussion of these contributions and their consequences.

## 2 Preliminaries

We model the interaction of an intelligent agent with its environment using the formalism of Partially Observable Markov Decision Processes (POMDPs). A POMDP is a discrete-time dynamical system with state $s_t \in \mathcal{S}$. In time $t$, the system emits an observation $o_t \in \mathcal{O}$ with probability $\sigma(o_t|s_t)$. It then receives from the interacting agent an input action $a_t \in \mathcal{A}$, and transitions to a new state $s_{t+1}$ with probability $p(s_{t+1}|s_t, a_t)$. For our purposes here, the sets $\mathcal{S}$, $\mathcal{O}$ and $\mathcal{A}$ are finite, and we are only concerned with stationary (time-invariant) POMDPs, where the model parameters $p$ and $\sigma$ are fixed for every time step.

A reactive agent has no internal memory state, and can only base its actions on the most recent observation. The agent consists of two modules, the sensor making the observation $o_t$ and the actuator taking the action $a_t$ (Figure 1). The reactive policy $\pi$ of the agent is implemented by linking the two modules through a communication channel, such that the action $a_t$ is taken with probability $\pi_t(a_t|o_t)$ in reaction to observation $o_t$. The policy is called periodic with period $\mathcal{T}$ if $\pi_t = \pi_{t+\mathcal{T}}$ for every time step $t$. The policy is called stationary if it has period 1, i.e. $\pi_t$ is fixed for every time step.

The model and the policy together induce a stochastic process over the variables $\{s_t, o_t, a_t\}$ (Figure 2). Due to the agent's lack of memory, the states $\{s_t\}$ form a Markov chain. In the following we always assume that the process is ergodic. This implies that, if the agent policy has period $\mathcal{T}$, then for each phase $0 \leq t < \mathcal{T}$ there exists a unique marginal distribution $\bar{p}_t(s_t)$ that is a fixed point of the $\mathcal{T}$-step forward recursion

$$\bar{p}_t(s_{t+\mathcal{T}}) = \sum_{s_t} \bar{p}_t(s_t) P_{t,\pi}(s_{t+\mathcal{T}}|s_t)$$

with

$$P_{t,\pi}(s_{t+\mathcal{T}}|s_t) = \sum_{s_{t+1},\ldots,s_{t+\mathcal{T}-1}} \prod_{\tau=t}^{t+\mathcal{T}-1} P_{\pi_\tau}(s_{\tau+1}|s_\tau)$$

and

$$P_{\pi_\tau}(s_{\tau+1}|s_\tau) = \sum_{o_\tau, a_\tau} \sigma(o_\tau|s_\tau) \pi_\tau(a_\tau|o_\tau) p(s_{\tau+1}|s_\tau, a_\tau).$$

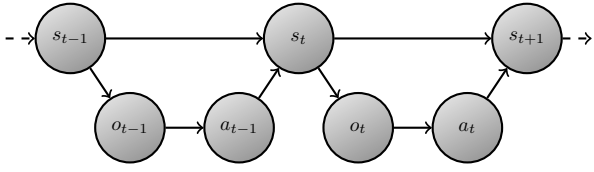

Figure 2: Graphical model of a reactive agent interacting with its environment

These marginal distributions are therefore periodic with the same period $\mathcal{T}$, i.e. $\bar{p}_t = \bar{p}_{t+\mathcal{T}}$, and inside a cycle the phases are linked through the 1-step forward recursion

$$\bar{p}_{t+1}(s_{t+1}) = \sum_{s_t} \bar{p}_t(s_t) P_{\pi_t}(s_{t+1}|s_t). \quad (1)$$

The marginal distributions also induce beliefs

$$b_t(s_t|o_t) = \frac{\bar{p}_t(s_t)\sigma(o_t|s_t)}{\bar{\sigma}_t(o_t)},$$

with

$$\bar{\sigma}_t(o_t) = \sum_{s_t} \bar{p}_t(s_t)\sigma(o_t|s_t).$$

The belief is the posterior distribution of the state given the observation.

In this paper we will have the agent incur an external nominal cost $c(s_t, a_t)$ when it takes action $a_t$ in state $s_t$, and measure the quality of a policy by the long-term average expected cost

$$\mathcal{C} = \lim_{T\to\infty} \frac{1}{T} \sum_{t=0}^{T-1} \mathrm{E}[c(s_t, a_t)]$$

in the stochastic process that the policy induces. If the policy has period $\mathcal{T}$ and the process is at its periodic marginal distribution, then

$$\mathcal{C} = \frac{1}{\mathcal{T}} \sum_{t=0}^{\mathcal{T}-1} \sum_{s_t, o_t, a_t} \bar{p}_t(s_t)\sigma(o_t|s_t)\pi_t(a_t|o_t)c(s_t, a_t).$$

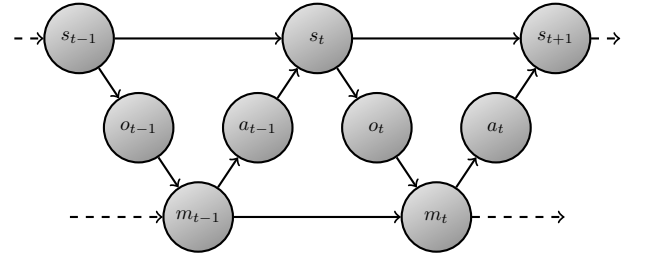

Figure 3: Graphical model of a retentive agent interacting with its environment

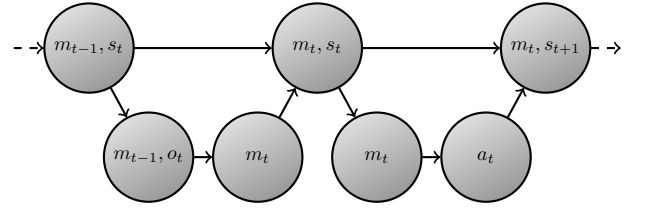

Figure 4: Reduction from a retentive policy to a reactive policy

This undiscounted expected cost is appropriate for studying stationary processes. In contrast, discounting the cost by $\gamma^t$ emphasizes transient effects, up to horizon $O(\frac{1}{1-\gamma})$. A related fault with discounting in reactive policies is discussed in [20].

## 3 Reduction from retentive to reactive policies

Consider a retentive agent [2] [6] interacting with a POMDP (Figure 3). The agent has an internal state $m_t \in \mathcal{M}$, and an inference policy $q_t$ controlling it, such that with probability $q_t(m_t|m_{t-1}, o_t)$ the memory state $m_{t-1}$ is updated to $m_t$ upon observing $o_t$ in time step $t$. The control policy $\pi_t(a_t|m_t)$ is allowed to depend not only on the most recent observation, but on the summary of the entire observable history represented in $m_t$.

In a given POMDP, retentive policies $(q, \pi)$ are much more expressive and powerful than reactive policies. Interestingly, however, there exists another (time-variant) POMDP in which $\pi' = (q, \pi)$ can be implemented as a reactive policy (Figure 4). This new POMDP is similar in spirit to the cross-product MDP [11], and the distinction between them is discussed below.

Formally, let the state space of the new POMDP be $\mathcal{S}' = \mathcal{M} \times \mathcal{S}$, the observation space $\mathcal{O}' = \mathcal{M} \times (\mathcal{O} \cup \{\bot\})$, where $\bot$ is some null-observation symbol, and the action space $\mathcal{A}' = \mathcal{M} \cup \mathcal{A}$. Let the dynamics advance at twice the frequency, with each time step tak-ing half as long. The state at time $t$ is $s'_t = (m_{t-1}, s_t)$, and it emits an observation with distribution

$$\sigma'_t((m_{t-1}, o_t)|(m_{t-1}, s_t)) = \sigma(o_t|s_t).$$

The agent, upon observing $(m_{t-1}, o_t)$, can apply its inference policy to generate the next memory state $m_t$. It then takes the "action" of committing $m_t$ to "external storage"

$$p'_t((m_t, s_t)|(m_{t-1}, s_t), m_t) = 1.$$

In this new state at time $t + \frac{1}{2}$, the committed memory state is observable

$$\sigma'_{t+\frac{1}{2}}((m_t, \perp)|(m_t, s_t)) = 1,$$

and the agent can apply its control policy to take the action $a_t$, thus controlling the transition

$$p'_{t+\frac{1}{2}}((m_t, s_{t+1})|(m_t, s_t), a_t) = p(s_{t+1}|s_t, a_t).$$

Note that it should be inadmissible for the agent to commit a memory state in a non-integer time step, or take an action in an integer one. This can be enforced by penalizing the wrong type of action, which is the main reason that the new POMDP needs to be time-variant.

Assuming that the agent follows these restrictions, the new POMDP induces the same stochastic process over the variables $\{s_t, o_t, m_t, a_t\}$ as the original one for any given policy, establishing the reduction.

Our reduction is related to the cross-product MDP of [11]. However, the two models have different formulations that serve their different purposes — where the cross-product MDP creates structure to be exploited in planning algorithms, our formulation flattens this structure to reduce the problem to a simpler one. To achieve this, instead of the implicit restriction in [11] that policies depend only on the agent state, we model the same constraint explicitly as partial observability. Furthermore, by breaking each time step into two we avoid the exponential action space of the cross-product MDP.

Lastly, an important issue to consider is the memory state space $\mathcal{M}$. The standard approach in the reinforcement learning literature is to have $\mathcal{M}$ be the belief space, the simplex of distributions over $\mathcal{S}$, and $q$ the Bayesian inference policy[1]. Such a choice would make $\mathcal{S}'$, $\mathcal{O}'$ and $\mathcal{A}'$ uncountable, as opposed to our usual assumption that these sets are finite.

---

[1]It is also common to have actions as part of the observable history, which our notation allows but does not require.

Alternatively, we can have in $\mathcal{M}$ only reachable beliefs. If the support of the inference policy $q$ is finite[2], then over a finite horizon only a finite number of beliefs are reachable. Unfortunately, due to the "curse of history", this number is exponential in the horizon, which renders this reduction — and indeed many existing approaches to POMDPs — impractical.

This difficulty underlines the need for selective attention. Theoretically, the support of $m_t$ needs never be larger than that of $(m_{t-1}, o_t)$, at least in terms of sufficient inference. However, it should practically be much smaller than that — roughly the same size as the support of $m_{t-1}$ — if the agent is to interact with the system for significant horizons without exploding in complexity. The ability of the agent to selectively attend to its input, whether from sensors or from memory, and to retain not all, but only the most useful information, is key to reducing this complexity.

This is the approach taken by Finite State Controllers (FSCs) [15], where the number of memory states is fixed. Several heuristic algorithms exist for finding a good FSC, however this problem is hard and highly non-convex. The policy of a FSC is time-invariant, and as we see in Section 5 a stationary Bellman-optimal solution is generally not stable.

## 4  Minimum-information principle

Our guiding principle in formalizing selective attention is the reduction of information complexity, as measured by the Shannon mutual information between the observation $o_t$ and the action $a_t$. We first present the principle, and then justify it by relating it to source coding. We note that numerous other justifications and connections exist, some discussed previously [9] [24] [25] [16] [8], and some should be explored further, particularly in the context of POMDP planning.

The pointwise mutual information between $o_t$ and $a_t$ in time step $t$ is given by

$$i_t(o_t, a_t) = \log \frac{\pi_t(a_t|o_t)}{\bar{\pi}_t(a_t)},$$

with

$$\bar{\pi}_t(a_t) = \sum_{o_t} \bar{\sigma}_t(o_t) \pi_t(a_t|o_t). \tag{2}$$

This can be thought of as the internal informational cost of choosing action $a_t$ in reaction to observation $o_t$. The long-term average expectation of this internal cost, similar to the external cost, is

$$\mathcal{I} = \lim_{T \to \infty} \frac{1}{T} \sum_{t=0}^{T-1} \mathrm{E}[i_t(o_t, a_t)].$$

---

[2]For example, the Bayesian inference policy is deterministic.

If the policy has period $\mathcal{T}$ and the process is at its periodic marginal distribution then

$$\mathcal{I} = \frac{1}{\mathcal{T}} \sum_{t=0}^{\mathcal{T}-1} \sum_{o_t, a_t} \bar{\sigma}_t(o_t) \pi_t(a_t|o_t) i_t(o_t, a_t)$$

$$= \frac{1}{\mathcal{T}} \sum_{t=0}^{\mathcal{T}-1} \mathrm{D}_{\mathrm{KL}}[\pi_t \| \bar{\pi}_t] = \frac{1}{\mathcal{T}} \sum_{t=0}^{\mathcal{T}-1} \mathrm{I}[o_t; a_t].$$

Here $\mathrm{D}_{\mathrm{KL}}[\pi_t \| \bar{\pi}_t]$ is the Kullback-Leibler divergence of $\pi_t$ from $\bar{\pi}_t$, and $\mathrm{I}[o_t; a_t]$ is the Shannon mutual information between $o_t$ and $a_t$.

$\mathrm{D}_{\mathrm{KL}}[\pi_t \| \bar{\pi}_t]$ is a measure of the cognitive effort required for the agent to diverge from a passive, uncontrolled policy $\bar{\pi}_t$ to an active, controlled policy $\pi_t$. Unlike [9] [24] [25], we allow the passive policy, as well as the active one, to be designed or evolved. The uncontrolled policy that minimizes the informational cost $\mathcal{I}$ is the appropriate marginal distribution of the action (2) [5].

Among agents incurring external cost $\mathcal{C} \leq C$, the simplest agent, in some sense, minimizes the internal cost $\mathcal{I}$. In other words, the agent needs to trade off its external and internal costs. To link these views, the Lagrange multiplier $\beta$ corresponding to the constraint $\mathcal{C} \leq C$ in the optimization of $\mathcal{I}$ is a conversion rate between the two types of cost. We can then write the total cost as

$$\mathcal{F} = \tfrac{1}{\beta} \mathcal{I} + \mathcal{C}.$$

$\mathcal{F}$ is called the free energy, due to its similarity to the quantity of the same name in statistical physics, with $\beta$ taking the part of the inverse temperature.

For a given $\beta$, the agent chooses its policy so as to minimize the free energy, under two constraints. First, the dynamics of the system follow the forward recursion (1). Second, $\bar{p}_t$, $\pi_t(\cdot|o_t)$ and $\bar{\pi}_t$ need to be probability distributions, each summing to 1. The constraints that they are non-negative can be ignored, since they will be either inactive or weakly active.

This gives for horizon $T$ the Lagrangian $\mathcal{L}_{\bar{p}, \pi, \bar{\pi}}$

$$= \frac{1}{T} \sum_{t=0}^{T-1} \Bigg( \sum_{s_t, o_t, a_t} \bar{p}_t(s_t) \sigma(o_t|s_t) \pi_t(a_t|o_t) f_t(s_t, o_t, a_t)$$

$$+ \sum_{s_{t+1}} \nu_{t+1}(s_{t+1}) \Bigg( \sum_{s_t} \bar{p}_t(s_t) P_{\pi_t}(s_{t+1}|s_t) - \bar{p}_{t+1}(s_{t+1}) \Bigg)$$

$$- \varphi_t \Bigg( \sum_{s_t} \bar{p}_t(s_t) - 1 \Bigg) + \eta_t \Bigg( \sum_{a_t} \bar{\pi}_t(a_t) - 1 \Bigg)$$

$$+ \sum_{o_t} \lambda_t(o_t) \Bigg( \sum_{a_t} \pi_t(a_t|o_t) - 1 \Bigg) \Bigg),$$

with

$$f_t(s_t, o_t, a_t) = \tfrac{1}{\beta} i_t(o_t, a_t) + c(s_t, a_t).$$

### 4.1 Necessary conditions for optimality

This optimization problem is far from convex, and no efficient algorithm is known for finding the global optimum. Indeed, as $\beta$ tends to infinity, the agent's policy becomes deterministic, and some problems involving deterministic reactive policies are known to be NP-complete [10].

Nevertheless, we can consider local minima by finding the first-order necessary conditions for a solution to be optimal. That is, we differentiate the Lagrangian by each of its parameters, and require that this derivative equals 0.

For $\bar{p}$, this gives us a backward recursion

$$\nu_t(s_t) = \sum_{o_t, a_t} \sigma(o_t|s_t) \pi_t(a_t|o_t) f_t(s_t, o_t, a_t)$$

$$+ \sum_{s_{t+1}} P_{\pi_t}(s_{t+1}|s_t) \nu_{t+1}(s_{t+1}) - \varphi_t. \quad (3)$$

Due to overconstraining, we have some degrees of freedom in choosing the multipliers to satisfy the Karush-Kuhn-Tucker conditions [4]. If the policy has period $\mathcal{T}$, we will choose $\varphi_t$ to also have period $\mathcal{T}$ and satisfy

$$\frac{1}{\mathcal{T}} \sum_{t=0}^{\mathcal{T}-1} \varphi_t = \mathcal{F},$$

so that $\nu_t$ also has period $\mathcal{T}$. Thus $\nu_t(s_t)$ measures the fluctuation from the average free energy $\mathcal{F}$ of the state $s_t$ in phase $t$ of the cycle.

The first-order necessary conditions for $\pi$ are

$$\pi_t(a_t|o_t) = \frac{1}{Z_t(o_t)} \bar{\pi}_t(a_t) \exp(-\beta d_t(o_t, a_t)) \quad (4)$$

with

$$d_t(o_t, a_t) = \sum_{s_t} b_t(s_t|o_t) c(s_t, a_t)$$

$$+ \sum_{s_t, s_{t+1}} b_t(s_t|o_t) p(s_{t+1}|s_t, a_t) \nu_{t+1}(s_{t+1})$$

and the normalizing partition function

$$Z_t(o_t) = \sum_{a_t} \bar{\pi}_t(a_t) \exp(-\beta d_t(o_t, a_t)),$$

and for $\bar{\pi}$ we have (2) as promised.

As $\beta$ tends to infinity, the optimal policy in (4) becomes deterministic. Together with (3), it becomes a Bellman equation [3].

For finite $\beta$, on the other hand, the optimal policy is stochastic, which is a welcome outcome in many respects. The best deterministic reactive policy is generally arbitrarily worse than the optimal stochastic reactive policy [20]. Optimality in reactive policies requires stochasticity. Unfortunately, many planning algorithms rely on the smaller space of deterministic policies (e.g. [14]), and others lack a principle by which to gauge the optimal amount of uncertainty in the agent's actions (e.g. [2]). We propose minimum information as such a principle.

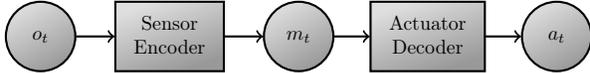

Figure 5: Reactive policy as a source-coding problem

Furthermore, in practice the model used for planning is itself uncertain. Using deterministic policies could overfit to the available data, and hinder further learning [22]. Information considerations provide a principled way of fitting the uncertainty of the policy to the uncertainty of the model [18].

In reinforcement learning, it is common to use soft-max to obtain stochastic planning and exploration policies [2] [22]. Note that soft-max is a special case of (4), with the uniform prior instead of the marginal $\bar{\pi}_t$. That information theory provides a better principle for stochasticity in optimization is further illustrated in the next subsection.

### 4.2 Sequential rate-distortion

The form of (4) and (2) may be familiar as the solution to the rate-distortion problem of lossy source coding [5]. Indeed, minimum-information optimal control can be construed as a sequential rate-distortion problem [23].

The reactive agent's policy is a channel from its sensor to its actuator (Figure 5). Following the classic model of such a channel, the sensor can be considered an encoder which, upon observing the "source" $o_t$, chooses a "codeword" $m_t$. It transmits it to the actuator, a decoder which then "reconstructs" the intended $a_t$.

For a given time step $t$, and with a source distribution $\bar{\sigma}_t(o_t)$ and a distortion function $d_t(o_t, a_t)$ fixed, this would be a standard source-coding problem. Let a feasible agent be one achieving at most $D$ expected distortion

$$\sum_{o_t, a_t} \bar{\sigma}_t(o_t) \pi_t(a_t|o_t) d_t(o_t, a_t) \leq D.$$

Now suppose we are interested in the feasible agent with the simplest internal state space, as measured by the size of the "codebook" $\mathcal{M}$. The rate-distortion theorem [5] states that the simplest feasible agent is the one minimizing $\mathrm{I}[o_t; a_t]$.

In a sequential rate-distortion problem, the solution $\pi_t$ in time step $t$ affects future source distributions $\bar{\sigma}_\tau$ in time steps $\tau > t$, as well as past distortions $d_\tau$ in time steps $\tau < t$. This creates a coupling between the forward inference process of computing marginal distributions, and the backward control process of computing value functions. This is further complicated in partially observable processes, where $d_t$ also depends on the forward inference process through the beliefs $b_t$.

Coupling makes sequential rate-distortion complex, both conceptually and computationally. Conceptually, the results of rate-distortion theory are no longer known to hold in the sequential case. If we nevertheless accept the minimum information as a solid guiding principle in our optimization, we find that this optimization is computationally hard. We can optimize the policy in each time step given the other time steps with algorithms like Blahut-Arimoto [13]. However the forward-backward algorithm for finding the overall policy is only guaranteed to converge to a local optimum [6].

### 4.3 Optimization algorithm

The forward recursion (1), the backward recursion (3), the optimal policy (4) and its marginal (2) are necessary conditions for a solution to be optimal. They also provide an algorithm for finding a good solution: iteratively plug the current solution in the right-hand side of one of the equations, to obtain a better solution, until (asymptotically) no such improvement is possible. Many existing algorithms employ a similar scheme. For example, in the Generalized Policy Iteration algorithm for planning in MDPs [22], there is some schedule for alternating between[3] policy evaluation, a variant of (3), and policy improvement, a variant of (4) with $\beta \to \infty$.

A sophisticated schedule can guarantee that the solution improves monotonically with each iteration [6]. Here we suggest the following simpler schedule, for which such a guarantee does not hold, but which empirically converges to good solutions in practice.

Repeat until convergence:

1. Compute the marginal $\bar{\pi}$ given the current solution for $\pi$, by applying (2).

2. Compute the value function $\nu$ given the current solution for $\bar{p}$, $\pi$ and $\bar{\pi}$. This can be done by iteratively applying (3) until it converges, or by solving it as a system of linear equations.

---
[3] A forward equation is not needed in fully observable problems if attention is not selective.

3. In a forward algorithm, until convergence to a limit cycle:

    (a) Compute the marginal $\bar{p}_t$ given the current solution for $\bar{p}_{t-1}$ and $\pi_{t-1}$, by applying (1).
    (b) Compute the optimal policy $\pi_t$ given the current solution for $\bar{p}_t$, $\bar{\pi}$ and $\nu$, by applying (4).

## 5 Periodicity in reactive policies

Throughout the previous sections, we always referred to periodic reactive policies rather than stationary ones, even though the POMDP itself is assumed to be stationary. Periodic reactive policies may seem to be a contradiction in terms, since their actions depend not only on the most recent observation, but also on the time $t$. They require a clock to be available to the actuator, with period that is a multiple of the policy period.

We argue that periodic policies must inevitably be a part of the solution concept of POMDPs with selective attention. When paying full attention to inputs, in the form of exact Bayesian inference, we can restrict the discussion to stationary policies [19]. When attention is partial, there are significant drawbacks to considering only stationary policies.

One drawback is that the best stationary policy is generally arbitrarily worse than the optimal periodic policy. Adapting the example in [20], consider the POMDP illustrated in Figure 6. This model has 2 states, 1 (uninformative) observation and 2 actions. The actions deterministically set the next state, and a reward (negative cost) is given for switching to the other state.

The optimal stationary retentive policy for this POMDP is to have two internal memory states, each indicating a different action, and switch between them in each time step. This policy gets the reward in each time step, but incurs 1 bit of internal cost[4].

On the other hand, a stationary reactive policy in an unobservable POMDP is just a fixed distribution over the actions, and it can be no better in this instance than the uniform distribution. This policy yields only half the expected reward, but incurs no internal cost.

Lastly, the reactive policy of period 2 which alternates between the actions also receives the full reward, at seemingly no internal cost. In fact, this would seemingly also be the preferred retentive solution, if the internal cost is taken into consideration.

Of course, counting no internal cost for a periodic pol-icy is cheating. Instead of paying attention to its sensors or memory, the agent is paying attention to a clock, but that attention is still a burden on internal resources.

---
[4]See [6] for the definition of the internal cost of a retentive policy.

Similar to the informational cost between $o_t$ and $a_t$, we need to add a term for the informational cost between $t$ and $a_t$. For a reactive policy with period $\mathcal{T}$, this cost term can naturally be defined by

$$I[t; a_t] = \frac{1}{\mathcal{T}} \sum_{t=0}^{\mathcal{T}-1} D_{KL}[\bar{\pi}_t \| \bar{\pi}]$$

$$= \frac{1}{\mathcal{T}} \sum_{t=0}^{\mathcal{T}-1} \sum_{a_t} \bar{\pi}_t(a_t) \log \frac{\bar{\pi}_t(a_t)}{\bar{\pi}(a_t)},$$

with

$$\bar{\pi}(a) = \frac{1}{\mathcal{T}} \sum_{t=0}^{\mathcal{T}-1} \bar{\pi}_t(a).$$

Here we use the fact that the phase of the cycle is distributed uniformly during the process.

Adding the term $I[t; a_t]$ to the free energy is equivalent to asserting that a clock is observable to the agent, and that attention to it is as costly as to any other part of the observation. The pointwise informational cost is now

$$\tilde{i}_t(o_t, a_t) = \log \frac{\pi_t(a_t|o_t)}{\bar{\pi}(a_t)},$$

and the average expected internal cost is

$$\tilde{\mathcal{I}} = \frac{1}{\mathcal{T}} \sum_{t=0}^{\mathcal{T}-1} I[o_t; a_t] + I[t; a_t]$$

$$= I[o_t; a_t|t] + I[t; a_t] = I[t, o_t; a_t].$$

The values of $\tilde{f}_t$, $\tilde{\nu}_t$ and $\tilde{d}_t$ change accordingly, and the optimal policy is now

$$\pi_t(a_t|o_t) = \frac{1}{\tilde{Z}_t(o_t)} \bar{\pi}(a_t) \exp(-\beta \tilde{d}_t(o_t, a_t)),$$

with the proper partition function $\tilde{Z}_t(o_t)$.

This allows us to consider policies which are "softly periodic", in that they attend to some but not all time information. Figure 7 shows the information-cost curve for the POMDP in Figure 6, and Figure 8 shows the final-state diagram for the iterative algorithm with the schedule in Section 4.3.

Interestingly, this problem exhibits a bifurcation at $\beta = 1$. Below this value, information is too costly, and the optimal solution is the stationary uniform policy. At $\beta = 1$, the system undergoes a period-doubling bifurcation, and above this value the optimal policy becomes periodic with period 2 — the two phases of the cycle are given by the two branches in Figure 8. The "hardness" of this periodicity, as measured by the information $I[t;a_t]$, grows continuously from 0, and tends to 1 bit as $\beta$ tends to infinity.

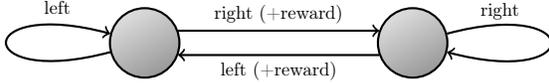

Figure 6: An unobservable POMDP, having an optimal policy of period 2

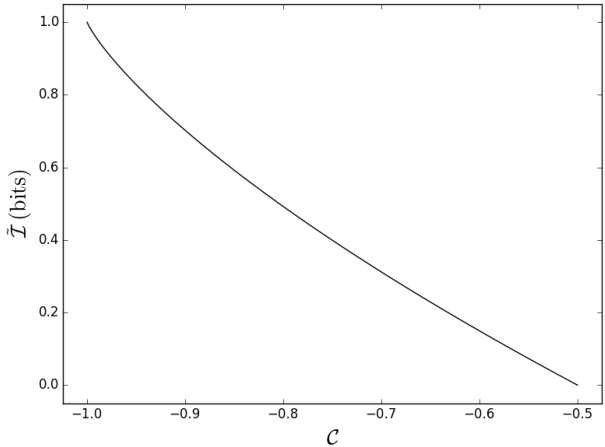

Figure 7: Information-cost curve for the POMDP in Figure 6; Points on the curve were achieved by the algorithm with different values of $\beta$, and points above it are achievable; The curve is convex, with slope $-\beta$

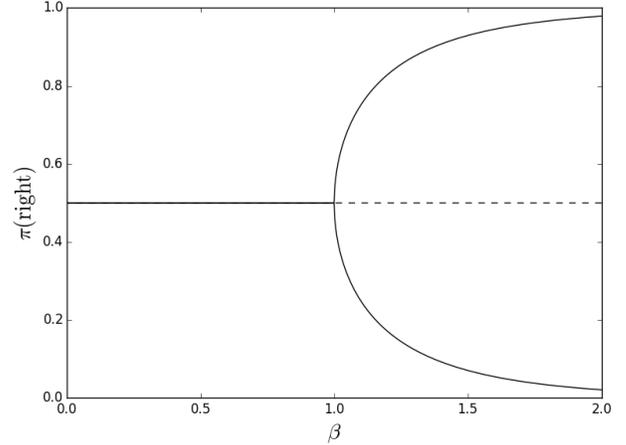

Figure 8: Finite-state diagram for the iterative algorithm applied to the POMDP in Figure 6, as a function of the cost conversion rate $\beta$; Points on the curve are the probability of taking the action "right" in each phase of the limit cycle of the algorithm, when run to convergence with the given $\beta$

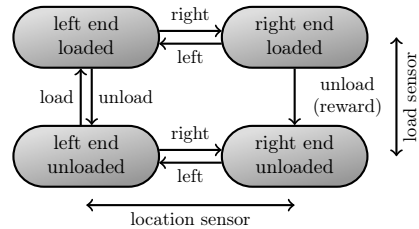

Figure 9: A POMDP of a robot moving items from the left end of a corridor to the right one; Shown actions succeed with probability 0.8, otherwise the state remains the same; Location sensor correct with probability 0.88, load sensor with probability 0.7

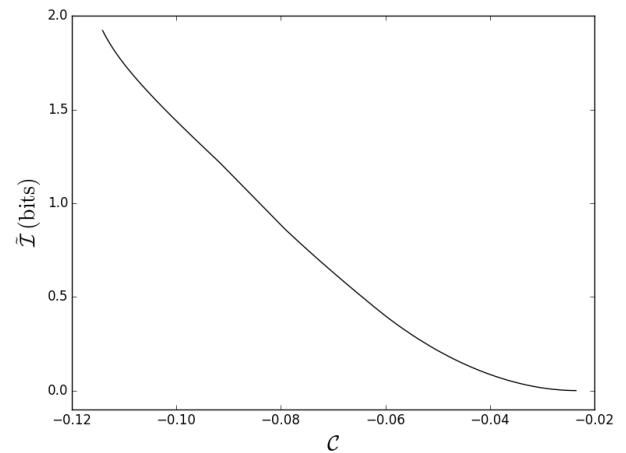

Figure 10: Information-cost curve for the POMDP in Figure 9

Above the critical point, a third solution exists, which is a fixed point of the optimization schedule (the dashed line in Figure 8). This solution is the optimal stationary reactive policy, but it is an unstable fixed point: starting the optimization from a small perturbation of this solution does not converge back to it, but diverges until it reaches the periodic solution. Thus we have a supercritical pitchfork bifurcation [21].

The instability of the stationary solution is another paramount reason for allowing periodic policies. It would be practically impossible to find a stationary solution using Bellman-like variational methods, as the one presented in this paper. In contrast, approaches such as policy-gradient methods [2] generally can find stationary solutions, but these are generally not fixed points of a Bellman recursion, and are thus not Bellman-optimal [3].

### 5.1 Robot example

As another example, consider the POMDP illustrated in Figure 9. Here a robot is engaged in moving items

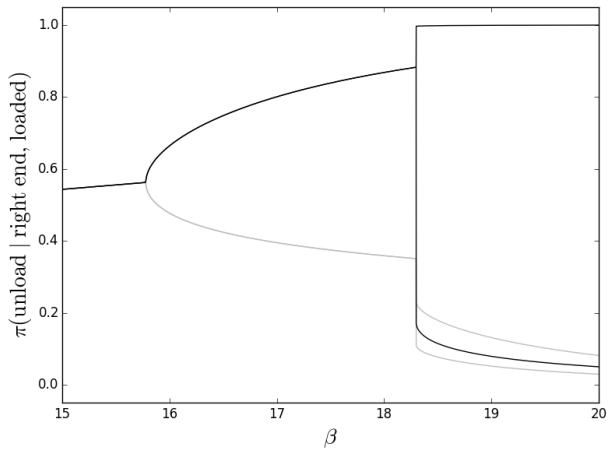

Figure 11: Finite-state diagram for the POMDP in Figure 9; Points on the curve are the probability of taking the action "unload" when the sensors indicate "right end" and "loaded", in each phase of the limit cycle; The lower branch is grayed out for clarity

from the left end of a corridor to the right one. The robot can be in one of 4 states: it can be at either end of the corridor, and it can be carrying an item or not. It has 4 actions: to move to the left end of the corridor, or to the right, or to pick up or put down an item. However, an action can fail with probability 0.2, leaving the robot at the same state. The robot can only pick up an item at the left end of the corridor, and it receives a reward for dropping an item at the right end.

The robot has 4 possible observations from two binary sensors, telling it its position and whether or not it is carrying an item. The location sensor is more reliable, showing the correct position with probability 0.88. The load sensor only shows the correct load state with probability 0.7. The parameters were selected for visual clarity of the results.

Figure 10 shows the information-cost curve for this problem. Here there are two phase transitions, where the period doubles to 2, then again to 4 (Figure 11). When attention is scarce, the robot's actions are more uniformly random. In this situation, the sensors, although noisy, carry more relevant information than the clock, since they better correspond to the actual state.

As attention increases with $\beta$, the robot relies more and more on its sensors. Conditional on the observation, the robot makes its actions less and less stochastic. At some point, the policy is reliable enough that the clock has more relevant information about the load state than the noisy sensor. At that point, a pitchfork bifurcation occurs, and the robot begins to rely mostly on the parity of the clock to decide when to move, and on the location sensor to decide where to move and whether to load or unload.

With the parameters above, as $\beta$ keeps increasing, the clock eventually becomes even more reliable than the location sensor, and a second period doubling occurs, to period 4. Asymptotically as $\beta$ grows to infinity, the clock signal takes precedence over both sensors, and the agent unloads its cargo on schedule even if its sensors tells it that it is dislocated or empty handed.

## 6 Discussion

In this paper we presented three novel results involving reactive agents interacting with partially observable systems. We have motivated the focus on reactive policies through a reduction from retentive policies, introduced a principle and an algorithm for optimizing reactive policies, and explored a surprising aspect of their phenomenology.

We conclude with a few remarks on the implication of each contribution.

### 6.1 Selective attention as clustering

Information-constrained clustering can also be construed as source coding [16], so that the data to be clustered is considered the source, and the cluster centroids the reconstruction. Following the relation we show between selective attention and source coding, we can think of a reactive policy as a soft clustering of observations into actions.

With the information constraint removed, the clustering becomes hard, mapping each data point to its closest centroid. Similarly in our case, as $\beta$ grows the policy becomes more deterministic, until at $\beta \to \infty$ it always picks the optimal action for each observation.

The implication of viewing reactive policies as clustering is that actions should generally be simpler, and never more complex, than the observations on which they are based. Indeed, there is a duality between observations and actions, and between selective attention (the retained part of the observation) and selective action (the intended part of the action, as divergence from the prior $\bar{\pi}$). Information that is not retained cannot be used for choosing actions, and there is no point in retaining information that is not used.

### 6.2 Implications of selective attention for retentive agents

In this paper we have focused on reactive agents, and introduced the minimum-information principle for optimal selective attention. However, as the reduction in Section 3 shows, this has implications for retentive agents as well.

The effect of selective attention is to make internal states less complex than their inputs, by discarding information that is not useful enough. When applied to the inference policy, this leads to approximate inference, that trades off the external value of information in guiding actions with its internal cost in information complexity. In fact, an inference process in POMDPs is equivalent to sequential clustering. With each new observation $o_t$, the pair $(m_{t-1}, o_t)$ is clustered into a new internal state $m_t$.

The major challenge when planning in POMDPs is approximating the Bayesian belief in such a way that allows efficient planning and execution, while not losing too much value. Selective attention, and in this case retention, is precisely such a principle. The application of this approach to retentive agents is left for future work.

### 6.3 Policy bifurcations and chaos theory

We have discovered the occurrence of bifurcations in the optimization process of reactive policies. It presents many of the characterizing features of chaos theory of iterated functions, such as period doubling and slow convergence near the bifurcation points. We expect to see many more such features in other, more complex systems. We conjecture that systems with more states, perhaps infinitely many, can present a cascade of bifurcations, leading to aperiodicity and chaos.

A full investigation of the bearings of the theories of bifurcation and chaos to optimal control in dynamical systems is beyond the scope of this report. To the extent that such a connection exists, it could be of profound philosophical implications, as it could indicate that intelligent agents interacting with complex environments must choose among the following alternatives:

- Plan with very little attention of their inputs
- Plan for very short horizons
- Plan with some degree of inability to identify their own value function or predict their own future actions.

# Chapter 3

# Minimum-Information LQG Control

## 3.1 Part I: Memoryless Controllers

# Minimum-Information LQG Control
# Part I: Memoryless Controllers

Roy Fox[†] and Naftali Tishby[†]

*Abstract*— With the increased demand for power efficiency in feedback-control systems, communication is becoming a limiting factor, raising the need to trade off the external cost that they incur with the capacity of the controller's communication channels. With a proper design of the channels, this translates into a sequential rate-distortion problem, where we minimize the rate of information required for the controller's operation under a constraint on its external cost. Memoryless controllers are of particular interest both for the simplicity and frugality of their implementation and as a basis for studying more complex controllers. In this paper we present the optimality principle for memoryless linear controllers that utilize minimal information rates to achieve a guaranteed external-cost level. We also study the interesting and useful phenomenology of the optimal controller, such as the principled reduction of its order.

## I. INTRODUCTION

The modern technology industry is deploying artificial sensing-acting agents everywhere [1]. From smart-home devices to manufacturing robots to outdoor vehicles and from nanoscale machines to space rockets, these agents sense their environment and act on it in a perception-action cycle [2].

When these agents are centrally controlled or when the sensors and the actuators are distributed, this control process relies on the ability to communicate the observations to the controller and the intended actions to the actuators. Autonomous agents likewise require sufficient capacity for the internal communication between their sensor and actuator components. As devices become smaller and more ubiquitous, power efficiency and physical restrictions dictate that communication become a limiting factor in the agent's operation.

Classic optimal control theory [3] is unconcerned with the costs and the limitations of communicating the information needed for the controller's operation. In the past two decades, however, a large body of research has been dedicated to this issue ([4]–[7] and references therein).

The perception-action cycle between a controller and its environment (Figure 1) consists of multiple channels and the capacity of any of them can be limited. Accordingly, various information rates can be considered. Our guiding principle in this work is to measure the information complexity of the controller's internal representation by asking *"How much information does the controller have on the past?"*. The past is informative of the future [8] and some information in past observations is useful in controlling the future. We therefore seek a trade-off between the external cost incurred by the system and the internal cost of the communication resources spent by the controller in reducing that external cost. This trade-off is often formulated as an optimization problem, where one cost is constrained and the other minimized.

When the controller has no internal memory, it can only attend to its most recent input observation, perhaps selectively. The degree of this attention, measured by the amount of Shannon information about the input observation that is utilized in the output control, is a lower bound on the required capacity of the communication channel between the controller's sensor and its actuator (see Figure 3).

Our motivation in considering memoryless controllers is twofold. First, there are applications in which having any significant memory capacity within the controller is impractical. When the system is complex and the controller's hardware and resources are limited, they may be inadequate for maintaining any significant representation of the environment. In this case, a memoryless controller is the more cost-effective solution and sometimes the only feasible one. Memoryless controllers have been studied before, particularly in the contexts of delay [9]–[11] and discrete state-spaces [12]–[14].

Second, we show in Part II of this work [15] how to formulate the problem of optimizing a bounded retentive (memory-utilizing) controller as an equivalent problem of optimizing a bounded memoryless controller. This reduction enables us to reuse the solution derived in this paper in solving the bounded retentive control problem.

Much of the related existing research has been concerned with the issue of stabilizability of an unstable plant over communication channels that are limited in some way: quantization [16]–[20], noise [21]–[23], delay [24] and fading [25]. Our current work reduces in the stabilizable case to known results, and this analysis will be included in an upcoming paper.

Other early publications proposed heuristic approximate solutions to the problem of optimal control with finite precision [26], [27]. More recently, the problem of optimal control over limited-capacity channels has been studied, with various information patterns in the sensor-side encoder and the actuator-side decoder: unlimited encoder and decoder memory with full feedback [28]–[31], unlimited encoder memory and memoryless decoder [32], and unlimited decoder memory with some feedback to the encoder [33].

A special case of our current work was studied in [34]. Their setting is fully observable and scalar, whereas we treat the much more general setting of partially observable vector spaces. Our main result reduces in this simple case to one of their solutions, implying that their other proposed solution is never optimal.

[†]School of Computer Science and Engineering, The Hebrew University, {royf,tishby}@cs.huji.ac.il

[*]This work was supported by the DARPA MSEE Program, the Gatsby Charitable Foundation, the Israel Science Foundation and the Intel ICRI-CI Institute

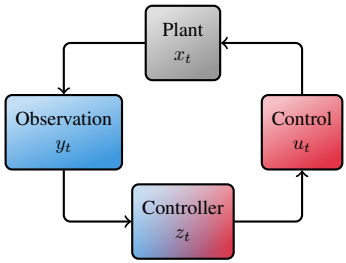

Fig. 1. Block diagram of a closed-loop control system

In this paper we make three contributions. First, we present a method for designing memoryless linear controllers that utilize minimal information rates to achieve a guaranteed external cost level. To our knowledge, this is the first treatment of information considerations in continuous-space control problems where neither the controller's sensor nor its actuator have unbounded memory capacity.

Second, we derive a solution that has a particularly explicit form, allowing direct numerical computation. Unlike classic controllers, which are designed by separable forward and backward Riccati equations [35], our forward and backward recursions are coupled. Yet each forward and backward step is given in closed form up to eigenvalue decomposition (EVD) operations. This is in contrast to the semidefinite programs (SDP) in [29], [31], [36], which require external solvers.

Third, we study the interesting and useful phenomenology of the optimal controller. It manifests a water-filling effect [37], which is a principled criterion for the selection of the active controller modes and their magnitudes. By trading off external cost to reduce the controller's communication resources we also reduce its order in a principled way.

In Section II we define the LQG task that the controller should solve. In Section III we present the memoryless control model and the information considerations involved. In Section IV we find the conditions satisfied by the optimal linear solution and discuss its intriguing phenomenology. More discussion and an illustrative example can be found in Part II of this work [15].

## II. CONTROL TASK

We consider the closed-loop control problem depicted schematically in Figure 1, where an agent (controller) is interacting with its environment (plant). When the plant is in state $x_t \in \mathbb{R}^n$, it emits an observation $y_t \in \mathbb{R}^k$, takes in a control input $u_t \in \mathbb{R}^\ell$ and undergoes a stochastic state transition. The goal of the controller is to reduce the long-term average expectation of some cost rate $\mathcal{J}_t(x_t, u_t)$.

A controller $\pi$ defines the possibly stochastic mapping from the observable history $y^t = \{y_\tau\}_{\tau \leq t}$ into the control $u_t$. The plant and the controller, under some initial conditions, jointly induce a stochastic process over the infinite sequence of variables $\{x_t, y_t, u_t\}$.

Our focus in this work is on discrete-time systems with linear dynamics, Gaussian noise and quadratic cost rate (LQG). For simplicity, all elements are taken to be homogeneous, i.e. centered at the origin, and time-invariant. We note that all our results hold without these assumptions, with the appropriate adjustments, as usual in LQG problems [3].

*Definition 1:* A linear-Gaussian time-invariant (LTI) plant $\langle A, B, C, \Sigma_\xi, \Sigma_\epsilon \rangle$ has state dynamics

$$x_{t+1} = Ax_t + Bu_t + \xi_t; \qquad \xi_t \sim \mathcal{N}(0, \Sigma_\xi),$$

where $A \in \mathbb{R}^{n \times n}$, $B \in \mathbb{R}^{n \times \ell}$, $\Sigma_\xi \in \mathbb{S}_+^n$ is in the positive-semidefinite cone and $\xi_t$ is independent of $(x^t, y^t, u^t) = \{x_\tau, y_\tau, u_\tau\}_{\tau \leq t}$. The observation dynamics are

$$y_t = Cx_t + \epsilon_t; \qquad \epsilon_t \sim \mathcal{N}(0, \Sigma_\epsilon),$$

where $C \in \mathbb{R}^{k \times n}$, $\Sigma_\epsilon \in \mathbb{S}_+^k$ and $\epsilon_t$ is independent of $(y^{t-1}, u^{t-1}, x^t)$.

*Definition 2:* A linear-quadratic-Gaussian (LQG) task $\langle A, B, C, \Sigma_\xi, \Sigma_\epsilon, Q, R \rangle$ involves a LTI plant and the cost rate

$$\mathcal{J}_t = \tfrac{1}{2}(x_t^\intercal Q x_t + u_t^\intercal R u_t),$$

where $Q \in \mathbb{S}_+^n$ and $R \in \mathbb{S}_+^\ell$. The task is to achieve a low long-term average expected cost rate, with respect to the distribution induced by the plant and the controller $\pi$

$$\mathcal{J}_\pi = \limsup_{T \to \infty} \frac{1}{T} \sum_{t=1}^{T} \mathbb{E}_\pi[\mathcal{J}_t]. \qquad (1)$$

We are particularly interested in controllers which are time-invariant, i.e. have $\pi(u_t | y^t)$ independent of $t$, and which induce a stationary process, independent of any initial conditions. In a stationary process, the marginal joint distribution of $(x_t, y_t, u_t)$ is time-invariant and we can replace the long-term average expected cost rate (1) with the expected cost rate in the stationary marginal distribution.

We denote by $\Sigma_x \in \mathbb{S}_+^n$ and $\Sigma_y \in \mathbb{S}_+^k$, respectively, the stationary covariances of the state and of the observation, assuming they exist and are finite. They are related through

$$\Sigma_y = C \Sigma_x C^\intercal + \Sigma_\epsilon .$$

If $x_t$ and $y_t$ are jointly Gaussian with mean 0, they satisfy the reverse relation

$$x_t = K y_t + \kappa_t; \qquad \kappa_t \sim \mathcal{N}(0, \Sigma_\kappa),$$

where the residual state noise $\kappa_t$ is independent of $y_t$ (but not of the past of the process), and

$$K = \Sigma_x C^\intercal \Sigma_y^\dagger$$
$$\Sigma_\kappa = \Sigma_x - \Sigma_x C^\intercal \Sigma_y^\dagger C \Sigma_x,$$

with $\cdot^\dagger$ the Moore-Penrose pseudoinverse. If the entire process has mean 0, the stationary expected cost rate (1) is given by

$$\mathcal{J}_\pi = \tfrac{1}{2}(\mathrm{tr}(Q \Sigma_x) + \mathrm{tr}(R \Sigma_u)), \qquad (2)$$

where $\Sigma_u \in \mathbb{S}_+^\ell$ is the stationary control covariance.

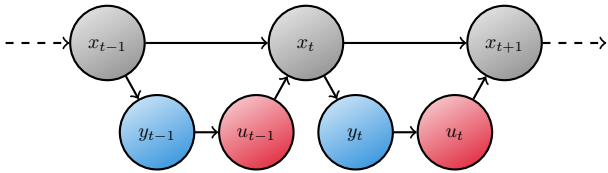

Fig. 2. Bayesian network of memoryless control

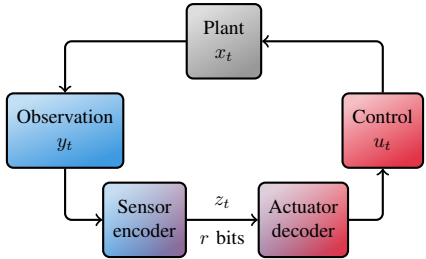

Fig. 3. The communication channel from the sensor to the actuator

## III. BOUNDED MEMORYLESS CONTROLLERS

### A. Control model

In this section we introduce memoryless controllers with bounded communication resources. A memoryless controller is simply a possibly stochastic mapping from its input observation $y_t$ into its output control $u_t$ without any memory of past observations.

*Definition 3:* A controller is memoryless if the control depends only on the most recent observation; that is, $u_t$ is independent of $(y^{t-1}, u^{t-1}, x^t)$ given $y_t$.

A system including a memoryless controller satisfies the Bayesian network in Figure 2.

Optimization over the space of all measurable control laws is hard to analyze and the optimal controller can be hard to implement. It is therefore practical to require the control law to have a certain form, most commonly the linear-Gaussian time-invariant (LTI) form. LTI controllers induce, jointly with a LTI plant, a Gaussian stochastic process. When the process is stable, it has a unique stationary distribution that is independent of any initial conditions. Linear controllers with limited memory are known not to be optimal for all control problems [38], [39]. The conditions under which there exists an optimal memoryless controller which is LTI, so that no performance is lost by focusing our attention on such controllers, are beyond the scope of this paper.

*Definition 4:* A memoryless linear-Gaussian time-invariant (LTI) controller has control law of the form

$$u_t = H y_t + \eta_t; \qquad \eta_t \sim \mathcal{N}(0, \Sigma_\eta), \qquad (3)$$

where $H \in \mathbb{R}^{\ell \times k}$, $\Sigma_\eta \in \mathbb{S}_+^\ell$ and $\eta_t$ is independent of $y_t$.

### B. Information considerations

Our controller is bounded and operates under limitations on its capacity to process the observation and produce the control. To measure this internal complexity of the controller, we consider a memoryless communication channel from the sensor to the actuator with limited capacity (Figure 3).

For example, we can consider a noiseless binary channel and measure the controller complexity by the number $r$ of bits per time step that it transmits from its sensor to its actuator. This requires the controller's sensor to perform lossy source coding of the observation $y_t$ by compressing it into a binary string representation $z_t \in \{0,1\}^r$. This representation is transmitted losslessly and reconstructed by the controller's actuator as a control $u_t$. Since the controller is memoryless, both the encoder and the decoder are memoryless.

In this sense, the dynamical control problem can be thought of as a sequential rate-distortion (SRD) problem [28], [36]. Unlike the standard one-shot rate-distortion (RD) problem [37], [40], in a SRD problem the output distribution affects the future of the process. This often creates a coupling between the forward inference process that determines the marginal distributions and the backward control process that determines the cost-to-go, i.e. the distortion. We note that without control [36] the decoder only affects the controller part of the future trajectory; however, this distinction is of minor consequence for the SRD aspect of the problem [41].

Following rate-distortion theory, we find that the bit rate $r$ required for this process is linked to the Shannon mutual information between the observation and the control, defined by

$$\mathbb{I}[y_t; u_t] = \mathbb{E}\left[\log \frac{f(y_t, u_t)}{f(y_t) f(u_t)}\right],$$

where $f$ denotes the various probability density functions, as indicated by their arguments. The bit rate is bounded from below by the information rate due to the data-processing inequality [37]

$$\mathbb{I}[y_t; u_t] \leq \mathbb{I}[y_t; z_t] \leq \mathbb{H}[z_t] \leq r \log 2,$$

where

$$\mathbb{H}[z_t] = -\mathbb{E}[\log \Pr(z_t)]$$

is the discrete Shannon entropy of $z_t$.

In classic information theory, this bound is made asymptotically tight by jointly encoding a long block of observations and jointly decoding a long block of controls. In our setting, this is impossible due to the causal nature of the plant-controller interaction. Thus, unfortunately, the bound is generally not tight for discrete channels. We can nevertheless expect it to be a good approximation, if we draw intuition from the stabilizability problem, where the informational lower bound is approximated by a known upper bound [42].

In applications, it is often possible to make design choices regarding the channel itself. If we can design the channel to be perfectly matched to the optimal LTI control law, no block coding will be needed [43]. When the controller is LTI, it is more practical to take the channel in Figure 3 to be itself linear-Gaussian instead of binary. There exists an additive Gaussian noise channel with a signal power cost that is perfectly matched to our optimal controller in Theorem 1. With such a channel, the information rate is optimally equal to the channel capacity and a constraint on the information rate $\mathbb{I}[y_t; u_t]$ is equivalent to a constraint on the expected power available for transmission on the channel. We develop these results in the Supplementary Material[1] (SM), Appendix I.

We are thus interested in a LTI controller $\pi$ that minimizes the long-term average

$$\mathcal{I}_\pi = \limsup_{T \to \infty} \frac{1}{T} \sum_{t=1}^{T} \mathcal{I}_t \qquad (4)$$

of the controller's internal information rate $\mathcal{I}_t = \mathbb{I}[y_t; u_t]$, under the constraint that it achieves some guarantee level $c$ of expected cost rate.

*Problem 1:* Given a LQG task, the bounded memoryless LTI controller optimization problem is

$$\min_\pi \quad \mathcal{I}_\pi$$
$$\text{s.t.} \quad \mathcal{J}_\pi \leq c,$$

with $\mathcal{I}_\pi$ as in (4), where $\mathcal{I}_t = \mathbb{I}[y_t; u_t]$, and with $u_t$ as in (3).

## IV. MAIN RESULT

### A. Optimality conditions

In this section we derive the optimality conditions for a bounded memoryless LTI controller. These conditions are summarized in Theorem 1 below.

Analysis of Problem 1 starts with considering the minimum mean square error (MMSE) estimators

$$\hat{x}_{y_t} = \mathbb{E}[x_t | y_t] = K y_t$$
$$\hat{x}_{u_t} = \mathbb{E}[x_t | u_t] = \Sigma_{x;u} \Sigma_u^\dagger u_t,$$

respectively for the state given the observation and the control. Here $\Sigma_{x;u} = \mathbb{E}[x_t u_t^\mathsf{T}]$ is the covariance matrix between $x_t$ and $u_t$. This implies that $\hat{x}_{y_t}$ and $\hat{x}_{u_t}$ are also 0-mean and jointly Gaussian with the other variables. At this point, it is useful to state a few properties of MMSE estimators of Gaussian variables.

*Lemma 1:* Let $x$ and $\hat{x}$ be 0-mean jointly Gaussian random variables. The following properties are equivalent:
1) There exists a random variable $u$, jointly Gaussian with $x$, such that $\hat{x}(u) = \arg\min_{\hat{x}} \mathbb{E}[\|\hat{x} - x\|^2 | u] = \mathbb{E}[x|u]$.
2) $\Sigma_{\hat{x};x} = \Sigma_{\hat{x}}$.
3) $\Sigma_{x|\hat{x}} = \Sigma_x - \Sigma_{\hat{x}}$, where $\Sigma_{x|\hat{x}}$ is the conditional covariance matrix of $x$ given $\hat{x}$, implying $\Sigma_x \succeq \Sigma_{\hat{x}}$.
4) $\hat{x} = \mathbb{E}[x|\hat{x}]$.

Such $\hat{x}$ is called a minimum mean square error (MMSE) estimator (of $u$) for $x$.

*Proof:* See SM, Appendix II. □

Since the conditional covariance $\Sigma_{x|u}$ of $x_t$ given $u_t$ is deterministic, i.e. is not a random variable, the conditional expectation of $x_t$ given $u_t$, i.e. $\hat{x}_{u_t}$, is a sufficient statistic of $u_t$ for $x_t$, satisfying the Markov chain $x_t — \hat{x}_{u_t} — u_t$. This suggests that the stochastic control process satisfies the Bayesian network in Figure 4, where the control is based on $\hat{x}_{u_t}$ instead of directly on $y_t$.

[1]Available at https://arxiv.org/abs/1606.01946

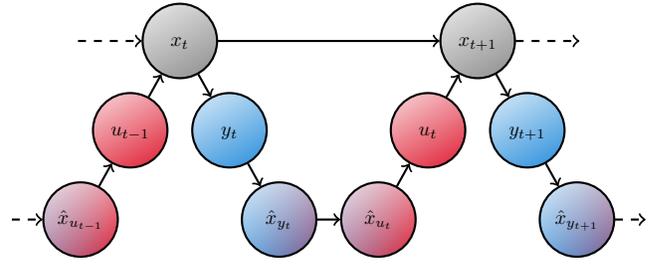

Fig. 4. Bayesian network of memoryless estimator-based control

*Lemma 2:* The bounded memoryless LTI controller optimization problem (Problem 1) is solved by a control law of the form

$$\hat{x}_{y_t} = K y_t \qquad (5a)$$
$$\hat{x}_{u_t} = W \hat{x}_{y_t} + \omega_t; \qquad \omega_t \sim \mathcal{N}(0, \Sigma_\omega) \qquad (5b)$$
$$u_t = L \hat{x}_{u_t}, \qquad (5c)$$

where $W \in \mathbb{R}^{n \times n}$, $\Sigma_\omega \in \mathbb{R}^{n \times n}$, $L \in \mathbb{R}^{\ell \times n}$, $\omega_t$ is independent of $y_t$, $\hat{x}_{u_t}$ is a MMSE estimator for $\hat{x}_{y_t}$ and

$$\mathbb{I}[y_t; u_t] = \mathbb{I}[\hat{x}_{y_t}; \hat{x}_{u_t}]. \qquad (6)$$

*Proof:* See SM, Appendix III. □

Lemma 2 allows us to derive optimality conditions for Problem 1. The stationary state covariance satisfies

$$\Sigma_x = \begin{bmatrix} A & B \end{bmatrix} \begin{bmatrix} \Sigma_x & \Sigma_{x;u} \\ \Sigma_{u;x} & \Sigma_u \end{bmatrix} \begin{bmatrix} A & B \end{bmatrix}^\mathsf{T} + \Sigma_\xi \qquad (7)$$
$$= (A + BL) \Sigma_{\hat{x}_u} (A + BL)^\mathsf{T} + A \Sigma_{x|\hat{x}_u} A^\mathsf{T} + \Sigma_\xi.$$

The mutual information between jointly Gaussian variables [37] is given by

$$\mathbb{I}[\hat{x}_{y_t}; \hat{x}_{u_t}] = \tfrac{1}{2}(\log |\Sigma_{\hat{x}_y}|_\dagger - \log |\Sigma_{\hat{x}_y | \hat{x}_u}|_\dagger), \qquad (8)$$

where $|\cdot|_\dagger$ is the pseudodeterminant, i.e. the product of the positive eigenvalues. This holds if $\Sigma_{\hat{x}_y}$ and $\Sigma_{\hat{x}_y | \hat{x}_u}$ have the same range and thus the same number of positive eigenvalues; otherwise, the mutual information between $y_t$ and $u_t$ is infinite.

With the target (8) and the constraints (7) and $\mathcal{J}_\pi \leq c$, where $\mathcal{J}_\pi$ is given by (2), the Lagrangian of Problem 1 can be written as

$$\mathcal{F}_{\Sigma_x, \Sigma_{\hat{x}_u}, L, S; \beta} = \tfrac{1}{2}(\beta^{-1}(\log |\Sigma_{\hat{x}_y}|_\dagger - \log |\Sigma_{\hat{x}_y | \hat{x}_u}|_\dagger) \qquad (9)$$
$$+ \operatorname{tr}(Q \Sigma_x) + \operatorname{tr}(RL \Sigma_{\hat{x}_u} L^\mathsf{T})$$
$$+ \operatorname{tr}(S((A+BL)\Sigma_{\hat{x}_u}(A+BL)^\mathsf{T}$$
$$+ A \Sigma_{x|\hat{x}_u} A^\mathsf{T} + \Sigma_\xi - \Sigma_x))).$$

Here $\beta > 0$ is the Lagrange multiplier corresponding to the constraint $\mathcal{J}_\pi \leq c$ and serving as the marginal trade-off coefficient between the external cost and the information rate, $\frac{\beta}{2} S \in \mathbb{R}^{n \times n}$ is the multiplier of the constraint (7) and for convenience the entire Lagrangian is divided by $\beta$. As in rate-distortion theory, $\mathcal{F}$ can be minimized for any given value of $\beta$. The $\beta$ that corresponds to a specific expected cost-rate guarantee level $c$ can then be found using a binary search. The case $\beta = 0$ corresponds to the minimization of information without any cost constraint.

*Theorem 1:* Given $\beta$, the Lagrangian (9) is minimized by a controller satisfying the forward equations

$$\Sigma_x = (A + BL)\Sigma_{\hat{x}_u}(A + BL)^\intercal \qquad (10a)$$
$$\quad + A\Sigma_{x|\hat{x}_u} A^\intercal + \Sigma_\xi$$
$$\Sigma_y = C\Sigma_x C^\intercal + \Sigma_\epsilon \qquad (10b)$$
$$K = \Sigma_x C^\intercal \Sigma_y^\dagger \qquad (10c)$$
$$\Sigma_{\hat{x}_y} = K\Sigma_y K^\intercal, \qquad (10d)$$

the backward equations

$$M = \beta^{-1} C^\intercal K^\intercal (\Sigma_{\hat{x}_y|\hat{x}_u}^\dagger - \Sigma_{\hat{x}_y}^\dagger) KC \qquad (10e)$$
$$S = Q + A^\intercal S A - M, \qquad (10f)$$
$$L = -(R + B^\intercal SB)^\dagger B^\intercal SA \qquad (10g)$$
$$N = L^\intercal (R + B^\intercal SB) L \qquad (10h)$$

and the control-based estimator covariance

$$\Sigma_{\hat{x}_u} = \Sigma_{\hat{x}_y}^{1/2} V D V^\intercal \Sigma_{\hat{x}_y}^{1/2}, \qquad (10i)$$

the latter determined by the eigenvalue decomposition (EVD)

$$V\Lambda V^\intercal = \Sigma_{\hat{x}_y}^{1/2} N \Sigma_{\hat{x}_y}^{1/2} \qquad (10j)$$

having $V$ orthogonal with $n - \mathrm{rank}(\Sigma_{\hat{x}_y})$ columns spanning the kernel of $\Sigma_{\hat{x}_y}$ and $\Lambda = \mathrm{diag}\{\lambda_i\}$ and by the active mode coefficient matrix

$$D = \mathrm{diag}\begin{cases} 1 - \beta^{-1}\lambda_i^{-1} & \lambda_i > \beta^{-1} \\ 0 & \lambda_i \leq \beta^{-1} \end{cases}. \qquad (10k)$$

*Proof:* See SM, Appendix IV. □

The spectral analysis in (10j)–(10k) implies that in (10e) the signal-to-noise-ratio (SNR) matrix $Z = \Sigma_{\hat{x}_y|\hat{x}_u}^\dagger - \Sigma_{\hat{x}_y}^\dagger$ satisfies

$$Z = \Sigma_{\hat{x}_y|\hat{x}_u}^\dagger - \Sigma_{\hat{x}_y}^\dagger = \Sigma_{\hat{x}_y}^{\dagger/2} V ((I - D)^{-1} - I) V^\intercal \Sigma_{\hat{x}_y}^{\dagger/2}$$
$$= \beta \Sigma_{\hat{x}_y}^{\dagger/2} V D \Lambda V^\intercal \Sigma_{\hat{x}_y}^{\dagger/2}$$

and that the information rate is

$$\mathcal{I}_\pi = \tfrac{1}{2}(\log|\Sigma_{\hat{x}_y}|_\dagger - \log|\Sigma_{\hat{x}_y|\hat{x}_u}|_\dagger) \qquad (11)$$
$$= -\log|I - D| = \sum_i \max(0, \log \beta \lambda_i).$$

As shown in the SM, Appendix I, given an additive Gaussian noise channel $w_t \to \hat{w}_t$ with noise covariance $I - D$, the optimal encoder and decoder are now given by

$$w_t = D^{1/2} V^\intercal \Sigma_{\hat{x}_y}^{\dagger/2} \hat{x}_{y_t}$$
$$\hat{x}_{u_t} = \Sigma_{\hat{x}_y}^{1/2} V D^{1/2} \hat{w}_t,$$

which can be summarized in the form (5b), with

$$W = \Sigma_{\hat{x}_u} \Sigma_{\hat{x}_y}^\dagger$$
$$\Sigma_\omega = \Sigma_{\hat{x}_y}^{1/2} V D(I - D) V^\intercal \Sigma_{\hat{x}_y}^{1/2}.$$

Alternatively, the controller can be given in the form (3), with

$$H = LWK$$
$$\Sigma_\eta = L\Sigma_\omega L^\intercal.$$

Interestingly, Theorem 1 also shows that $S$ corresponds to the cost-to-go Hessian, with respect to the state, as in classic control theory. The difference is that here $S$ also accumulates the non-quadratic information cost and is only the Hessian in an average sense. In the form given in Theorem 1, $M$ is positive semidefinite, but $S$ may not be. This is not problematic if we view $S$ as the Lagrange multiplier of the equality constraint (7), but it is undesired for a cost-to-go Hessian. The positive semidefiniteness of $S$ is discussed further and restored in Part II [15, Section III-C].

Theorem 1 gives the first-order necessary conditions for a solution to be optimal; namely, that the gradient of the Lagrangian (9) is 0 with respect to each parameter. It additionally includes two more conditions, one which is higher-order and the other non-necessary. First, the condition on $\Sigma_{\hat{x}_u}$ (10i) is necessary but not first-order, being a solution to a semidefinite program (see SM, Appendix V). Second, the condition on $L$ (10g) is the least-square solution of a possibly underdetermined system, which means that it may not hold for all optimal solutions but that it does hold for some globally optimal solution.

Problem 1 is highly non-convex and has many local optima that satisfy the first-order necessary conditions. By including the two higher-order and non-necessary conditions, we exclude many of these local optima, although some remain (see Part II [15, Section IV]). This merits further study of the fixed-point structure of this problem.

### B. Phenomenology

To better understand the optimal solution of Theorem 1, consider its phenomenology as $\beta$ spans its range from 0 to $\infty$. The following is the SRD extension of a standard result in one-shot RD theory [37].

*Lemma 3:* Let $\mathcal{I}(\mathcal{J})$ be the minimal information rate achievable by a controller that incurs cost at rate at most $\mathcal{J}$. This information-cost function is monotonically decreasing, convex, and its slope is

$$\partial_\mathcal{J} \mathcal{I} = -\beta, \qquad (12)$$

for $\beta$ the Lagrange multiplier corresponding to the expected cost-rate guarantee level $c = \mathcal{J}$.

*Proof:* For any $\beta$, let

$$\pi^* = \arg\min_\pi \{\beta^{-1} \mathcal{I}_\pi + \mathcal{J}_\pi\}.$$

$\pi^*$ achieves the optimum in Problem 1 when $c = \mathcal{J}_{\pi^*}$. Take

$$\mathcal{I} = \mathcal{I}_{\pi^*}; \qquad \mathcal{J} = \mathcal{J}_{\pi^*}; \qquad \mathcal{F} = \beta^{-1}\mathcal{I} + \mathcal{J}.$$

Then the slope equation follows by fixing $\beta$ while $\mathcal{J}$ and $\mathcal{I}$ vary and noting that at the optimum

$$\partial_\mathcal{J} \mathcal{F} = \beta^{-1} \partial_\mathcal{J} \mathcal{I} + 1 = 0.$$

Monotonicity follows directly from the definition of Problem 1. Convexity can also be shown directly; however, it follows more easily from the slope equation (12) by considering that $\mathcal{J}$ is non-increasing in $\beta$ and thus

$$\partial^2_{\mathcal{J}^2} \mathcal{I} = -\partial_{\mathcal{J}} \beta \geq 0. \qquad \square$$

We now turn to consider how the controller order is increased as $\beta$ is increased from 0 to $\infty$. This phenomenon is known as a water-filling effect [36], [37], and is made explicit in the form of the optimal information rate $\mathcal{I}_\pi$ (11). Note, however, that in the SRD problem the water-filling effect is self-consistent, in that $\Lambda$ itself depends on $\beta$.

*Definition 5:* The order of a LTI controller is $\mathrm{rank}(\Sigma_{\hat{x}_u})$. For the optimal solution (10i), this equals $\mathrm{rank}(D)$, the number of active modes.

Let us consider a stable plant, having all eigenvalues of $A$ inside the unit circle. We note that our results hold more generally and extend known results [18] when the plant is unstable but stabilizable and detectable. However, the analysis of this case when $\beta \to 0$ is more involved and is presented separately in an upcoming paper.

When $\beta = 0$, we are only interested in minimizing $\mathcal{I}_\pi$ and therefore take an order-0 controller, having $D = 0$, $\Sigma_{\hat{x}_u} = 0$ and $M = 0$. $\Sigma_x$ and $S$ satisfy the uncontrolled Lyapunov equations

$$\Sigma_x = A \Sigma_x A + \Sigma_\xi$$
$$S = Q + A^\intercal S A.$$

$L$ and $N$ can be set accordingly, despite the fact that no attention to the observation is spent and no control is possible. Computing the EVD of $\Sigma_{\hat{x}_y}$ and applying (10j), we can retrieve $\Lambda$.

As we increase $\beta$, this uncontrolled solution remains constant as long as $\beta \leq \lambda_1^{-1}$, the inverse of the largest eigenvalue in $\Lambda$. At that first critical point, the controller undergoes a phase transition, where its order increases from 0 to 1 (or higher if $\lambda_1$ is not unique in $\Lambda$).

Note that $\Lambda$ contains the same eigenvalues as the matrix

$$\Sigma_{N^{1/2}\hat{x}_y} = N^{1/2} \Sigma_{\hat{x}_y} N^{1/2},$$

which represents the value of the information that the observation has on the state, in terms of the cost reduction it allows. Thus an order-1 controller observes and controls the state mode that provides the largest decrease in cost per bit of observed information, in keeping with (12).

Beyond the first phase transition, the optimal solution does change with $\beta$ and so does $\Lambda$. Eventually, $\beta$ meets $\lambda_i^{-1}(\beta)$, for each $i = 2, \ldots, \mathrm{rank}(\Sigma_{N^{1/2}\hat{x}_y})$ in turn and further phase transitions occur, increasing the controller order until it reaches $\mathrm{rank}(\Sigma_{N^{1/2}\hat{x}_y})$.

As long as $\beta$ is finite, even after the last phase transition, the information rate must be finite. Since the controller lacks the capacity to attend to any mode with perfect fidelity, it must maintain some uncertainty in all modes and accordingly $D \prec I$ and $\Sigma_{\hat{x}_u} \prec \Sigma_{\hat{x}_y}$. As $\beta \to \infty$, the SNR matrix $Z = \Sigma^\dagger_{\hat{x}_y|\hat{x}_u} - \Sigma^\dagger_{\hat{x}_y}$ grows to infinity in modes having $\lambda_i > 0$, as does the information rate in these modes.

The $\beta = \infty$ case marks a qualitative change in the optimization problem. We are no longer concerned with the information rate and only wish to minimize the expected cost rate $\mathcal{J}_\pi$. The optimal solution here is underdetermined with respect to useless modes where $\lambda_i = 0$. Despite having no value in decreasing $\mathcal{J}_\pi$, at $\beta = \infty$ (but not for $\beta \to \infty$) these modes may be observed for free. This allows us to simplify the solution to

$$D = I$$
$$\Sigma_{\hat{x}_u} = \Sigma_{\hat{x}_y}$$
$$M = C^\intercal K^\intercal \Sigma^{\dagger/2}_{\hat{x}_y} \Sigma^{1/2}_{\hat{x}_y} N \Sigma^{1/2}_{\hat{x}_y} \Sigma^{\dagger/2}_{\hat{x}_y} K C$$
$$= C^\intercal K^\intercal N K C.$$

It is interesting to note the impact of the observability on $M$ at $\beta = \infty$. When the plant is unobservable, we have $C = K = 0$ and thus $M = 0$. When observability is full, we have $C = K = I$ and thus $M = N$. For partial observability models, $N - M$ is not necessarily positive semidefinite, which will become important in the reduced retentive control problem (see Part II [15, Section III-C]).

In the classic control problem, where observability is partial but the memory and the sensory capacities are unbounded, the memory state is maintained by the Kalman filter and we have $M = N$ and

$$S = Q + A^\intercal S A - N,$$

independent of the forward inference process. Note, however, that $S$ in that case is the Hessian of the certainty-equivalent cost-to-go with respect to $\hat{x}_t$, instead of $x_t$.

Thus either full and unbounded ($\beta = \infty$) observability or bounded ($\beta < \infty$) sensing with unbounded memory [31] are sufficient for recovering the separation principle of classic control theory. In the more general case, the backward control process (10f) is coupled with the forward inference process (10a).

## V. DISCUSSION

In this paper we introduce the problem of optimal memoryless LQG control with bounded channel capacity. We present the solution and discuss some of its properties and phenomenology.

Part of our motivation in considering memoryless controllers is that the problem of retentive (memory-utilizing) control can be reduced to the problem of memoryless control. This is discussed in detail in Part II of this work [15, Section III-B]. The two control models are also compared there (Section IV) using an illustrative example.

One attractive aspect of our solution is its principled reduction of the controller order. In many applications, the controller's information rate is a more natural measure of its complexity than the dimension of its support. Nevertheless, a hard constraint on the order is sometimes required, alongside a soft constraint on the information rate, leading to an algorithmically challenging open question.

The controllers considered in this paper have linear-Gaussian control laws. This class of controllers does not solve optimally all control problems and is particularly prone to suboptimality in memory-constrained settings [38], [39]. Nevertheless, we conjecture that there exist some moderately strong conditions under which the bounded memoryless control problem discussed here is solved optimally by an LTI controller.

## 3.2 Part II: Retentive Controllers

# Minimum-Information LQG Control
# Part II: Retentive Controllers

Roy Fox† and Naftali Tishby†

*Abstract*— Retentive (memory-utilizing) sensing-acting agents may operate under limitations on the communication between their sensing, memory and acting components, requiring them to trade off the external cost that they incur with the capacity of their communication channels. In this paper we formulate this problem as a sequential rate-distortion problem of minimizing the rate of information required for the controller's operation under a constraint on its external cost. We reduce this bounded retentive control problem to the memoryless one, studied in Part I of this work [1], by viewing the memory reader as one more sensor and the memory writer as one more actuator. We further investigate the structure of the resulting optimal solution and demonstrate its interesting phenomenology.

## I. INTRODUCTION

In a feedback-control system, the internal state of the agent interacts with the external state of the world through sensors that pay attention to the agent's environment and actuators that apply intention to it, in a perception-action cycle [2]. This interaction is limited by external constraints on observability and controllability, as well as internal constraints on the information-processing resources available to the controller.

In Part I of this work [1], we focused on memoryless controllers that have no internal memory and can only attend to their most recent input observation. We discussed how the communication from the sensor to the actuator is central to the agent's ability to act upon the perceived information. The degree of this attention, measured by the amount of Shannon information about the input observation that is utilized in the output control, is a lower bound on the required capacity of the communication channel between the controller's sensor and its actuator. When this capacity for internal communication is limited, the agent needs to trade off some external cost for reducing the rate at which it transmits information.

A related but often overlooked resource is memory bandwidth. We can think of memory as a communication channel from the past internal state of the controller to its future internal state. When memory resources are remote, communication constraints apply to them as well. Even local memory is limited by its capacity to store information and by the capacity of the internal communication channels to and from the memory components. This limitation is evidenced by the hierarchical design of memory in modern digital computers, which places larger capacity on the channels to closer but smaller cache memory components [3].

When the controller is retentive (memory-utilizing), it does maintain an internal memory state which can have information on more than the most recent observation. As in Part I, our guiding principle in this work is to measure the information complexity of the controller's internal representation by asking *"How much information does the controller have on the past?"*. The retentive controller receives information of the past through both memory and sensory channels (Figure 2) and the amount of information that it keeps of the past is a lower bound on the total capacity of both these channels [4].

In a sense, we can consider the reader of the memory state to be one more sensor and the writer of the memory state to be one more actuator. This suggests a reduction from the retentive case to the memoryless case, in which the memory state is considered external and part of the world state [5], [6]. This memory component is fully observable, fully controllable, has no process noise and incurs no cost. Rather than redevelop our results for the retentive controllers similarly to Part I, this reduction allows us to reuse those results and underlines the structure of the solution.

In this paper we make two contributions. First, we present a method for the design of controllers that are optimal under a constraint on both their memory and sensory channel capacity. To our knowledge, this is the first explicit treatment of the channel capacity of the memory process in the context of continuous state-space systems.

Second, we provide a reduction from the problem of bounded retentive control to the problem of bounded memoryless control. This reduction is conceptually convenient and constructive, allowing us to treat both problems using the same framework and providing insight into the structure of the optimal retentive controller.

In Section II we define the LQG task and restate the results of Part I. In Section III we present the retentive control model, its reduction to memoryless control and the structure of the resulting optimal solution. In Section IV we illustrate our results with an example.

## II. PRELIMINARIES

### A. Control task

We consider the same closed-loop control problem detailed in Part I [1, Section II]. In time $t$, a plant in state $x_t \in \mathbb{R}^n$ emits an observation $y_t \in \mathbb{R}^k$, takes in a control input $u_t \in \mathbb{R}^\ell$ and undergoes a stochastic state transition. We focus on discrete-time systems with linear dynamics,

†School of Computer Science and Engineering, The Hebrew University, {royf,tishby}@cs.huji.ac.il
*This work was supported by the DARPA MSEE Program, the Gatsby Charitable Foundation, the Israel Science Foundation and the Intel ICRI-CI Institute

Gaussian noise and quadratic cost rate (LQG). For simplicity, all elements are taken to be homogeneous, i.e. centered at the origin, and time-invariant. We note that all our results hold without these assumptions, with the appropriate adjustments, as usual in LQG problems [7].

*Definition 1:* A linear-Gaussian time-invariant (LTI) plant $\langle A, B, C, \Sigma_\xi, \Sigma_\epsilon \rangle$ has state dynamics

$$x_{t+1} = Ax_t + Bu_t + \xi_t; \qquad \xi_t \sim \mathcal{N}(0, \Sigma_\xi), \quad (1)$$

where $A \in \mathbb{R}^{n \times n}$, $B \in \mathbb{R}^{n \times \ell}$, $0 \preceq \Sigma_\xi \in \mathbb{S}_+^n$ and $\xi_t$ is independent of $(x^t, y^t, u^t)$. The observation dynamics are

$$y_t = Cx_t + \epsilon_t; \qquad \epsilon_t \sim \mathcal{N}(0, \Sigma_\epsilon), \quad (2)$$

where $C \in \mathbb{R}^{k \times n}$, $\Sigma_\epsilon \in \mathbb{S}_+^k$ and $\epsilon_t$ is independent of $(y^{t-1}, u^{t-1}, x^t)$, where we denote $x^t = \{x_\tau\}_{\tau \le t}$, etc.

*Definition 2:* A linear-quadratic-Gaussian (LQG) task $\langle A, B, C, \Sigma_\xi, \Sigma_\epsilon, Q, R \rangle$ involves a LTI plant and the cost rate

$$\mathcal{J}_t = \tfrac{1}{2}(x_t^\intercal Q x_t + u_t^\intercal R u_t),$$

where $Q \in \mathbb{S}_+^n$ and $R \in \mathbb{S}_+^\ell$. The task is to achieve a low long-term average expected cost rate, with respect to the distribution induced by the plant and the controller $\pi$

$$\mathcal{J}_\pi = \limsup_{T \to \infty} \frac{1}{T} \sum_{t=1}^T \mathbb{E}_\pi[\mathcal{J}_t].$$

As motivated in Part I, we are particularly interested in linear-Gaussian time-invariant (LTI) controllers, which induce, jointly with a LTI plant, a stationary Gaussian process, independent of any initial conditions. With $\Sigma_x \in \mathbb{S}_+^n$, $\Sigma_y \in \mathbb{S}_+^k$ and $\Sigma_u \in \mathbb{S}_+^\ell$, respectively the stationary covariances of the state, the observation and the control, we have

$$\Sigma_y = C \Sigma_x C^\intercal + \Sigma_\epsilon,$$

and the reverse relation

$$x_t = K y_t + \kappa_t; \qquad \kappa_t \sim \mathcal{N}(0, \Sigma_\kappa)$$
$$K = \Sigma_x C^\intercal \Sigma_y^\dagger$$
$$\Sigma_\kappa = \Sigma_x - \Sigma_x C^\intercal \Sigma_y^\dagger C \Sigma_x,$$

with $\cdot^\dagger$ the Moore-Penrose pseudoinverse. Assuming that the process has mean 0, the stationary expected cost rate is

$$\mathcal{J}_\pi = \tfrac{1}{2}(\mathrm{tr}(Q \Sigma_x) + \mathrm{tr}(R \Sigma_u)).$$

*B. Bounded memoryless control*

In this section we restate the main result of Part I [1, Section IV].

*Definition 3:* A memoryless linear-Gaussian time-invariant (LTI) controller has control law of the form

$$u_t = H y_t + \eta_t; \qquad \eta_t \sim \mathcal{N}(0, \Sigma_\eta), \quad (3)$$

where $H \in \mathbb{R}^{\ell \times k}$, $\Sigma_\eta \in \mathbb{S}_+^\ell$ and $\eta_t$ is independent of $(u^{t-1}, x^t, y^t)$.

The controller is bounded and operates under limitations on its capacity to process the observation and produce the control. Namely, with the Shannon information rate

$$\mathcal{I}_t = \mathbb{I}[y_t; u_t] = \mathbb{E}\left[\log \frac{f(y_t, u_t)}{f(y_t) f(u_t)}\right], \quad (4)$$

where $f$ denotes the various probability density functions, as indicated by their arguments, we are interested in a LTI controller $\pi$ that minimizes the long-term average rate

$$\mathcal{I}_\pi = \limsup_{T \to \infty} \frac{1}{T} \sum_{t=1}^T \mathcal{I}_t, \quad (5)$$

under the constraint that it achieves some guarantee level $c$ of expected cost rate.

*Problem 1:* Given a LQG task, the bounded memoryless LTI controller optimization problem is

$$\min_\pi \quad \mathcal{I}_\pi$$
$$\text{s.t.} \quad \mathcal{J}_\pi \le c,$$

with $\mathcal{I}_\pi$ as in (5), where $\mathcal{I}_t = \mathbb{I}[y_t; u_t]$, and with $u_t$ as in (3).

To solve the optimization problem, we consider the minimum mean square error (MMSE) estimators

$$\hat{x}_{y_t} = \mathbb{E}[x_t | y_t] = K y_t$$
$$\hat{x}_{u_t} = \mathbb{E}[x_t | u_t] = \Sigma_{x;u} \Sigma_u^\dagger u_t,$$

respectively for the state given the observation and the control. Since $\hat{x}_{u_t}$ is a sufficient statistic of $u_t$ for $x_t$, we can reverse their causality, basing $u_t$ on $\hat{x}_{u_t}$ instead of vice versa. This puts the control law in the form

$$\hat{x}_{y_t} = K y_t$$
$$\hat{x}_{u_t} = W \hat{x}_{y_t} + \omega_t; \qquad \omega_t \sim \mathcal{N}(0, \Sigma_\omega)$$
$$u_t = L \hat{x}_{u_t}.$$

The optimal memoryless controller satisfies the conditions of Theorem 1 in Part I, Section IV-A, restated below in algorithmic form. To numerically find the optimal solution, we can interpret these conditions as update equations, which we apply iteratively until a fixed point is reached.

We split the equations into three parts, a forward iteration (Algorithm 1) updating the marginal distributions, a backward iteration (Algorithm 2) updating the cost-to-go and the control policy, and an eigenvalue decomposition (EVD) for finding the control-based estimator covariance (Algorithm 3). We can alternate between Algorithms 1, 2 and 3, iterating until the solution converges to a fixed point of the equations.

### III. BOUNDED RETENTIVE CONTROLLERS

*A. Control model*

In this section we discuss retentive (memory-utilizing) controllers with bounded communication resources. A retentive controller has an internal memory state $z_t$ in some space $\mathcal{Z}$. The memory allows the controller to output a control that indirectly depends on past input observations rather than only on the most recent observation. The controller takes as input an observation $y_t$ and outputs a control $u_t$, while also making a memory state transition from $z_{t-1}$ to $z_t$. Thus, in each time step, there are two inputs, $z_{t-1}$ and $y_t$, and two outputs, $z_t$ and $u_t$.

*Definition 4:* A controller is retentive if it satisfies the following independence properties:

**Algorithm 1** Forward iteration
**function** FORWARD($\Sigma_x, \Sigma_{\hat{x}_u}, L$)
    Update
$$\Sigma_x \leftarrow (A + BL)\,\Sigma_{\hat{x}_u}(A + BL)^\mathsf{T}$$
$$\qquad + A(\Sigma_x - \Sigma_{\hat{x}_u})A^\mathsf{T} + \Sigma_\xi$$
$$\Sigma_y \leftarrow C\,\Sigma_x\,C^\mathsf{T} + \Sigma_\epsilon$$
$$K \leftarrow \Sigma_x\,C^\mathsf{T}\,\Sigma_y^\dagger$$
$$\Sigma_{\hat{x}_y} \leftarrow K\,\Sigma_y\,K^\mathsf{T}$$
**end function**

**Algorithm 2** Backward iteration
**function** BACKWARD($\Sigma_{\hat{x}_y}, \Sigma_{\hat{x}_u}, K, S; \beta$)
    Update
$$M \leftarrow \beta^{-1}C^\mathsf{T}K^\mathsf{T}(\Sigma_{\hat{x}_y|\hat{x}_u}^\dagger - \Sigma_{\hat{x}_y}^\dagger)KC$$
$$S \leftarrow Q + A^\mathsf{T}SA - M$$
$$L \leftarrow -(R + B^\mathsf{T}SB)^\dagger B^\mathsf{T}SA$$
$$N \leftarrow L^\mathsf{T}(R + B^\mathsf{T}SB)L$$
**end function**

**Algorithm 3** Activation of control-based estimator modes
**function** ACTIVATION($\Sigma_{\hat{x}_y}, N; \beta$)
    Update
$$V, \Lambda \leftarrow \mathrm{EVD}(\Sigma_{\hat{x}_y}^{1/2}\,N\,\Sigma_{\hat{x}_y}^{1/2})$$
with $n - \mathrm{rank}(\Sigma_{\hat{x}_y})$ columns of $V$ spanning $\ker(\Sigma_{\hat{x}_y})$
$$D \leftarrow \mathrm{diag}\left\{\begin{array}{ll} 1 - \beta^{-1}\lambda_i^{-1} & \lambda_i > \beta^{-1} \\ 0 & \lambda_i \leq \beta^{-1} \end{array}\right\}$$
$$\Sigma_{\hat{x}_u} \leftarrow \Sigma_{\hat{x}_y}^{1/2}\,VDV^\mathsf{T}\,\Sigma_{\hat{x}_y}^{1/2}$$
**end function**

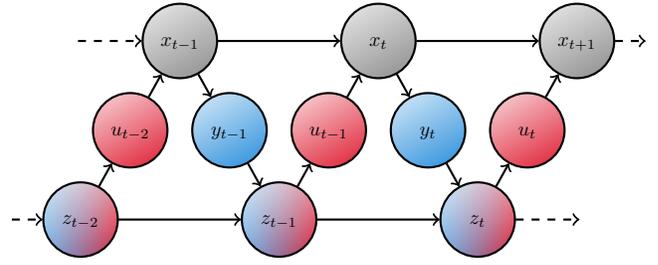

Fig. 1. Bayesian network of retentive control

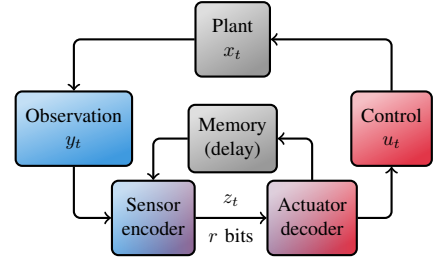

Fig. 2. Block diagram of a closed-loop retentive control system, with a communication channel from the sensor-reader to the actuator-writer

1) The memory state depends only on the previous memory state and the current observation; that is, $z_t$ is independent of $(z^{t-2}, y^{t-1}, u^{t-1}, x^t)$ given $z_{t-1}$ and $y_t$.
2) The control depends only on the memory state; that is, $u_t$ is independent of $(z^{t-1}, u^{t-1}, x^t, y^t)$ given $z_t$.

A system including a retentive controller satisfies the Bayesian network in Figure 1.

As motivated in Part I for the memoryless case, we are particularly interested in controllers where both the memory state update and the control are linear-Gaussian and time-invariant (LTI), since they are easier to optimize and implement. Linear controllers with limited memory are known not to be optimal for all control problems [8], [9]. The conditions under which such controllers are optimal for our bounded control problem are beyond our current scope.

*Definition 5:* A retentive linear-Gaussian time-invariant (LTI) controller has memory state space that is a vector space $\mathcal{Z} = \mathbb{R}^d$ and control law of the form

$$z_t = Fz_{t-1} + Gy_t + \zeta_t; \qquad \zeta_t \sim \mathcal{N}(0, \Sigma_\zeta) \quad (6\text{a})$$
$$u_t = Lz_t + \nu_t; \qquad \nu_t \sim \mathcal{N}(0, \Sigma_\nu) \quad (6\text{b})$$

where $F \in \mathbb{R}^{d \times d}$, $G \in \mathbb{R}^{d \times k}$, $\Sigma_\zeta \in \mathbb{S}_+^d$, $L \in \mathbb{R}^{m \times d}$, $\Sigma_\nu \in \mathbb{S}_+^m$, $\zeta_t$ is independent of $(z_{t-1}, y_t)$ and $\nu_t$ is independent of $z_t$.

We are interested in reducing the information complexity of implementing this controller. To measure this complexity, we consider the capacity of a memoryless communication channel from the sensor-reader to the actuator-writer (Figure 2). The encoder and the decoder themselves are memoryless, but the memory component has perfect fidelity, making everything written by the actuator available for the sensor to read in the next step.

We could use $\mathcal{Z} = \{0, 1\}^r$, the set of $r$-bit strings, instead of the vector space $\mathbb{R}^d$, to indicate that the controller can process at most $r$ bits of information per time step

$$\mathbb{I}[z_{t-1}, y_t; z_t, u_t] = \mathbb{I}[z_{t-1}, y_t; z_t] \leq \mathbb{H}[z_t] \leq r \log 2.$$

As in the memoryless case (Part I [1, Section III-B]), the information rate is generally not a tight lower bound on the capacity of a discrete memory, but here again, if the controller is LTI, there exists a perfectly matched memoryless additive Gaussian noise channel. As shown in the Supplementary Material[1] (SM), Appendix I, the capacity of this channel optimally equals the information rate $\mathbb{I}[z_{t-1}, y_t; z_t, u_t]$ and a constraint on the information rate is equivalent to a constraint on the power available for transmission on the channel.

The retentive controller optimization problem is therefore similar to Problem 1, but with the information rate including both the memory and the sensory channels.

---
[1]Available at https://arxiv.org/abs/1606.01947

*Problem 2:* Given a LQG task, the bounded retentive LTI controller optimization problem is

$$\min_{\pi} \quad \mathcal{I}_\pi$$
$$\text{s.t.} \quad \mathcal{J}_\pi \leq c,$$

with $\mathcal{I}_\pi$ as in (5), where

$$\mathcal{I}_t = \mathbb{I}[z_{t-1}, y_t; z_t, u_t], \quad (7)$$

and with $z_t$ and $u_t$ as in (6).

Note that here there is no additional constraint or cost on the precision of $u_t$ given $z_t$, implying that optimally $\Sigma_\nu = 0$.

There is an interesting connection between the retentive information rate $\mathcal{I}_\pi$ and the long-term average of the directed information rate [10], [11], defined by

$$\mathbb{I}[\{y_t\} \to \{z_t\}] = \limsup_{T \to \infty} \frac{1}{T} \mathbb{I}[y^T \to z^T]$$
$$= \limsup_{T \to \infty} \frac{1}{T} \sum_{t=1}^{T} \mathbb{I}[y^t; z_t | z^{t-1}].$$

By the independence properties of the retentive controller and by the chain rule for information [12], we have

$$\mathbb{I}[z_{t-1}, y_t; z_t, u_t] = \mathbb{I}[z_{t-1}, y_t; z_t]$$
$$= \mathbb{I}[z^{t-1}, y^t; z_t]$$
$$= \mathbb{I}[z^{t-1}; z_t] + \mathbb{I}[y^t; z_t | z^{t-1}].$$

We can thus define the following extension of the concept of directed information.

*Definition 6:* The retentive directed information from the sequence of observations $y^T$ to the sequence of memory states $z^T$ is

$$\mathbb{I}[y^T \twoheadrightarrow z^T] = \sum_{t=1}^{T} \mathbb{I}[z^{t-1}, y^t; z_t].$$

Since $\mathbb{I}[y^T \twoheadrightarrow z^T] \geq \mathbb{I}[y^T \to z^T]$, the retentive directed information rate is always a tighter lower bound on the capacity of the channel in Figure 2. Despite the apparent similarity to Figure 2 in [11], notice that their encoder and decoder have unlimited memory of $z^t$ and $u^t$. This justifies their use of directed information, regardless of the residual term $\mathbb{I}[z^{t-1}; z_t]$ being infinite in their optimal controller.

Some further properties of the retentive directed information can be found in the SM, Appendix VI.

### B. Reduction to memoryless controllers

We can analyze the bounded retentive control problem (Problem 2) directly using the same tools developed in Part I [1, Section IV-A] for Problem 1. Fortunately, there is no need to repeat that entire treatment, since a simple and insightful reduction will allow us to reuse the results obtained there.

We start by reformulating the problem. The following relaxation and Lemma 1 that shows its equivalence to the original problem allow us to reverse the causality between $u_t$ and $z_t$. We need a new notation for the resulting time-shifted memory state sequence and define for each $t$

$$m_t = z_{t-1}.$$

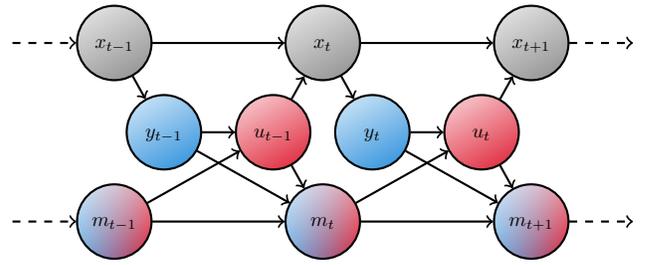

Fig. 3. Bayesian network of relaxed retentive control

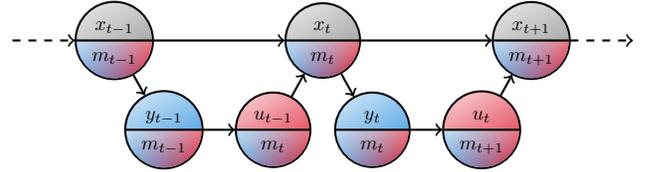

Fig. 4. Bayesian network of relaxed retentive control, redrawn in the form of memoryless control

*Definition 7:* A retentive controller is relaxed if $u_t$ is not required to be independent of $(m_t, y_t)$ given $m_{t+1}$. Thus the relaxed controller satisfies the Bayesian network in Figure 3 and its control law is given by $\pi(u_t, m_{t+1} | m_t, y_t)$.

*Lemma 1:* The relaxed controller optimization problem is equivalent to the original Problem 2.

*Proof:* The following proof does not assume that the controller is linear-Gaussian and holds for the LTI controller as a special case.

Let $\pi$ be a controller satisfying the Bayesian network in Figure 3. We construct a controller $\tilde{\pi}$ with $\tilde{z}_t = (u_t, m_{t+1})$ for each $t$, such that

$$\tilde{\pi}(\tilde{z}_t | \tilde{z}_{t-1}, y_t) = \pi(u_t, m_{t+1} | m_t, y_t)$$
$$\tilde{\pi}(u_t | \tilde{z}_t) = \delta_{\tilde{z}_t = (u_t, \cdot)}.$$

This controller satisfies the Bayesian network in Figure 1 and

$$\mathbb{I}_{\tilde{\pi}}[\tilde{z}_{t-1}, y_t; \tilde{z}_t, u_t] = \mathbb{I}_\pi[(u_{t-1}, m_t), y_t; (u_t, m_{t+1})]$$
$$= \mathbb{I}_\pi[m_t, y_t; u_t, m_{t+1}].$$

Thus the controller $\tilde{\pi}$ is feasible for the unrelaxed Problem 2 and has the same performance as the relaxed controller $\pi$, since it induces a stochastic process with the same distribution and information rate. □

The structure in Figure 3 can now be redrawn as in Figure 4. Comparing this Bayesian network to the one in Part I, Figure 2, we have clearly reduced the bounded retentive control problem to a special case of the bounded memoryless control problem, as stated formally in the following lemma.

*Lemma 2:* The bounded retentive LTI controller optimization problem (Problem 2) for the LQG task $\langle A_x, B_{x;u}, C_{y;x}, \Sigma_\xi, \Sigma_\epsilon, Q_x, R_u \rangle$ is equivalent to the bounded memoryless LTI controller optimization problem (Problem 1) for the LQG task $\langle A, B, C, \Sigma_{\tilde{\xi}}, \Sigma_{\tilde{\epsilon}}, Q, R \rangle$, where

$$A = \begin{bmatrix} A_x & 0 \\ 0 & 0 \end{bmatrix}; \quad B = \begin{bmatrix} B_{x;u} & 0 \\ 0 & I \end{bmatrix}; \quad C = \begin{bmatrix} C_{y;x} & 0 \\ 0 & I \end{bmatrix}$$

$$\Sigma_{\tilde{\xi}} = \begin{bmatrix} \Sigma_\xi & 0 \\ 0 & 0 \end{bmatrix}; \quad \Sigma_{\tilde{\epsilon}} = \begin{bmatrix} \Sigma_\epsilon & 0 \\ 0 & 0 \end{bmatrix}$$

$$Q = \begin{bmatrix} Q_x & 0 \\ 0 & 0 \end{bmatrix}; \quad R = \begin{bmatrix} R_u & 0 \\ 0 & 0 \end{bmatrix}.$$

Here all matrices are extended by $d$ rows and $d$ columns.

*Proof:* Given the retentive control stochastic process $\{x_t, m_t, y_t, u_t\}$, we consider the memoryless control stochastic process $\{\tilde{x}_t, \tilde{y}_t, \tilde{u}_t\}$ with

$$\tilde{x}_t = \begin{bmatrix} x_t \\ m_t \end{bmatrix}; \quad \tilde{y}_t = \begin{bmatrix} y_t \\ m_t \end{bmatrix}; \quad \tilde{u}_t = \begin{bmatrix} u_t \\ m_{t+1} \end{bmatrix}.$$

The dynamics for this process can easily be seen to be given by (1), (2), with $A$, $B$, $C$, $\Sigma_{\tilde{\xi}}$ and $\Sigma_{\tilde{\epsilon}}$ as in the lemma. The cost rate applies only to the $x_t$ and $u_t$ parts

$$\mathcal{J}_t = \tfrac{1}{2} \left( \begin{bmatrix} x_t \\ m_t \end{bmatrix}^\intercal \begin{bmatrix} Q_x & 0 \\ 0 & 0 \end{bmatrix} \begin{bmatrix} x_t \\ m_t \end{bmatrix} + \begin{bmatrix} u_t \\ m_{t+1} \end{bmatrix}^\intercal \begin{bmatrix} R_u & 0 \\ 0 & 0 \end{bmatrix} \begin{bmatrix} u_t \\ m_{t+1} \end{bmatrix} \right).$$

The information rate is

$$\mathcal{I}_t = \mathbb{I}[\tilde{y}_t; \tilde{u}_t] = \mathbb{I}[m_t, y_t; u_t, m_{t+1}],$$

where the left-hand side is taken as in (4) and the right-hand side as in (7), as required. □

### C. Structure of the optimal solution

We can substitute the form of the reduction in Lemma 2 into the optimal solution in Section II-B, to study more explicitly the structure of the optimal solution in the retentive case. The detailed derivations can be found in the SM, Appendix VII.

For the backward process, it is useful to borrow notation from the forward process and denote

$$S = \begin{bmatrix} S_x & S_{x;m} \\ S_{m;x} & S_m \end{bmatrix}$$

$$S_{x|m} = S_x - S_{x;m} S_m^\dagger S_{m;x}$$

$$S_{u|m} = R + B^\intercal S_{x|m} B.$$

Then we can find the feedback gain

$$L = -(R + B^\intercal S B)^\dagger B^\intercal S A$$
$$= \begin{bmatrix} L_{u;x|m} & 0 \\ -S_m^\dagger S_{m;x}(A_x + B_{x;u} L_{u;x|m}) & 0 \end{bmatrix}, \quad (8)$$

with a memory-conditioned form of the classic feedback gain

$$L_{u;x|m} = -S_{u|m}^\dagger B_{x;u}^\intercal S_{x|m} A_x.$$

The memory-conditioned cost reduction matrix is

$$N = L^\intercal (R + B^\intercal S B) L = \begin{bmatrix} N_{x|m} & 0 \\ 0 & 0 \end{bmatrix},$$

with

$$N_{x|m} = A_x^\intercal (S_x - S_{x|m} + S_{x|m} B_{x;u} S_{u|m}^\dagger B_{x;u}^\intercal S_{x|m}) A_x.$$

Thus $\operatorname{rank}(D) \le \operatorname{rank}(N) \le n$, with $D$ the mode activation matrix (see Algorithm 3), implying that at most $n$ modes can be active.

The $d$ rightmost columns in (8) are 0, implying that $\tilde{u}_t$ depends only on the state estimator $\hat{x}_{\tilde{u}_t} = \mathbb{E}[x_t | \tilde{u}_t]$ of $x_t$ and not on an estimator of the memory component $m_t$. Since $\hat{x}_{\tilde{y}_t} = \mathbb{E}[x_t | \tilde{y}_t]$ is a sufficient statistic of $\tilde{y}_t$ for $x_t$, we also have the Markov chain

$$x_t \;\text{—}\; \hat{x}_{\tilde{y}_t} \;\text{—}\; \tilde{y}_t \;\text{—}\; \hat{\tilde{x}}_{\tilde{y}_t} \;\text{—}\; \hat{\tilde{x}}_{\tilde{u}_t} \;\text{—}\; \hat{x}_{\tilde{u}_t} \;\text{—}\; \tilde{u}_t,$$

with

$$\hat{\tilde{x}}_{\tilde{y}_t} = \mathbb{E}[\tilde{x}_t | \tilde{y}_t] = \mathbb{E}\left[ \begin{bmatrix} x_t \\ m_t \end{bmatrix} \,\Big|\, \begin{bmatrix} y_t \\ m_t \end{bmatrix} \right]$$

and similarly for $\hat{\tilde{x}}_{\tilde{u}_t}$. This implies that we need only consider the first component $\hat{x}_{\tilde{y}_t}$ of $\hat{\tilde{x}}_{\tilde{y}_t}$, which is obtained from the observation $\tilde{y}_t$ using

$$K = \Sigma_x C^\intercal \Sigma_{\tilde{y}}^\dagger$$
$$= \begin{bmatrix} K_{x;y|m} & (I - K_{x;y|m} C_{y;x}) \Sigma_{x;m} \Sigma_m^\dagger \end{bmatrix},$$

where

$$K_{x;y|m} = \Sigma_{x|m} C_{y;x}^\intercal \Sigma_{y|m}^\dagger$$

is the Kalman gain that performs optimal inference in the classic LQG task [7].

Crucially, we see that $\hat{x}_{\tilde{y}_t}$ depends on $m_t$ only through

$$\hat{x}_{m_t} = \mathbb{E}[x_t | m_t] = \Sigma_{x;m} \Sigma_m^\dagger m_t.$$

This implies that, for a controller $\pi$, we can design an equivalent controller $\pi'$ whose memory state is the MMSE estimator $m_t' = \hat{x}_{m_t}$. The feedback gain for $\pi'$ is

$$L' = \begin{bmatrix} I & 0 \\ 0 & \Sigma_{x;m} \Sigma_m^\dagger \end{bmatrix} L.$$

Note that, since $m_t'$ is a sufficient statistic of $m_t$ for $x_t$, we have $\Sigma_{x|m'} = \Sigma_{x|m}$ and $K_{x;y|m'} = K_{x;y|m}$. Thus

$$K' = \begin{bmatrix} K_{x;y|m} & I - K_{x;y|m} C_{y;x} \end{bmatrix},$$

with $\Sigma_{x;m'} \Sigma_{m'}^\dagger = \Sigma_{m'} \Sigma_{m'}^\dagger$ in the second component omitted due to its redundancy.

The controllers $\pi$ and $\pi'$ generate the same control $u_t$ and thus incur the same external cost. At the same time, since $m_t'$ is a function of $m_t$, by the data-processing inequality the information rate of $\pi'$ is at most that of $\pi$. Thus any controller can be converted into a MMSE controller without loss of performance, allowing us to consider the MMSE controller canonical. In particular, this proves again that $d = n$ is always sufficient for representing the memory state.

We now diverge from the solution given in Section II-B, which has freedom in its choice of memory representation, and is therefore not guaranteed to be a MMSE controller. Instead, we explicitly constrain the controller to be MMSE, which in return enables us to relax some of the conditions given in Section II-B, which are now not necessary (and indeed do not hold at the optimum), as discussed below.

Constraining the controller to be MMSE imposes the structure

$$\Sigma_{\tilde{x}} = \begin{bmatrix} \Sigma_{x|m} + \Sigma_m & \Sigma_m \\ \Sigma_m & \Sigma_m \end{bmatrix},$$

parameterized by $\Sigma_{x|m}$ and $\Sigma_m$. The reduced number of independent parameters leaves $M$ overparameterized (see SM, Appendix VII) and we can choose, without loss of performance, the structure

$$M = \begin{bmatrix} M_{x|m} + M_m & -M_m \\ -M_m & M_m \end{bmatrix}$$

with

$$M_{x|m} = \beta^{-1} Z$$
$$M_m = \beta^{-1}(C_{y;x}^\intercal K_{x;y|m}^\intercal Z K_{x;y|m} C_{y;x} - Z),$$

where $Z = \Sigma^\dagger_{\hat{x}_{\tilde{y}}|\hat{x}_{\tilde{u}}} - \Sigma^\dagger_{\hat{x}_{\tilde{y}}}$ is the signal-to-noise-ratio (SNR) matrix for the channel $\hat{x}_{\tilde{y}_t} \to \hat{x}_{\tilde{u}_t}$. Due to the shrinkage effect of $K_{x;y|m} C_{y;x}$

$$M_m \preceq 0 \preceq M_{x|m} + M_m.$$

The Hessian of the cost-to-go now has the form

$$S = Q + A^\intercal S A - M$$
$$= \begin{bmatrix} Q_x + A_x^\intercal S_x A_x - M_{x|m} - M_m & M_m \\ M_m & -M_m \end{bmatrix}$$

and the second-order expansion of the cost-to-go, at the optimum, has the form

$$\tilde{x}_t^\intercal S \tilde{x}_t = x_t^\intercal (Q_x + A_x^\intercal S_x A_x - \beta^{-1} Z) x_t$$
$$- (m_t - x_t)^\intercal M_m (m_t - x_t).$$

The first term measures the divergence of the state $x_t$ from 0 and the second the divergence of the controller's estimator $m_t$ from the true state $x_t$, which is the expected form for a MMSE controller. Both terms link the SNR matrix $Z$ to the cost reduction. In this form, $S$ is again positive semidefinite, while now $M$ is generally not.

Finally, when $\beta = \infty$, we can recover the classic LQG results. Similarly to Part I [1, Section IV-B], we can substitute $N_{x|m}$ for $\beta^{-1} Z$, to recover the algebraic Riccati equation

$$S_{x|m} = Q_x + A_x^\intercal S_x A_x - N_{x|m}$$
$$= Q_x + A_x^\intercal (S_{x|m} - S_{x|m} B_{x;u} S_{u|m}^\dagger B_{x;u}^\intercal S_{x|m}) A_x.$$

## IV. EXAMPLE

As a simple example, consider the double mass-spring-damper system in Figure 5, adapted from [13]. The

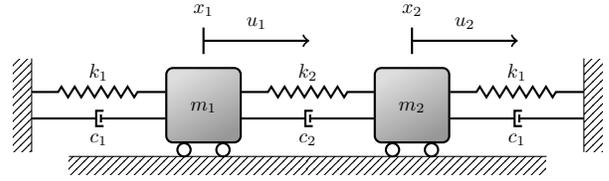

Fig. 5. Double mass-spring-damper system; masses: $m_1 = 5\,\mathrm{kg}$, $m_2 = \sqrt{15}\,\mathrm{kg}$; spring constants: $k_1 = 1\,\mathrm{N/m}$, $k_2 = 0.5\,\mathrm{N/m}$; damping coefficients: $c_1 = c_2 = 1\,\mathrm{N\cdot sec/m}$

continuous-time dynamics of this system are given by

$$A = \begin{bmatrix} 0 & 1 & 0 & 0 \\ -\frac{k_1+k_2}{m_1} & -\frac{c_1+c_2}{m_1} & \frac{k_2}{m_1} & \frac{c_2}{m_1} \\ 0 & 0 & 0 & 1 \\ \frac{k_2}{m_2} & \frac{c_2}{m_2} & -\frac{k_1+k_2}{m_2} & -\frac{c_1+c_2}{m_2} \end{bmatrix}$$

$$B = \begin{bmatrix} 0 & 0 \\ \frac{1}{m_1} & 0 \\ 0 & 0 \\ 0 & \frac{1}{m_2} \end{bmatrix} \quad C = \begin{bmatrix} 1 & 0 & 0 & 0 \\ 0 & 0 & 1 & 0 \end{bmatrix},$$

with $m_1 = 5\,\mathrm{kg}$, $m_2 = \sqrt{15}\,\mathrm{kg}$, $k_1 = 1\,\mathrm{N/m}$, $k_2 = 0.5\,\mathrm{N/m}$ and $c_1 = c_2 = 1\,\mathrm{N\cdot sec/m}$. We discretize the time using the Tustin transformation with sampling frequency 20Hz and consider the isotropic noises and cost rates

$$\Sigma_\xi = I \qquad \Sigma_\epsilon = I \qquad Q = I \qquad R = I.$$

For the memoryless control problem, we initialize a solution with $\Sigma_x = S = 0$. For the retentive control problem, we apply the reduction in Lemma 2 to obtain a reduced plant and then initialize a solution using the classic LQG controller, as described in Section III-C. To the initial solution, we apply the forward-backward iterations of Section II-B, with fixed $\beta$, until convergence to a fixed point, suspected as a global optimum. To improve running time, we employ a reverse-annealing scheme, decreasing $\beta$ gradually over its range and using the fixed point for one value of $\beta$ to initialize the iterations for the next value of $\beta$.

Figures 6 and 7 show, respectively, the resulting cost-log-beta and cost-information curves, demonstrating that even this simple example exhibits interesting phenomenology.

We see that both the memoryless (blue) and the retentive (green) controllers undergo phase transitions as $\beta$ increases. The system is controllable and observable, allowing the retentive controller to undergo 4 phase transitions, until it fully remembers and controls all modes of the system. However, the rank-2 matrices $B$ and $C$ only allow the memoryless controller to undergo 2 phase transitions and reach order $d = 2$.

In the first phase transition, the controllers begin controlling a single mode, in order to reduce the external cost, at the expense of communication resources. This is not depicted in the cost-information plot (Figure 7), since below this critical point the information is 0 and the cost is fixed.

The second and fourth phase transitions involve memory and only occur in the retentive controller. Below these critical point, a hypothetical order-2 retentive controller is worse than the order-1 controller, in terms of the target $\mathcal{F}$, the total external and internal cost-to-go it incurs. At the critical point, the order-2 controller overtakes the order-1 controller, already with a significantly reduced cost rate and a significant information rate (see red dots in Figures 6 and 7). The critical point is where the ratio between these costs is $\beta^{-1}$ (see (12) in Part I [1, Section IV-B]).

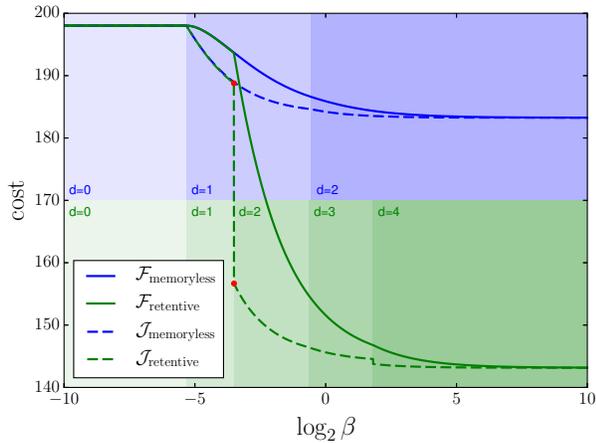

Fig. 6. Cost-log-beta curve for the double mass-spring-damper problem. Memoryless control (blue) generally incurs higher cost than retentive control (green). The Lagrangian $\mathcal{F}$ (solid) is continuous, whereas the external cost $\mathcal{J}$ (dashed) is discontinuous in the retentive case in phase transitions 2 (red dots) and 4. Background shades indicate the controller order $d$, with boundaries at critical points.

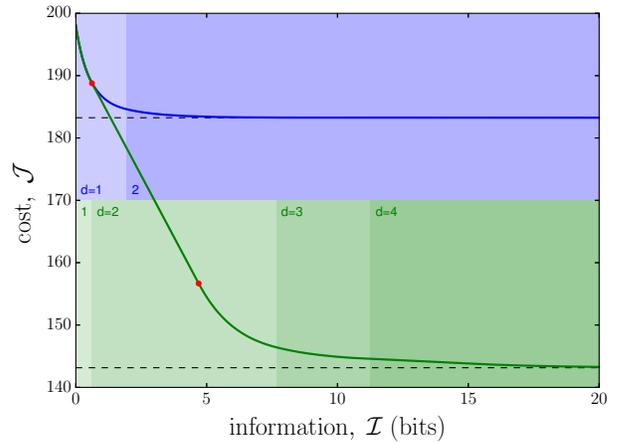

Fig. 7. Cost-information curve for the double mass-spring-damper problem. Memoryless control (blue) incurs higher cost than retentive control (green) after phase transition 2 (red dots). The asymptotic costs at $\beta = \infty$ (dashed black) can be approximated with very little information and a reduced order.

The third phase transition is again common to the memoryless and the retentive controllers, although by now the retentive controller has committed to memory much valuable information, reducing the cost much beyond the capabilities of the memoryless controller.

## V. Discussion

In this paper we introduce the problem of optimal LQG control with bounded channel capacity in both the memory and the sensory channels. We show how to reduce this problem to that of bounded memoryless LQG control, study the structure of the resulting solution and illustrate its interesting phenomenology with a simple example.

One aspect of this phenomenology that merits further study is the existence of suboptimal fixed points of the iterative algorithm (Section II-B). For example, around the second critical point in the double mass-spring-damper system (Section IV), both an order-1 controller and a retentive order-2 controller are fixed points. Before the phase transition, one of these solutions is stable, while the other is metastable and suboptimal, and at the phase transition they switch. This resembles well-studied phenomena in statistical physics.

LQG control with constraints on the sensory channel capacity has now been studied in the regime of unlimited memory [11], no memory (Part I of this work [1]) and in this paper, a shared channel capacity for sensing and memory. More generally, the memory and the sensory channels can be separate, with their relative costs ranging from 0 (no memory) to 1 (shared capacity) to $\infty$ (unlimited memory) including any intermediate value. This memory-sensory trade-off has been studied in the context of finite-state systems [4] and further insight can be gained from studying this more general problem in the LQG context.

## 3.3 Supplementary Material

Supplementary material for Part I and Part II.

# Minimum-Information LQG Control
# Supplementary Material

Roy Fox[†] and Naftali Tishby[†]

## APPENDIX I
## PERFECTLY MATCHED CHANNEL

In this appendix we construct a channel that is perfectly matched to the sequential source code derived in Theorem 1, in Part I of this paper [1, Section III-B]. Recall that in a perfectly matched source-channel pair the optimal source coding and the optimal channel coding can be implemented jointly for single letters, without requiring longer blocks. This allows us to use them in a perception-action cycle, where we cannot accumulate a block of inputs before emitting an output.

The main results of [2], applied to our setting, can be summarized as follows. We wish to find a memoryless channel into which we can input an encoding $w_t = g(\hat{x}_{y_t})$, such that $\hat{x}_{u_t} = h(\hat{w}_t)$ can be decoded from the channel output $\hat{w}_t$. Suppose that we are concerned with the power needed to transmit $w_t$ and thus the input cost is $w_t^\intercal w_t$. Then the source $\hat{x}_{y_t}$ and the channel $w_t \to \hat{w}_t$ are perfectly matched if there exist an encoder and a decoder such that

1) The Kullback-Leibler divergence $\mathbb{D}[f(\hat{w}_t|w_t)\|f(\hat{w}_t)]$ between the conditional and marginal densities of $\hat{w}_t$, as a function of $w_t$, equals $c_1 w_t^\intercal w_t + c_2$, for some constants $c_1 \geq 0$ and $c_2$; and
2) $f(\hat{x}_{u_t}|\hat{x}_{y_t})$ satisfies the conditions in Theorem 1.

To meet these conditions, we can choose the channel, the encoder and the decoder to have

$$w_t = D^{1/2} V^\intercal \Sigma_{\hat{x}_y}^{\dagger/2} \hat{x}_{y_t}$$
$$\hat{w}_t = w_t + v_t; \qquad v_t \sim \mathcal{N}(0, I - D)$$
$$\hat{x}_{u_t} = \Sigma_{\hat{x}_y}^{1/2} V D^{1/2} \hat{w}_t,$$

with $D$ and $V$ as in Theorem 1. Then

$$\Sigma_w = D$$
$$\Sigma_{\hat{w}} = I$$
$$\Sigma_{\hat{x}_u} = \Sigma_{\hat{x}_y}^{1/2} V D V^\intercal \Sigma_{\hat{x}_y}^{1/2} = \Sigma_{\hat{x}_u;\hat{x}_y},$$

and it can be verified that

$$\mathbb{D}[f(\hat{w}_t|w_t)\|f(\hat{w}_t)] = \tfrac{1}{2} w_t^\intercal \Sigma_{\hat{w}}^{-1} w_t + \text{const},$$

as required.

The capacity of the additive Gaussian noise channel with noise covariance $I - D$, under the appropriate expected power constraint, is indeed achieved by a Gaussian input with covariance $D$ and is equal to the information rate in Theorem 1. As shown in [2], this means that constraining the expected power $\Sigma_w$ is equivalent to constraining the information rate $\mathbb{I}[\hat{x}_{y_t}; \hat{x}_{u_t}]$.

Note, however, that the matched channel noise covariance depends on the constraint, through the solution in Theorem 1. Moreover, this result is not applicable when the best channel available to the designer of the controller is not the matched channel above, in which case both the channel and the sequential source coding generally need to be adapted.

## APPENDIX II
## PROOF OF LEMMA 1 OF PART I

In this appendix we restate and prove Lemma 1 of Part I [1, Section IV-A].

*Lemma 1:* Let $x$ and $\hat{x}$ be 0-mean jointly Gaussian random variables. The following properties are equivalent:

1) There exists a random variable $u$, jointly Gaussian with $x$, such that $\hat{x}(u) = \arg\min_{\hat{x}} \mathbb{E}[\|\hat{x} - x\|^2|u] = \mathbb{E}[x|u]$.
2) $\Sigma_{\hat{x};x} = \Sigma_{\hat{x}}$.
3) $\Sigma_{x|\hat{x}} = \Sigma_x - \Sigma_{\hat{x}}$, where $\Sigma_{x|\hat{x}}$ is the conditional covariance matrix of $x$ given $\hat{x}$, implying $\Sigma_x \succeq \Sigma_{\hat{x}}$.
4) $\hat{x} = \mathbb{E}[x|\hat{x}]$.

Such $\hat{x}$ is called a minimum mean square error (MMSE) estimator (of $u$) for $x$.

*Proof:* (1 $\implies$ 2) Assume without loss of generality that $u$ has mean 0. Then

$$\hat{x} = \Sigma_{x;u} \Sigma_u^\dagger u,$$

implying

$$\Sigma_{\hat{x};x} = \Sigma_{x;u} \Sigma_u^\dagger \Sigma_{u;x} = \Sigma_{\hat{x}}.$$

(2 $\implies$ 3)

$$\Sigma_{x|\hat{x}} = \Sigma_x - \Sigma_{x;\hat{x}} \Sigma_{\hat{x}}^\dagger \Sigma_{\hat{x};x} = \Sigma_x - \Sigma_{\hat{x}}.$$

(3 $\implies$ 4) Since $x$ and $\hat{x}$ are 0-mean and jointly Gaussian, we can write for some $T$

$$x = T\hat{x} + \xi; \qquad \xi \sim \mathcal{N}(0, \Sigma_{x|\hat{x}}),$$

implying

$$\Sigma_x = T \Sigma_{\hat{x}} T^\intercal + \Sigma_x - \Sigma_{\hat{x}},$$

thus without loss of generality $T = I$.

[†]School of Computer Science and Engineering, The Hebrew University, {royf,tishby}@cs.huji.ac.il
[*]This work was supported by the DARPA MSEE Program, the Gatsby Charitable Foundation, the Israel Science Foundation and the Intel ICRI-CI Institute

($4 \implies 1$) Taking $u = \hat{x}$, we have

$$\arg\min_{\hat{x}'} \mathbb{E}[\|\hat{x}' - x\|^2 | u]$$
$$= \arg\min_{\hat{x}'} (\hat{x}'^\intercal \hat{x}' - 2\hat{x}'^\intercal \mathbb{E}[x|u]) + \mathbb{E}[x^\intercal x | u],$$

which is optimized by $\hat{x}' = \mathbb{E}[x|u]$. □

## APPENDIX III
### PROOF OF LEMMA 2 OF PART I

In this appendix we restate and prove Lemma 2 of Part I [1, Section IV-A].

*Lemma 2:* The bounded memoryless LTI controller optimization problem (Problem 1) is solved by a control law of the form

$$\hat{x}_{y_t} = K y_t \tag{5a}$$
$$\hat{x}_{u_t} = W \hat{x}_{y_t} + \omega_t; \qquad \omega_t \sim \mathcal{N}(0, \Sigma_\omega) \tag{5b}$$
$$u_t = L \hat{x}_{u_t}, \tag{5c}$$

where $W \in \mathbb{R}^{n \times n}$, $\Sigma_\omega \in \mathbb{R}^{n \times n}$, $L \in \mathbb{R}^{\ell \times n}$, $\omega_t$ is independent of $y_t$, $\hat{x}_{u_t}$ is a MMSE estimator for $\hat{x}_{y_t}$ and

$$\mathbb{I}[y_t; u_t] = \mathbb{I}[\hat{x}_{y_t}; \hat{x}_{u_t}]. \tag{6}$$

*Proof:* Consider a LTI controller $\pi$ of the form

$$u_t = H y_t + \eta_t; \qquad \eta_t \sim \mathcal{N}(0, \Sigma_\eta), \tag{III.1}$$

satisfying the Markov network

$$\begin{array}{ccc} x_t & \!\!\!-\!\!\! & y_t & \!\!\!-\!\!\! & u_t \\ & & | & & | \\ & & \hat{x}_{y_t} & & \hat{x}_{u_t}. \end{array}$$

We now construct a controller $\pi'$ with control law $u'_t$ based on the estimator $\hat{x}'_{u_t}$ by defining the Markov chain

$$x_t \;-\; y_t \;-\; \hat{x}_{y_t} \;-\; u''_t \;-\; \hat{x}'_{u_t} \;-\; u'_t$$

such that each consecutive pair of variables has the same joint distribution as their unprimed namesakes. Since $\hat{x}_{y_t}$ is a sufficient statistic of $y_t$ for $x_t$, we have the Markov chain $x_t - \hat{x}_{y_t} - y_t - u_t$, implying that $u''_t$ has the same joint distribution with $x_t$ as $u_t$ does. Likewise, $\hat{x}'_{u_t}$ has the same joint distribution with $x_t$ as $\hat{x}_{u_t}$ does. Since $\hat{x}_{u_t}$ is a sufficient statistic of $u_t$ for $x_t$, we have that $u'_t$ also has the same joint distribution with $x_t$ as $u_t$ does.

Thus the controller $\pi'$ induces the same stochastic process $\{x_t, u'_t\}$ and the same external cost. Note that $u'_t$ may not have the same joint distribution with $y_t$ as $u_t$ does and due to the data-processing inequality [3]

$$\mathbb{I}[y_t; u_t] \geq \mathbb{I}[\hat{x}_{y_t}; u_t] = \mathbb{I}[\hat{x}_{y_t}; u''_t]$$
$$\geq \mathbb{I}[\hat{x}_{y_t}; \hat{x}'_{u_t}] \geq \mathbb{I}[y_t; u'_t].$$

Therefore $\pi'$ performs at least as well as $\pi$ and equally well when $\pi$ is optimal, proving (6).

$\hat{x}'_{u_t}$ is a MMSE estimator for $\hat{x}_{y_t}$ since

$$\mathbb{E}[\hat{x}_{y_t} | \hat{x}'_{u_t}] = \mathbb{E}[\mathbb{E}[x_t | y_t] | \hat{x}_{u'_t}]$$
$$= \mathbb{E}[x_t | \hat{x}'_{u_t}] = \hat{x}'_{u_t},$$

where the second equality follows from $x_t - y_t - \hat{x}'_{u_t}$.

Finally, it may not be clear from the above analysis that $u'_t$ is optimally deterministic in $\hat{x}'_{u_t}$. If $u_t$ has covariance $\Sigma_\nu$ given $\hat{x}'_{u_t}$, the Lagrangian of the optimization problem ((9) in Part I) depends on $\Sigma_\nu$ only through the terms

$$\tfrac{1}{2}(\mathrm{tr}(R\,\Sigma_\nu) + \mathrm{tr}(SB\,\Sigma_\nu\,B^\intercal)).$$

Since $R + B^\intercal S B \succeq 0$ is positive semidefinite, we can take $\Sigma_\nu = 0$ without loss of performance, recovering the structure (5). Intuitively, the argument is that any noise added to $u'_t$, beyond $\hat{x}'_{u_t}$, is not helpful in compressing $x_t$ and can only increase the external cost without saving any communication cost.

In the other direction, let $u_t$ satisfy the form of Lemma 2. We can rewrite $u_t$ in the form (III.1), with

$$H = LWK$$
$$\Sigma_\eta = L\,\Sigma_\omega\,L^\intercal. \qquad \square$$

## APPENDIX IV
### PROOF OF THEOREM 1 OF PART I

In this appendix we restate and prove Theorem 1 of Part I [1, Section IV-A], which relies on the following Lagrangian developed there.

$$\mathcal{F}_{\Sigma_x, \Sigma_{\hat{x}_u}, L, S; \beta} = \tfrac{1}{2}(\beta^{-1}(\log|\Sigma_{\hat{x}_y}|_\dagger - \log|\Sigma_{\hat{x}_y | \hat{x}_u}|_\dagger) \tag{9}$$
$$+ \mathrm{tr}(Q\,\Sigma_x) + \mathrm{tr}(RL\,\Sigma_{\hat{x}_u}\,L^\intercal)$$
$$+ \mathrm{tr}(S((A+BL)\,\Sigma_{\hat{x}_u}(A+BL)^\intercal$$
$$+ A\,\Sigma_{x|\hat{x}_u}\,A^\intercal + \Sigma_\xi - \Sigma_x))).$$

*Theorem 1:* Given $\beta$, the Lagrangian (9) is minimized by a controller satisfying the forward equations

$$\Sigma_x = (A+BL)\,\Sigma_{\hat{x}_u}(A+BL)^\intercal \tag{10a}$$
$$\qquad + A\,\Sigma_{x|\hat{x}_u}\,A^\intercal + \Sigma_\xi$$
$$\Sigma_y = C\,\Sigma_x\,C^\intercal + \Sigma_\epsilon \tag{10b}$$
$$K = \Sigma_x\,C^\intercal\,\Sigma_y^\dagger \tag{10c}$$
$$\Sigma_{\hat{x}_y} = K\,\Sigma_y\,K^\intercal, \tag{10d}$$

the backward equations

$$M = \beta^{-1} C^\intercal K^\intercal (\Sigma_{\hat{x}_y | \hat{x}_u}^\dagger - \Sigma_{\hat{x}_y}^\dagger) K C \tag{10e}$$
$$S = Q + A^\intercal S A - M, \tag{10f}$$
$$L = -(R + B^\intercal S B)^\dagger B^\intercal S A \tag{10g}$$
$$N = L^\intercal (R + B^\intercal S B) L \tag{10h}$$

and the control-based estimator covariance

$$\Sigma_{\hat{x}_u} = \Sigma_{\hat{x}_y}^{1/2} V D V^\intercal \Sigma_{\hat{x}_y}^{1/2}, \tag{10i}$$

the latter determined by the eigenvalue decomposition (EVD)

$$V \Lambda V^\intercal = \Sigma_{\hat{x}_y}^{1/2} N \Sigma_{\hat{x}_y}^{1/2} \tag{10j}$$

having $V$ orthogonal with $n - \mathrm{rank}(\Sigma_{\hat{x}_y})$ columns spanning the kernel of $\Sigma_{\hat{x}_y}$ and $\Lambda = \mathrm{diag}\{\lambda_i\}$ and by the active mode coefficient matrix

$$D = \mathrm{diag}\left\{\begin{array}{ll} 1 - \beta^{-1}\lambda_i^{-1} & \lambda_i > \beta^{-1} \\ 0 & \lambda_i \leq \beta^{-1} \end{array}\right\}. \tag{10k}$$

*Proof:* The minimum of the Lagrangian (9) must satisfy the first-order optimality conditions, i.e. that the gradient with respect to each parameter is 0 at the optimum. We start by differentiating $\mathcal{F}$ by the feedback gain $L$

$$\partial_L \mathcal{F}_{\Sigma_x, \Sigma_{\hat{x}_u}, L, S; \beta} = RL\Sigma_{\hat{x}_u} + B^\intercal S(A+BL)\Sigma_{\hat{x}_u} = 0,$$

which we rewrite as

$$(R + B^\intercal SB)L\Sigma_{\hat{x}_u} = -B^\intercal SA\Sigma_{\hat{x}_u}.$$

As this equation shows, $L$ is underdetermined in the kernel of $\Sigma_{\hat{x}_u}$, since these modes are always 0 in $\hat{x}_{u_t}$ and have no effect on $u_t$. $L$ is also underdetermined in the kernel of $R + B^\intercal SB$, since these modes have no cost (immediate or future) and can be controlled in any way without affecting the solution's performance. Thus without loss of performance we can take

$$L = -(R + B^\intercal SB)^\dagger B^\intercal SA.$$

We substitute this solution back into the Lagrangian, to get

$$\mathcal{F}_{\Sigma_x, \Sigma_{\hat{x}_u}, S; \beta} = \tfrac{1}{2}(\beta^{-1}(\log|\Sigma_{\hat{x}_y}|_\dagger - \log|\Sigma_{\hat{x}_y|\hat{x}_u}|_\dagger) \quad \text{(IV.1)}$$
$$+ \operatorname{tr}(M\Sigma_x) - \operatorname{tr}(N\Sigma_{\hat{x}_u}) + \operatorname{tr}(S\Sigma_\xi)),$$

with

$$M = Q + A^\intercal SA - S$$
$$N = L^\intercal(R + B^\intercal SB)L$$
$$= A^\intercal SB(R + B^\intercal SB)^\dagger B^\intercal SA.$$

The problem of optimizing over $\Sigma_{\hat{x}_u}$ given the other parameters can now be written, up to constants, as the semidefinite program (SDP)

$$\max_{\Sigma_{\hat{x}_u}} \quad \log|\Sigma_{\hat{x}_y} - \Sigma_{\hat{x}_u}|_\dagger + \beta \operatorname{tr}(N\Sigma_{\hat{x}_u})$$
$$\text{s.t.} \quad 0 \preceq \Sigma_{\hat{x}_u} \preceq \Sigma_{\hat{x}_y}.$$

By Lemma V.1 in Appendix V, the optimum is achieved when $\Sigma_{\hat{x}_u}$ satisfies (10i)–(10k).

Finally, with $P = \Sigma_{\hat{x}_y}\Sigma_{\hat{x}_y}^\dagger$ the projection onto the support of $\hat{x}_{y_t}$ and since the range of $\Sigma_{\hat{x}_u}$ is contained in that subspace, we have

$$\partial_{(\Sigma_x)_{i,j}}(\log|\Sigma_{\hat{x}_y}|_\dagger - \log|\Sigma_{\hat{x}_y|\hat{x}_u}|_\dagger)$$
$$= -\partial_{(\Sigma_x)_{i,j}}\log|P - \Sigma_{\hat{x}_u}\Sigma_{\hat{x}_y}^\dagger|_\dagger$$
$$= -\partial_{(\Sigma_x)_{i,j}}\log|I - \Sigma_{\hat{x}_u}(P\Sigma_{\hat{x}_y}P)^\dagger|$$
$$= \operatorname{tr}((I - \Sigma_{\hat{x}_u}\Sigma_{\hat{x}_y}^\dagger)^{-1}\Sigma_{\hat{x}_u}\partial_{(\Sigma_x)_{i,j}}(P\Sigma_{\hat{x}_y}P)^\dagger).$$

The purpose of introducing $P$ is to notice that even if the range of $\Sigma_{\hat{x}_y}$ is increased, this has no effect on the Lagrangian, because these modes are orthogonal to the range of $\Sigma_{\hat{x}_u}$. This allows us to treat $P$ as constant, so that the range of $P\Sigma_{\hat{x}_y}P$ is constant in a neighborhood of the solution, and the derivative of the pseudoinverse is simplified in this case to

$$\partial_{(\Sigma_x)_{i,j}}(P\Sigma_{\hat{x}_y}P)^\dagger = -\Sigma_{\hat{x}_y}^\dagger(\partial_{(\Sigma_x)_{i,j}}\Sigma_{\hat{x}_y})\Sigma_{\hat{x}_y}^\dagger$$
$$= -\Sigma_{\hat{x}_y}^\dagger KCJ_{i,j}C^\intercal K^\intercal \Sigma_{\hat{x}_y}^\dagger,$$

with $J_{i,j}$ the matrix with 1 in position $(i,j)$ and 0 elsewhere. This yields

$$\partial_{\Sigma_x}\mathcal{F}_{\Sigma_x, \Sigma_{\hat{x}_u}, S; \beta}$$
$$= \tfrac{1}{2}(M - \beta^{-1}C^\intercal K^\intercal \Sigma_{\hat{x}_y}^\dagger(I - \Sigma_{\hat{x}_u}\Sigma_{\hat{x}_y}^\dagger)^{-1}\Sigma_{\hat{x}_u}\Sigma_{\hat{x}_y}^\dagger KC)$$
$$= \tfrac{1}{2}(M - \beta^{-1}C^\intercal K^\intercal \Sigma_{\hat{x}_y}^\dagger((I - \Sigma_{\hat{x}_u}\Sigma_{\hat{x}_y}^\dagger)^{-1} - I)KC)$$
$$= \tfrac{1}{2}(M - \beta^{-1}C^\intercal K^\intercal (\Sigma_{\hat{x}_y|\hat{x}_u}^\dagger - \Sigma_{\hat{x}_y}^\dagger)KC) = 0,$$

implying (10e). □

## APPENDIX V
### SEMIDEFINITE PROGRAM SOLUTION

In this appendix we state and prove the following solution to our SDP problem.

*Lemma V.1:* The semidefinite program

$$\max_{X \in \mathbb{S}_+^n} \quad \log|M_1 - X|_\dagger + \operatorname{tr}(M_2 X)$$
$$\text{s.t.} \quad X \preceq M_1,$$

with $M_1, M_2 \succeq 0$, is optimized by

$$X = M_1^{1/2} V D V^\intercal M_1^{1/2},$$

with the eigenvalue decomposition (EVD)

$$V\Lambda V^\intercal = M_1^{1/2} M_2 M_1^{1/2},$$

such that $V$ is orthogonal with $n - \operatorname{rank}(M_1)$ columns spanning the kernel of $M_1$ and $\Lambda = \operatorname{diag}\{\lambda_i\}$ and with

$$D = \operatorname{diag}\left\{\begin{array}{ll} 1 - \lambda_i^{-1} & \lambda_i > 1 \\ 0 & \lambda_i \leq 1 \end{array}\right\}.$$

*Proof:* Let the EVD of $M_1$ be

$$U\Psi U^\intercal = M_1,$$

with $U$ orthogonal and $\Psi$ diagonal, having

$$\Psi = \begin{bmatrix} \Psi_+ & 0 \\ 0 & 0_{(n-m)\times(n-m)} \end{bmatrix},$$

with $m = \operatorname{rank}(M_1)$. Let

$$\Psi^\ddagger = \Psi^\dagger + I - \Psi^\dagger\Psi = \begin{bmatrix} \Psi_+^{-1} & 0 \\ 0 & I \end{bmatrix}.$$

By changing the variable to

$$Y = \Psi^{\ddagger/2} U^\intercal X U \Psi^{\ddagger/2},$$

the constraint of the SDP becomes

$$Y \preceq I_{m,n} = \begin{bmatrix} I_{m\times m} & 0 \\ 0 & 0_{(n-m)\times(n-m)} \end{bmatrix}.$$

$Y$ must therefore be 0 outside the upper-left $m \times m$ block, and the SDP is equivalent, up to constants, to

$$\max_{Y \in \mathbb{S}_+^n} \quad \log|I_{m,n} - Y|_\dagger + \operatorname{tr}(\Psi^{1/2} U^\intercal M_2 U \Psi^{1/2} Y)$$
$$\text{s.t.} \quad Y \preceq I_{m,n}.$$

Let the EVD of the linear coefficient be

$$\bar{V}\Lambda\bar{V}^\intercal = \Psi^{1/2} U^\intercal M_2 U \Psi^{1/2},$$

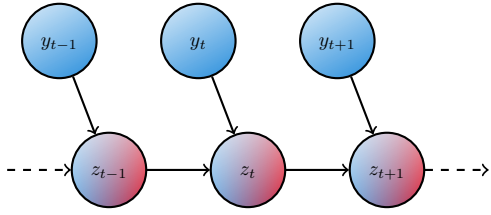

Fig. VI.1. Bayesian network of online inference from a sequence of independent observations

with

$$\bar{V} = \begin{bmatrix} \bar{V}_+ & 0 \\ 0 & I_{(n-m)\times(n-m)} \end{bmatrix}$$

orthogonal and preserving the kernel of $\Psi$ and $\Lambda = \mathrm{diag}\{\lambda_i\}$. We can again change the variable to

$$D = \bar{V}^{\mathsf{T}} Y \bar{V},$$

to get

$$\max_{D \in \mathbb{S}^n_+} \log |I_{m,n} - D|_\dagger + \mathrm{tr}(\Lambda D)$$

$$\text{s.t.} \quad D \preceq I_{m,n},$$

which can easily be solved using Hadamard's inequality [3], to find

$$D = \mathrm{diag} \left\{ \begin{array}{ll} 1 - \lambda_i^{-1} & \lambda_i > 1 \\ 0 & \lambda_i \leq 1 \end{array} \right\}.$$

Finally, the lemma follows by unmaking the variable changes and taking

$$V = U\bar{V}. \qquad \square$$

## APPENDIX VI
### PROPERTIES OF THE RETENTIVE DIRECTED INFORMATION

In this appendix we show how the retentive directed information (Definition 6 of Part II [4, Section III-A]) relates to the multi-information of Bayesian networks [5].

Consider the Bayesian network in Figure VI.1, which describes the process of online inference from a sequence of independent observations. The multi-information of this network, for horizon $T$, is equal to the retentive directed information

$$\mathbb{I}[y^T, z^T] = \mathbb{E}\left[\log \frac{f(y^T, z^T)}{\prod_{t=1}^T f(y_t)f(z_t)}\right]$$

$$= \sum_{t=1}^T \mathbb{E}\left[\log \frac{f(z_t|z^{t-1}, y^t)}{f(z_t)}\right] = \mathbb{I}[y^T \twoheadrightarrow z^T].$$

An important property of the directed information is that the mutual information between two sequences can be decomposed into the sum of directed information in both directions [6]

$$\mathbb{I}[x^T; z^T] = \mathbb{I}[x^T \to z^T] + \mathbb{I}[z^T \to x^T].$$

Interestingly, retentive directed information extends this property to the retentive control process (Figure 1 in Part II). This process can be thought of as consisting of four phases: observation, inference, control and state transition. Its multi-information can accordingly be decomposed [7] into the sum

$$\mathbb{I}[x^T, y^T, z^T, u^T] = \mathbb{I}[x^T \twoheadrightarrow y^T] + \mathbb{I}[y^T \twoheadrightarrow z^T] + \mathbb{I}[z^T \twoheadrightarrow u^T] + \mathbb{I}[u^T \twoheadrightarrow x^T].$$

## APPENDIX VII
### STRUCTURE OF THE OPTIMAL RETENTIVE CONTROLLER

In this appendix we derive the structure of the optimal retentive controller summarized in Part II [4, Section III-C].

For the structured feedback gain $L$ we find using the Schur complement that

$$(R + B^{\mathsf{T}} SB)^\dagger = \begin{bmatrix} R_u + B^{\mathsf{T}}_{x;u} S_x B_{x;u} & B^{\mathsf{T}}_{x;u} S_{x;m} \\ S_{m;x} B_{x;u} & S_m \end{bmatrix}^\dagger$$

$$= \begin{bmatrix} S^\dagger_{u|m} & -S^\dagger_{u|m} B^{\mathsf{T}}_{x;u} S_{x;m} S^\dagger_m \\ -S^\dagger_m S_{m;x} B_{x;u} S^\dagger_{u|m} & S^\dagger_{m|u} \end{bmatrix},$$

with

$$S^\dagger_{m|u} = S^\dagger_m + S^\dagger_m S_{m;x} B_{x;u} S^\dagger_{u|m} B^{\mathsf{T}}_{x;u} S_{x;m} S^\dagger_m,$$

and so

$$L = -(R + B^{\mathsf{T}} SB)^\dagger B^{\mathsf{T}} SA$$

$$= -(R + B^{\mathsf{T}} SB)^\dagger \begin{bmatrix} B^{\mathsf{T}}_{x;u} S_x A_x & 0 \\ S_{m;x} A_x & 0 \end{bmatrix}$$

$$= -\begin{bmatrix} S^\dagger_{u|m} B^{\mathsf{T}}_{x;u} S_{x|m} A_x & 0 \\ S^\dagger_m S_{m;x}(I - B_{x;u} S^\dagger_{u|m} B^{\mathsf{T}}_{x;u} S_{x|m}) A_x & 0 \end{bmatrix}$$

$$= \begin{bmatrix} L_{u;x|m} & 0 \\ -S^\dagger_m S_{m;x}(A_x + B_{x;u} L_{u;x|m}) & 0 \end{bmatrix},$$

with

$$L_{u;x|m} = -S^\dagger_{u|m} B^{\mathsf{T}}_{x;u} S_{x|m} A_x.$$

We also have

$$N = L^{\mathsf{T}}(R + B^{\mathsf{T}} SB)L$$

$$= A^{\mathsf{T}} SB(R + B^{\mathsf{T}} SB)^\dagger B^{\mathsf{T}} SA = \begin{bmatrix} N_{x|m} & 0 \\ 0 & 0 \end{bmatrix}$$

$$N_{x|m} = \begin{bmatrix} B^{\mathsf{T}}_{x;u} S_x A_x \\ S_{m;x} A_x \end{bmatrix}^{\mathsf{T}} L \begin{bmatrix} I \\ 0 \end{bmatrix}$$

$$= A^{\mathsf{T}}_x(S_x - S_{x|m} + S_{x|m} B_{x;u} S^\dagger_{u|m} B^{\mathsf{T}}_{x;u} S_{x|m}) A_x.$$

Dually, for the structured Kalman gain $K$ we find that

$$\Sigma^\dagger_{\tilde{y}} = \begin{bmatrix} \Sigma_y & \Sigma_{y;m} \\ \Sigma_{m;y} & \Sigma_m \end{bmatrix}^\dagger$$

$$= \begin{bmatrix} \Sigma^\dagger_{y|m} & -\Sigma^\dagger_{y|m} \Sigma_{y;m} \Sigma^\dagger_m \\ -\Sigma^\dagger_m \Sigma_{m;y} \Sigma^\dagger_{y|m} & \Sigma^\dagger_m + \Sigma^\dagger_m \Sigma_{m;y} \Sigma^\dagger_{y|m} \Sigma_{y;m} \Sigma^\dagger_m \end{bmatrix},$$

and so

$$K = \Sigma_x C^{\mathsf{T}} \Sigma^\dagger_{\tilde{y}}$$

$$= \begin{bmatrix} \Sigma_x C^{\mathsf{T}}_{y;x} & \Sigma_{x;m} \end{bmatrix} \begin{bmatrix} \Sigma_y & \Sigma_{y;m} \\ \Sigma_{m;y} & \Sigma_m \end{bmatrix}^\dagger$$

$$= \begin{bmatrix} K_{x;y|m} & (I - K_{x;y|m} C_{y;x}) \Sigma_{x;m} \Sigma^\dagger_m \end{bmatrix},$$

with
$$K_{x;y|m} = \Sigma_{x|m} C_{y;x}^\intercal \Sigma_{y|m}^\dagger.$$

Now constraining the controller to be MMSE, we have the structure
$$\Sigma_{\tilde{x}} = \begin{bmatrix} \Sigma_{x|m} + \Sigma_m & \Sigma_m \\ \Sigma_m & \Sigma_m \end{bmatrix}$$
$$K = \begin{bmatrix} K_{x;y|m} & I - K_{x;y|m} C_{y;x} \end{bmatrix},$$

which we employ in differentiating $\mathcal{F}$ (IV.1), to get

$$\partial_{\Sigma_{x|m}} \mathcal{F}_{\Sigma_{x|m}, \Sigma_m, \Sigma_{\hat{x}_{\tilde{u}}}, S; \beta} = \begin{bmatrix} I \\ 0 \end{bmatrix}^\intercal \partial_{\Sigma_{\tilde{x}}} \mathcal{F}_{\Sigma_{\tilde{x}}, \Sigma_{\hat{x}_{\tilde{u}}}, S; \beta} \begin{bmatrix} I \\ 0 \end{bmatrix}$$
$$= \tfrac{1}{2} \begin{bmatrix} I \\ 0 \end{bmatrix}^\intercal (M - \beta^{-1} C^\intercal K^\intercal Z K C) \begin{bmatrix} I \\ 0 \end{bmatrix}$$
$$= \tfrac{1}{2} \left( \begin{bmatrix} I \\ 0 \end{bmatrix}^\intercal M \begin{bmatrix} I \\ 0 \end{bmatrix} - \beta^{-1} C_{y;x}^\intercal K_{x;y|m}^\intercal Z K_{x;y|m} C_{y;x} \right) = 0$$
$$\partial_{\Sigma_m} \mathcal{F}_{\Sigma_{x|m}, \Sigma_m, \Sigma_{\hat{x}_{\tilde{u}}}, S; \beta} = \begin{bmatrix} I \\ I \end{bmatrix}^\intercal \partial_{\Sigma_{\tilde{x}}} \mathcal{F}_{\Sigma_{\tilde{x}}, \Sigma_{\hat{x}_{\tilde{u}}}, S; \beta} \begin{bmatrix} I \\ I \end{bmatrix}$$
$$= \tfrac{1}{2} \begin{bmatrix} I \\ I \end{bmatrix}^\intercal (M - \beta^{-1} C^\intercal K^\intercal Z K C) \begin{bmatrix} I \\ I \end{bmatrix}$$
$$= \tfrac{1}{2} \left( \begin{bmatrix} I \\ I \end{bmatrix}^\intercal M \begin{bmatrix} I \\ I \end{bmatrix} - \beta^{-1} Z \right) = 0,$$

with
$$Z = \Sigma_{\hat{x}_{\tilde{y}}|\hat{x}_{\tilde{u}}}^\dagger - \Sigma_{\hat{x}_{\tilde{y}}}^\dagger.$$

This leaves $M$ overparameterized and we can choose to give it the structure
$$M = \begin{bmatrix} M_{x|m} + M_m & -M_m \\ -M_m & M_m \end{bmatrix}$$

with
$$M_{x|m} = \beta^{-1} Z$$
$$M_m = \beta^{-1} (C_{y;x}^\intercal K_{x;y|m}^\intercal Z K_{x;y|m} C_{y;x} - Z).$$

# Chapter 4

# Minimum-KL Reinforcement Learning

## 4.1 Taming the Noise in Reinforcement Learning via Soft Updates

Published: Roy Fox*, Ari Pakman* and Naftali Tishby, *Taming the Noise in Reinforcement Learning via Soft Updates*, In Proceedings of the 32nd Conference on Uncertainty in Artificial Intelligence (UAI), 2016.

*These authors contributed equally to this work.

# Taming the Noise in
# Reinforcement Learning via Soft Updates

**Roy Fox**[*]
Hebrew University

**Ari Pakman**[*]
Columbia University

**Naftali Tishby**
Hebrew University

## Abstract

Model-free reinforcement learning algorithms, such as Q-learning, perform poorly in the early stages of learning in noisy environments, because much effort is spent unlearning biased estimates of the state-action value function. The bias results from selecting, among several noisy estimates, the apparent optimum, which may actually be suboptimal. We propose G-learning, a new off-policy learning algorithm that regularizes the value estimates by penalizing deterministic policies in the beginning of the learning process. We show that this method reduces the bias of the value-function estimation, leading to faster convergence to the optimal value and the optimal policy. Moreover, G-learning enables the natural incorporation of prior domain knowledge, when available. The stochastic nature of G-learning also makes it avoid some exploration costs, a property usually attributed only to on-policy algorithms. We illustrate these ideas in several examples, where G-learning results in significant improvements of the convergence rate and the cost of the learning process.

## 1 INTRODUCTION

The need to separate signals from noise stands at the center of any learning task in a noisy environment. While a rich set of tools to regularize learned parameters has been developed for supervised and unsupervised learning problems, in areas such as reinforcement learning there still exists a vital need for techniques that tame the noise and avoid overfitting and local minima.

One of the central algorithms in reinforcement learning is Q-learning [1], a model-free off-policy algorithm, which attempts to estimate the optimal value function $Q$, the cost-to-go of the optimal policy. To enable this estimation, a stochastic exploration policy is used by the learning agent to interact with its environment and explore the model. This approach is very successful and popular, and despite several alternative approaches developed in recent years [2, 3, 4], it is still being applied successfully in complex domains for which explicit models are lacking [5].

However, in noisy domains, in early stages of the learning process, the min (or max) operator in Q-learning brings about a bias in the estimates. This problem is akin to the "winner's curse" in auctions [6, 7, 8, 9]. With too little evidence, the biased estimates may lead to wrong decisions, which slow down the convergence of the learning process, and require subsequent unlearning of these suboptimal behaviors.

In this paper we present G-learning, a new off-policy information-theoretic approach to regularizing the state-action value function learned by an agent interacting with its environment in model-free settings.

This is achieved by adding to the cost-to-go a term that penalizes deterministic policies which diverge from a simple stochastic prior policy [10]. With only a small sample to go by, G-learning prefers a more randomized policy, and as samples accumulate, it gradually shifts to a more deterministic and exploiting policy. This transition is managed by appropriately scheduling the coefficient of the penalty term as learning proceeds.

In Section 4 we discuss the theoretical and practical aspects of scheduling this coefficient, and suggest that a simple linear schedule can perform well. We show that G-learning with this schedule reduces the value estimation bias by avoiding overfitting in its selection of the update policy. We further establish empirically the link between bias reduction and learning performance, that has been the underlying assumption in many approaches to reinforcement learning [11, 12, 13, 14]. The examples in Section 6 demonstrate the significant improvement thus obtained.

Furthermore, in domains where exploration incurs significantly higher costs than exploitation, such as the classic cliff domain [2], G-learning with an $\epsilon$-greedy exploration policy is exploration-aware, and chooses a less costly exploration policy, thus reducing the costs incurred during the learning process. Such awareness to the cost of exploration is usually attributed to on-policy algorithms, such as SARSA [2, 4] and Expected-SARSA [15, 16]. The remarkable finding that G-learning exhibits on-policy-like properties is illustrated in the example of Section 6.2.

---

[*]These authors contributed equally to this work.

In Section 2 we discuss the problem of learning in noisy environments. In Section 3 we introduce the penalty term, derive G-learning and prove its convergence. In Section 4 we determine a schedule for the coefficient of the information penalty term. In Section 5 we discuss related work. In Section 6 we illustrate the strengths of the algorithm through several examples.

## 2 LEARNING IN NOISY ENVIRONMENTS

### 2.1 NOTATION AND BACKGROUND

We consider the usual setting of a Markov Decision Process (MDP), in which an agent interacts with its environment by repeatedly observing its state $s \in S$, taking an action $a \in A$, with $A$ and $S$ finite, and incurring cost $c \in \mathbb{R}$. This induces a stochastic process $s_0, a_0, c_0, s_1, \ldots$, where $s_0$ is fixed, and where for $t \geq 0$ we have the Markov properties indicated by the conditional distributions $a_t \sim \pi_t(a_t|s_t)$, $c_t \sim \theta(c_t|s_t, a_t)$ and $s_{t+1} \sim p(s_{t+1}|s_t, a_t)$.

The objective of the agent is to find a time-invariant policy $\pi$ that minimizes the total discounted expected cost

$$V^\pi(s) = \sum_{t \geq 0} \gamma^t \, \mathrm{E}[c_t|s_0 = s], \qquad (1)$$

simultaneously for any $s \in S$, for a given discount factor $0 \leq \gamma < 1$. For each $t$, the expectation above is over all trajectories of length $t$ starting at $s_0 = s$. A related quantity is the state-action value function

$$Q^\pi(s, a) = \sum_{t \geq 0} \gamma^t \, \mathrm{E}[c_t|s_0 = s, a_0 = a]$$
$$= \mathrm{E}_\theta[c|s, a] + \gamma \, \mathrm{E}_p[V^\pi(s')|s, a], \qquad (2)$$

which equals the total discounted expected cost that follows from choosing action $a$ in state $s$, and then following the policy $\pi$.

If we know the distributions $p$ and $\theta$ (or at least $\mathrm{E}_\theta[c|s, a]$), then it is easy to find the optimal state-action value function

$$Q^*(s, a) = \min_\pi Q^\pi(s, a) \qquad (3)$$

using standard techniques, such as Value Iteration [17]. Our interest is in model-free learning, where the model parameters are unknown. Instead, the agent obtains samples from $p(s_{t+1}|s_t, a_t)$ and $\theta(c_t|s_t, a_t)$ through its interaction with the environment. In this setting, the Q-learning algorithm [1] provides a method for estimating $Q^*$. It starts with an arbitrary $Q$, and in step $t$ upon observing $s_t, a_t, c_t$ and $s_{t+1}$, performs the update

$$Q(s_t, a_t) \leftarrow (1 - \alpha_t) Q(s_t, a_t) \qquad (4)$$
$$+ \alpha_t \left( c_t + \gamma \sum_{a'} \pi(a'|s_{t+1}) Q(s_{t+1}, a') \right),$$

with some learning rate $0 \leq \alpha_t \leq 1$, and the greedy policy for $Q$ having

$$\pi(a|s) = \delta_{a, a^*(s)}; \qquad a^*(s) = \arg\min_a Q(s, a). \qquad (5)$$

$Q(s, a)$ is unchanged for any $(s, a) \neq (s_t, a_t)$. If the learning rate satisfies

$$\sum_t \alpha_t = \infty; \qquad \sum_t \alpha_t^2 < \infty, \qquad (6)$$

and the interaction itself uses an exploration policy that returns to each state-action pair infinitely many times, then $Q$ is a consistent estimator, converging to $Q^*$ with probability 1 [1, 17]. Similarly, if the update rule (4) uses a fixed update policy $\pi = \rho$, we call this algorithm $Q^\rho$-learning, because $Q$ converges to $Q^\rho$ with probability 1.

### 2.2 BIAS AND EARLY COMMITMENT

Despite the success of Q-learning in many situations, learning can proceed extremely slowly when there is noise in the distribution, given $s_t$ and $a_t$, of either of the terms of (2), namely the cost $c_t$ and the value of the next state $s_{t+1}$. The source of this problem is a negative bias introduced by the min operator in the estimator $\min_{a'} Q(s_{t+1}, a')$, when (5) is plugged into (4).

To illustrate this bias, assume that $Q(s, a)$ is an unbiased but noisy estimate of the optimal $Q^*(s, a)$. Then Jensen's inequality for the concave min operator implies that

$$\mathrm{E}[\min_a Q(s, a)] \leq \min_a Q^*(s, a), \qquad (7)$$

with equality only when $Q$ already reveals the optimal policy by having $\arg\min_a Q(s, a) = \arg\min_a Q^*(s, a)$ with probability 1, so that no further learning is needed. The expectation in (7) is with respect to the learning process, including any randomness in state transition, cost, exploration and internal update, given the domain.

This is an optimistic bias, causing the cost-to-go to appear lower than it is (or the reward-to-go higher). It is the well known "winner's curse" problem in economics and decision theory [6, 7, 8, 9], and in the context of Q-learning it was studied before in [3, 11, 12, 13]. A similar problem occurs when a function approximation scheme is used for $Q$ instead of a table, even in the absence of transition or cost noise, because the approximation itself introduces noise [18].

As the sample size increases, the variance in $Q(s,a)$ decreases, which in turn reduces the bias in (7). This makes the update policy (5) more optimal, and the update increasingly similar to Value Iteration.

## 2.3 THE INTERPLAY OF VALUE BIAS AND POLICY SUBOPTIMALITY

It is insightful to consider the effect of the bias not only on the estimated value function, but also on the real value $V^\pi$ of the greedy policy (5), since in many cases the latter is the actual output of the learning process. The central quantity of interest here is the gap $Q^*(s,a') - V^*(s)$, in a given state $s$, between the value of a non-optimal action $a'$ and that of the optimal action.

Consider first the case in which the gap is large compared to the noise in the estimation of the $Q(s,a)$ values. In this case, $a'$ indeed appears suboptimal with high probability, as desired. Interestingly, when the gap is very small relative to the noise, the learning agent should not worry, either. Confusing such $a'$ for the optimal action has a limited effect on the value of the greedy policy, since choosing $a'$ is near-optimal.

We conclude that the real value $V^\pi$ of the greedy policy (5) is suboptimal only in the intermediate regime, when the gap is of the order of the noise, and neither is small. The effect of the noise can be made even worse by the propagation of bias between states, through updates. Such propagation can cause large-gap suboptimal actions to nevertheless appear optimal, if they lead to a region of state-space that is highly biased.

## 2.4 A DYNAMIC OPTIMISM-UNCERTAINTY LOOP

The above considerations were agnostic to the exploration policy, but the bias reduction can be accelerated by an exploration policy that is close to being greedy. In this case, high-variance estimation is self-correcting: an estimated state value with optimistic bias draws exploration towards that state, leading to a decrease in the variance, which in turn reduces the optimistic bias. This is a dynamic form of optimism under uncertainty. While in the usual case the optimism is externally imposed as an initial condition [19], here it is spontaneously generated by the noise and self-corrected through exploration.

The approach we propose below to reduce the variance is motivated by electing to represent the uncertainty explicitly, and not indirectly through an optimistic bias. We notice that although *in the end* of the learning process one obtains the deterministic greedy policy from $Q(a,s)$ as in (5), *during* the learning itself the bias in $Q$ can be ameliorated by avoiding the hard min operator, and refraining from committing to a deterministic greedy policy. This can be achieved by adding to $Q$, at the early learning stage, a term that penalizes deterministic policies, which we consider next.

## 3 LEARNING WITH SOFT UPDATES

### 3.1 THE FREE-ENERGY FUNCTION $G$ AND G-LEARNING

Let us adopt, before any interaction with the environment, a simple stochastic prior policy $\rho(a|s)$. For example, we can take the uniform distribution over the possible actions. The *information cost* of a learned policy $\pi(a|s)$ is defined as

$$g^\pi(s,a) = \log \tfrac{\pi(a|s)}{\rho(a|s)}, \tag{8}$$

and its expectation over the policy $\pi$ is the Kullback-Leibler (KL) divergence of $\pi_s = \pi(\cdot|s)$ from $\rho_s = \rho(\cdot|s)$,

$$E_\pi[g^\pi(s,a)|s] = D_{KL}[\pi_s \| \rho_s]. \tag{9}$$

The term (8) penalizes deviations from the prior policy and serves to regularize the optimal policy away from a deterministic action. In the context of the MDP dynamics $p(s_{t+1}|s_t,a_t)$, similarly to (1), we consider the total discounted expected information cost

$$I^\pi(s) = \sum_{t \geq 0} \gamma^t E[g^\pi(s_t,a_t)|s_0=s]. \tag{10}$$

The discounting in (1) and (10) is justified by imagining a horizon $T \sim \text{Geom}(1-\gamma)$, distributed geometrically with parameter $1-\gamma$. Then the cost-to-go $V^\pi$ in (1) and the information-to-go $I^\pi$ in (10) are the total (undiscounted) expected $T$-step costs.

Adding the penalty term (10) to the cost function (1) gives

$$F^\pi(s) = V^\pi(s) + \tfrac{1}{\beta} I^\pi(s), \tag{11}$$
$$= \sum_{t \geq 0} \gamma^t E[\tfrac{1}{\beta} g^\pi(s_t,a_t) + c_t|s_0=s],$$

called the *free-energy function* by analogy with a similar quantity in statistical mechanics [10].

Here $\beta$ is a parameter that sets the relative weight between the two costs. For the moment, we assume that $\beta$ is fixed. In following sections, we let $\beta$ grow as the learning proceeds.

In analogy with the $Q^\pi$ function (2), let us define the *state-action free-energy function* $G^\pi(s,a)$ as

$$G^\pi(s,a) = E_\theta[c|s,a] + \gamma E_p[F^\pi(s')|s,a] \tag{12}$$
$$= \sum_{t \geq 0} \gamma^t E[c_t + \tfrac{\gamma}{\beta} g^\pi(s_{t+1},a_{t+1}))|s_0=s,a_0=a],$$

and note that it does not involve the information term at time $t = 0$, since the action $a_0 = a$ is already known. From the definitions (11) and (12) it follows that

$$F^\pi(s) = \sum_a \pi(a|s) \left[ \tfrac{1}{\beta} \log \tfrac{\pi(a|s)}{\rho(a|s)} + G^\pi(s,a) \right]. \quad (13)$$

It is easy to verify that, given the $G$ function, the above expression for $F^\pi$ has gradient 0 at

$$\pi(a|s) = \frac{\rho(a|s)e^{-\beta G(s,a)}}{\sum_{a'} \rho(a'|s)e^{-\beta G(s,a')}}, \quad (14)$$

which is therefore the optimal policy.

The policy (14) is the soft-min operator applied to $G$, with inverse-temperature $\beta$. When $\beta$ is small, the information cost is dominant, and $\pi$ approaches the prior $\rho$. When $\beta$ is large, we are willing to diverge much from the prior to reduce the external cost, and $\pi$ approaches the deterministic greedy policy for $G$.

Evaluated at the soft-greedy policy (14), the free energy (13) is

$$F^\pi(s) = -\tfrac{1}{\beta} \log \sum_a \rho(a|s) e^{-\beta G^\pi(s,a)}, \quad (15)$$

and plugging this expression into (12), we get that the optimal $G^*$ is a fixed point of the equation

$$G^*(s,a) = \mathrm{E}_\theta[c|s,a] \quad (16)$$

$$- \tfrac{\gamma}{\beta} \mathrm{E}_p \left[ \log \sum_{a'} \rho(a'|s')e^{-\beta G^*(s',a')} \right]$$

$$\equiv \mathbf{B}^*[G^*]_{(s,a)}. \quad (17)$$

Based on the above expression, we introduce G-learning as an off-policy TD-learning algorithm [2], that learns the optimal $G^*$ from the interaction with the environment by applying the update rule

$$G(s_t, a_t) \leftarrow (1 - \alpha_t) G(s_t, a_t) \quad (18)$$

$$+ \alpha_t \left( c_t - \tfrac{\gamma}{\beta} \log \left( \sum_{a'} \rho(a'|s_{t+1}) e^{-\beta G(s_{t+1}, a')} \right) \right).$$

### 3.2 THE ROLE OF THE PRIOR

Clearly the choice of the prior policy $\rho$ is significant in the performance of the algorithm. The prior policy can encode any prior knowledge that we have about the domain, and this can improve the convergence if done correctly. However an incorrect prior policy can hinder learning. We should therefore choose a prior policy that represents all of our prior knowledge, but nothing more. This prior policy has maximal entropy given the prior knowledge [20].

In our examples in Section 6, we use the uniform prior policy, representing no prior knowledge. Both in Q-learning and in G-learning, we could utilize the prior knowledge that moving into a wall is never a good action, by eliminating those actions. One advantage of G-learning is that it can utilize softer prior knowledge. For example, a prior policy that gives lower probability for moving into a wall represent the prior knowledge that such an action is usually (but not always) harmful, a type of knowledge that cannot be utilized in Q-learning.

We have presented G-learning in a fully parameterized formulation, where the function $G$ is stored in a lookup table. Practical applications of Q-learning often resort to approximating the function $Q$ through function approximations, such as linear expansions or neural networks [2, 3, 4, 21, 5]. Such an approximation generates inductive bias, which is another form of implicit prior knowledge. While G-learning is introduced here in its table form, preliminary results indicate that its benefits carry over to function approximations, despite the challenges posed by this extension.

### 3.3 CONVERGENCE

In this section we study the convergence of $G$ under the update rule (18). Recall that the supremum norm is defined as $|x|_\infty = \max_i |x_i|$. We need the following Lemma, proved in the Supplementary Material.

**Lemma 1.** *The operator $\mathbf{B}^*[G]_{(s,a)}$ defined in (17) is a contraction in the supremum norm,*

$$\left| \mathbf{B}^*[G_1] - \mathbf{B}^*[G_2] \right|_\infty \leq \gamma \left| G_1 - G_2 \right|_\infty. \quad (19)$$

The update equation (18) of the algorithm can be written as a stochastic iteration equation

$$G_{t+1}(s_t, a_t) = (1 - \alpha_t) G_t(s_t, a_t) \quad (20)$$
$$+ \alpha_t (\mathbf{B}^*[G_t]_{(s_t, a_t)} + z_t(c_t, s_{t+1}))$$

where the random variable $z_t$ is

$$z_t(c_t, s_{t+1}) = - \mathbf{B}^*[G_t]_{(s_t, a_t)} \quad (21)$$
$$+ c_t - \tfrac{\gamma}{\beta} \log \sum_{a'} \rho(a'|s_{t+1}) e^{-\beta G_t(s_{t+1}, a')}.$$

Note that $z_t$ has expectation 0. Many results exist for iterative equations of the type (20). In particular, given conditions (6) for $\alpha_t$, the contractive nature of $\mathbf{B}^*$, infinite visits to each pair $(s_t, a_t)$ and assuming that $|z_t| < \infty$, $G_t$ is guaranteed to converge to the optimal $G^*$ with probability 1 [17, 22].

## 4 SCHEDULING $\beta$

In the previous section, we showed that running G-learning with a fixed $\beta$ converges, with probability 1, to the optimal $G^*$ for that $\beta$, given by the recursion in (12)–(14).

When $\beta = \infty$, the equations for $G^*$ and $F^*$ degenerate into the equations for $Q^*$ and $V^*$, and G-learning becomes Q-learning. When $\beta = 0$, the update policy $\pi$ in (14) is equal to the prior $\rho$. This case, denoted Q$^\rho$-learning, converges to $Q^\rho$.

In an early stage of learning, Q$^\rho$-learning has an advantage over Q-learning, because it avoids committing to a deterministic policy based on a noisy $Q$ function. In a later stage of learning, when $Q$ is a more precise estimate of $Q^*$, Q-learning gains the advantage by updating with a better policy than the prior. This is demonstrated in section 6.1.

We would therefore like to schedule $\beta$ so that G-learning makes a smooth transition from Q$^\rho$-learning to Q-learning, just at the right pace to enjoy the early advantage of the former and the late advantage of the latter. As we argue below, such a $\beta$ always exists.

### 4.1 ORACLE SCHEDULING

To consider the effect of the $\beta$ scheduling on the correction of the bias (7), suppose that during learning we reach some $G$ that is an unbiased estimate of $G^*$. $G(s_t, a_t)$ would remain unbiased if we update it towards

$$c_t + \gamma G(s_{t+1}, a^*) \qquad (22)$$

with

$$a^* = \arg\min_{a'} G^*(s_{t+1}, a'), \qquad (23)$$

but we do not have access to this optimal action. If we use the update rule (18) with $\beta = 0$, we update $G(s_t, a_t)$ towards

$$c_t + \gamma \sum_{a'} \rho(a'|s_{t+1}) G(s_{t+1}, a'), \qquad (24)$$

which is always at least as large as (22), creating a positive bias. If we use $\beta = \infty$, we update $G(s_t, a_t)$ towards

$$c_t + \gamma \min_{a'} G(s_{t+1}, a'), \qquad (25)$$

which creates a negative bias, as explained in Section 2.2. Since the right-hand side of (18) is continuous and monotonic in $\beta$, there must be some $\beta$ for which this update rule is unbiased.

This is a non-constructive proof for the existence of a $\beta$ schedule that keeps the value estimators unbiased (or at least does not accumulate additional bias). We can imagine a scheduling oracle, and a protocol for the agent by which to consult the oracle and obtain the $\beta$ for its soft updates. At the very least, the oracle must be told the iteration index $t$, but it can also be useful to let $\beta$ depend on any other aspect of the learning process, particularly the current world state $s_t$.

### 4.2 PRACTICAL SCHEDULING

A good schedule should increase $\beta$ as learning proceeds, because as more samples are gathered the variance of $G$ decreases, allowing more deterministic policies. In the examples of Section 6 we adopted the linear schedule

$$\beta_t = kt, \qquad (26)$$

with some constant $k > 0$. Another possibility that we explored was to make $\beta$ inversely proportional to a running average of the Bellman error, which decreases as learning progresses. The results were similar to the linear schedule.

The optimal parameter $k$ can be obtained by performing initial runs with different values of $k$ and picking the value whose learned policy gives empirically the lower cost-to-go. Although this exploration would seem costly compared to other algorithms for which no parameter tuning is needed, these initial runs do not need to be carried for many iterations. Moreover, in many situations the agent is confronted with a class of similar domains, and tuning $k$ in a few initial domains leads to an improved learning for the whole class. This is the case in the domain-generator example in Section 6.1.

## 5 RELATED WORK

The connection between domain noise or function approximation, and the statistical bias in the $Q$ function, was first discussed in [18, 3]. An interesting modification of Q-learning to address this problem is Double-Q-learning [11, 14], which uses two estimators for the $Q$ function to alleviate the bias. Other modifications of Q-learning that attempt to reduce or correct the bias are suggested in [12, 13].

An early approach to Q-learning in continuous noisy domains was to learn, instead of the value function, the advantage function $A(s,a) = Q(s,a) - V(s)$ [23]. The algorithm represents $A$ and $V$ separately, and the optimal action is determined from $A(s,a)$ as $a^*(s) = \arg\min_a A(s,a)$. In noisy environments, learning $A$ is shown in some examples to be faster than learning $Q$ [23, 24].

More recently, it was shown that the advantage learning algorithm is a gap-increasing operator [25]. As discussed in Section 2.2, the action gap is a central factor in the generation of bias, and increasing the gap should also help reduce the bias. In Section 6.1 we compare our algorithm to the consistent Bellman operator $\mathcal{T}_C$, one of the gap-increasing algorithms introduced in [25].

For other works that study the effect of noise in Q-learning, although without identifying the bias (7), see [26, 27, 28].

Information considerations have received attention in recent years in various machine learning settings, with the free energy $F^\pi$ and similar quantities used as a design principle for policies in known MDPs [10, 29, 30]. Other works have used related methods for reinforcement learning [31, 32, 33, 34, 35]. A KL penalty similar to ours is used in [35], in settings with known reward and transition functions, to encourage "curiosity".

Soft-greedy policies have been used before for exploration [2, 36], but to our knowledge G-learning is the first TD-learning algorithm to explicitly use soft-greedy policies in its updates.

Particularly relevant to our work is the approach studied in [32]. There the policy is iteratively improved by optimizing it in each iteration under the constraint that it only diverges slightly, in terms of KL-divergence, from the empirical distribution generated by the previous policy. In contrast, in G-learning we measure the KL-divergence from a fixed prior policy, and in each iteration allow the divergence to grow larger by increasing $\beta$. Thus the two methods follow different information-geodesics from the stochastic prior policy to more and more deterministic policies.

This distinction is best demonstrated by considering the $\Psi$-learning algorithm presented in [33, 34], based on the same approach as [32]. It employs the update rule

$$\Psi(s_t, a_t) \leftarrow \Psi(s_t, a_t) \qquad (27)$$
$$+ \alpha_t(c_t + \gamma \bar{\Psi}(s_{t+1}) - \bar{\Psi}(s_t)),$$

with

$$\bar{\Psi}(s) = -\log \sum_a \rho(a|s) e^{-\Psi(s,a)}, \qquad (28)$$

which is closely related to our update of $G$ in (18).

Apart from lacking a $\beta$ parameter, the most important difference is that the update of $\Psi$ involves subtracting $\alpha_t \bar{\Psi}(s_t)$, whereas the update of $G$ involves subtracting $\alpha_t G(s_t, a_t)$. This seemingly minor modification has a large impact on the behavior of the two algorithms. The update of $G$ is designed to pull it towards the optimal state-action free energy $G^*$, for all state-action pairs. In contrast, subtracting the log-partition $\bar{\Psi}(s_t)$, in the long run pulls only $\Psi(s_t, a^*)$, with $a^*$ the optimal action, towards its true value, while for the other actions the values grow to infinity. In this sense, the $\Psi$-learning update (27) is an information-theoretic gap-increasing Bellman operator [25].

The growth to infinity of suboptimal values separates them from the optimal value, and drives the algorithm to convergence. In G-learning, this parallels the increase in $\beta$ with the accumulation of samples. However, there is a major benefit to keeping $G$ reliable in all its parameters, and controlling it with a separate $\beta$ parameter. In $\Psi$-learning, the $\Psi$ function penalizes actions it deems suboptimal. If early noise causes an error in this penalty, the algorithm needs to unlearn it - a similar drawback to that of Q-learning. In Section 6, we demonstrate the improvement offered by G-learning.

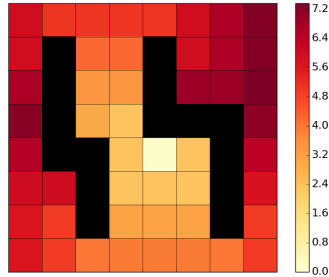

Figure 1: Gridworld domain. The agent can choose an adjacent square as the target to move to, and then may end up stochastically in a square adjacent to that target. The color scale indicates the optimal values $V^*$ with a fixed cost of 1 per step.

## 6 EXAMPLES

This section illustrates how G-learning improves on existing model-free learning algorithms in several settings. The domains we use are clean and simple, to demonstrate that the advantages of G-learning are inherent to the algorithm itself.

We schedule the learning rate $\alpha_t$ as

$$\alpha_t = n_t(s_t, a_t)^{-\omega}, \qquad (29)$$

where $n_t(s_t, a_t)$ is the number of times the pair $(s_t, a_t)$ was visited. This scheme is widely used, and is consistent with (6) for $\omega \in (1/2, 1]$. We choose $\omega = 0.8$, which is within the range suggested in [37].

We schedule $\beta$ linearly, as discussed in Section 4.2. In each case, we start with 5 preliminary runs of G-learning with various linear coefficients, and pick the coefficient with the lowest empirical cost. This coefficient is used in the subsequent test runs, whose results are plotted in Figure 2.

In all cases, we use a uniform prior policy $\rho$, a discount factor $\gamma = 0.95$, and 0 for the initial values ($Q_0 = 0$ in Q-learning, and similarly in the other algorithms). Except when mentioned otherwise, we employ random exploration, where $s_t$ and $a_t$ are chosen uniformly at the beginning of each time step, independently of any previous sample. This exploration technique is useful when comparing update rules, while controlling for the exploration process.

### 6.1 GRIDWORLD

Our first set of examples occurs in a gridworld of $8 \times 8$ squares, with some unavailable squares occupied by walls shown in black (Figure 1). The lightest square is the goal, and reaching it ends the episode.

At each time step, the agent can choose to move one square in any of the 8 directions (including diagonally), or stay in place. If the move is blocked by a wall or the edge of the board, it effectively attempts to stay in place. With some probability, the action performed by the agent is further followed by an additional random slide: with probability 0.15 to each vertically or horizontally adjacent available position, and with probability 0.05 to each diagonally adjacent available position.

The noise associated with these random transitions can be enhanced further by the possible variability in the costs incurred along the way. We consider three cases. In the first case, the cost in each step is fixed at 1. In the second case, the cost in each step is distributed normally i.i.d, with mean 1 and standard deviation 2. In the third case we define a distribution over domains, such that at the time of domain-generation the mean cost for each state-action is distributed uniformly i.i.d over $[1, 3]$. Once the domain has been generated and interaction begins, the cost itself in each step is again distributed normally i.i.d, with the generated mean and standard deviation 4.

We attempt to learn these domains using various algorithms. Figure 2 summarizes the results for Q-learning, G-learning, Double-Q-learning [11], $\Psi$-learning [33, 34] and the consistent Bellman operator $\mathcal{T}_C$ of [25]. We also include $Q^\rho$-learning, which performs updates as in (4) towards the prior policy $\rho$. Comparison with Speedy-Q-learning [12] is omitted, since it showed no improvement over vanilla Q-learning in these settings. In our experiments, these algorithms had comparable running times.

The $\beta$ scheduling used in G-learning is linear, with the coefficient $k$ equal to $10^{-3}$, $10^{-4}$, $5 \cdot 10^{-5}$ and $10^{-6}$, respectively for the fixed-cost, noisy-cost, domain-generator and cliff domains (see Section 6.2).

For each case, Figure 2 shows the evolution over 250,000 algorithm iterations of the following three measures, averaged over $N = 100$ runs:

1. Empirical bias, defined as

$$\frac{1}{Nn} \sum_{i=1}^{N} \sum_{s=1}^{n} (V_{i,t}(s) - V_i^*(s)), \quad (30)$$

   where $i$ indexes the $N$ runs and $s$ the $n$ states. Here $V_{i,t}$ is the greedy value based on the estimate obtained by each algorithm ($Q$, $G$, etc.), in iteration $t$ of run $i$. The optimal value $V_i^*$, computed via Value Iteration, varies between runs in the domain-generator case.

2. Mean absolute error in $V$

$$\frac{1}{Nn} \sum_{i=1}^{N} \sum_{s=1}^{n} |V_{i,t}(s) - V_i^*(s)|. \quad (31)$$

   A low bias could result from the cancellation of terms with high positive and negative biases. A convergence in the absolute error is more indicative of the actual convergence of the value estimates.

3. Increase in cost-to-go, relative to the optimal policy

$$\frac{1}{Nn} \sum_{i=1}^{N} \sum_{s=1}^{n} (V^{\pi_{i,t}}(s) - V_i^*(s)). \quad (32)$$

   This measures the quality of the learned policy. Here $\pi_{i,t}$ is the greedy policy based on the state-action value estimates, and $V^{\pi_{i,t}}$ is its value in the model, computed via Value Iteration.

An algorithm is better when these measures reach zero faster. As is clear in Figure 2, in the domains with noisy cost (Rows 2 and 3), G-learning dominates over all the other competing algorithms by the three measures. The results are statistically significant, but plotting confidence intervals would clutter the figure.

An important and surprising point of Figure 2 is that $Q^\rho$-learning always outperforms Q-learning initially, before degrading. The reason is that the Q-learning updates initially rely on very few samples, so these harmful updates need to be undone by later updates. $Q^\rho$-learning, on the other hand, updates in the direction of a uniform prior. This gives an early advantage in mapping out the local topology of the problem, before long-range effects start pulling the learning towards the suboptimal $Q^\rho$.

The power of G-learning is that it enjoys the early advantage of $Q^\rho$-learning, and smoothly transitions to the convergence advantage of Q-learning. When $\beta$ is small, the information cost $g_t$ (8) outweighs the external costs $c_t$, and we update towards $\rho$. As samples keep coming in, and our estimates improve, $\beta$ increases, and the updates gradually lean more towards a cost-optimizing policy. Unlike early stages in Q-learning, at this point $G_t$ is already a good estimate, and we avoid overfitting. As mentioned above, Figure 2 shows that this effect is more manifest in noisier scenarios.

Finally, Figure 3 shows running averages of the Bellman error for the different algorithms considered. The Bellman error in G-learning is the coefficient multiplying $\alpha_t$ in (18),

$$\Delta G_t \equiv c_t - \tfrac{\gamma}{\beta} \log \left( \sum_{a'} \rho(a'|s_{t+1}) e^{-\beta G_t(s_{t+1}, a')} \right)$$
$$- G_t(s_t, a). \quad (33)$$

When learning ends and $G = G^*$, the expectation of $\Delta G_t$ is zero (see (16)). Similar definitions hold for the other learning algorithms we compare with. As is clear from Figure 3, G-learning reaches zero average Bellman error faster than the competing methods, even while $\beta$ is still increasing in order to make $G^*$ converge to $Q^*$.

## 6.2 CLIFF WALKING

Cliff walking is a standard example in reinforcement learning [2], that demonstrates an advantage of on-policy algorithms such as SARSA [2, 4] and Expected-SARSA [15, 16] over off-policy learning approaches such as Q-learning. We use it to show another interesting strength of G-learning.

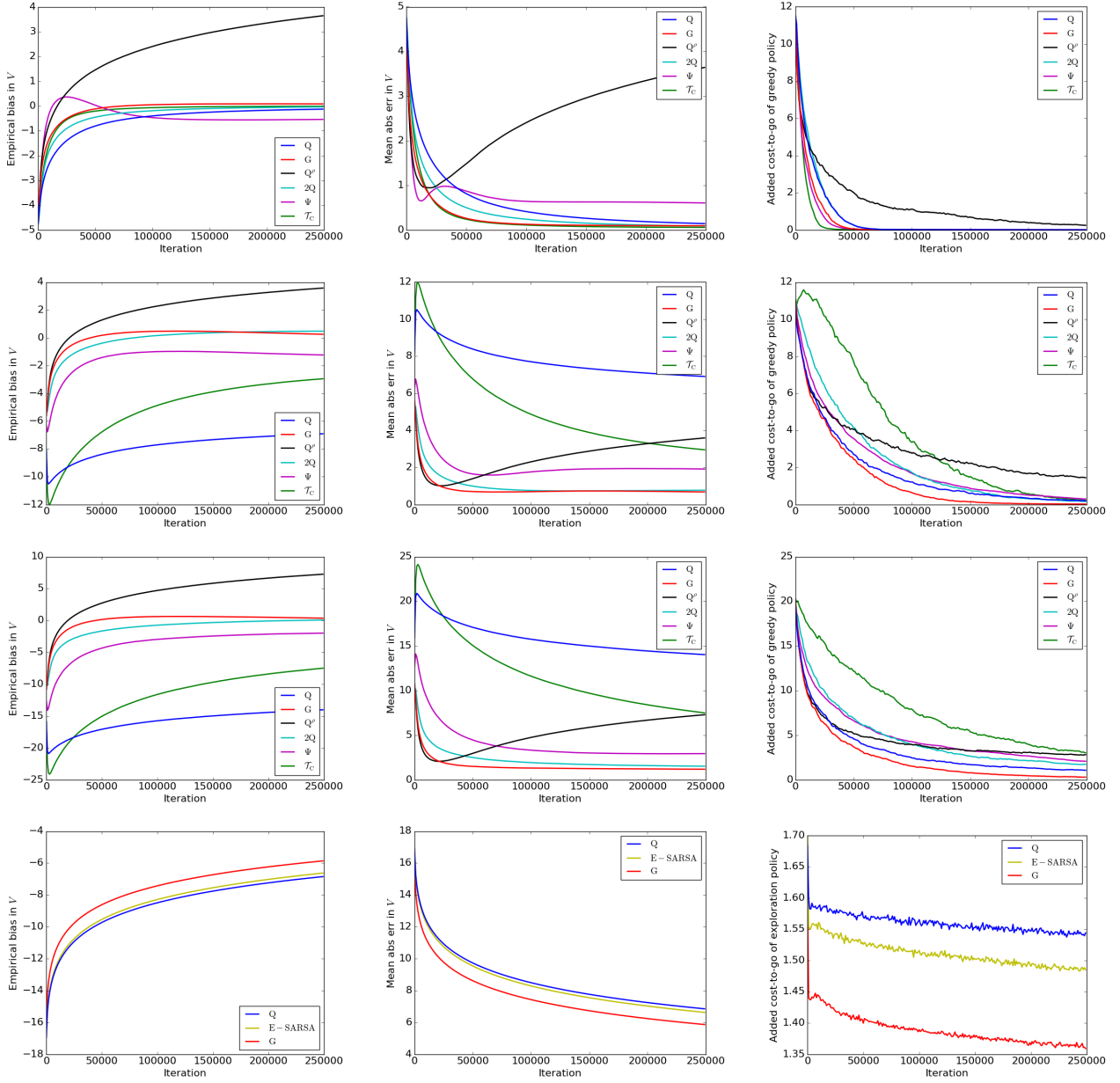

Figure 2: **Gridworld (Rows 1-3):** Comparison of Q-, G-, $Q^\rho$-, Double-Q-, $\Psi$- and $\mathcal{T}_C$-learning. **Row 1:** The cost in each step is fixed at 1. **Row 2:** The cost in each step is distributed as $\mathcal{N}(1, 2^2)$. **Row 3:** In each run, the domain is generated by drawing each $\mathrm{E}[c|s, a]$ uniformly over $[1, 3]$. The cost in each step is distributed as $\mathcal{N}(\mathrm{E}[c|s, a], 4^2)$. Note that in the noisy domains (Rows 2 and 3), G-learning dominates over all the other algorithms by the three measures. **Cliff (Row 4):** Comparison of Q- and G-learning, and Expected-SARSA. The cost in each step is 1, and falling off the cliff costs 5. **Left:** Empirical bias of $V$, relative to $V^*$ (30). **Middle:** Mean absolute error between $V$ and $V^*$ (31). **Right:** Value of greedy policy, with the baseline $V^*$ subtracted (32); except in Row 4, which shows the value of the exploration policy.

In this example, the agent can walk on the grid in Figure 4 horizontally or vertically, with deterministic transitions. Each step costs 1, except when the agent walks off the cliff (the bottom row), which costs 5, or reaches the goal (lower right corner), which costs 0. In either of these cases, the position resets to the lower left corner.

Exploration is now on-line, with $s_t$ taken from the end of the previous step. The exploration policy in our simulations is $\epsilon$-greedy with $\epsilon = 0.1$, i.e. with probability $\epsilon$ the agent chooses a random action, and otherwise it takes deterministically the one that seems optimal. In practice, $\epsilon$ can be decreased after the learning phase, however it is also common to keep $\epsilon$ fixed for continued exploration [2].

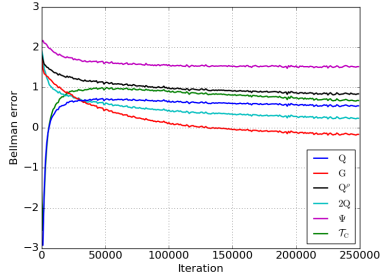

Figure 3: Running average of the Bellman error in the gridworld domain-generator example for Q-, G-, $Q^\rho$-, Double-Q-, $\Psi$- and $\mathcal{T}_C$-learning. The results for the other two gridworlds of Figure 2 are similar.

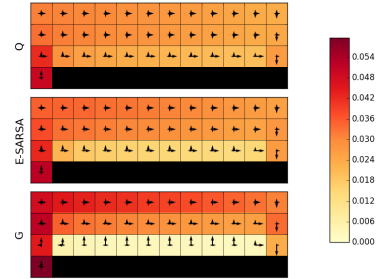

Figure 4: Cliff domain. The agent can choose a horizontally or vertically adjacent square, and moves there deterministically. The color scale and the arrow lengths indicate, respectively, the frequency of visiting each state and of making each transition, in the first 250,000 iterations of Q-learning, Expected-SARSA and G-learning. The near-greedy exploration policy of Q-learning has higher chance of taking the shortest path near the edge of the cliff at the bottom, than that of G-learning. As an off-policy algorithm, Q-learning fails to optimize for the exploration policy, whereas G-learning succeeds.

In this setting, as shown in the bottom row of Figure 2, an off-policy algorithm like Q-learning performs poorly in terms of the value of its exploration policy, and the empirical cost it incurs. It learns a rough estimate of $Q^*$ quickly, and then tends to use it and walk on the edge of the cliff. This leads to the agent occasionally exploring the possibility of falling off the cliff. In contrast, an on-policy algorithm like Expected-SARSA [15, 16] learns the value of its exploration policy, and quickly manages to avoid the cliff.

Figure 4 compares Q-learning, G-learning and Expected-SARSA in this domain, and shows that G-learning learns to avoid the cliff even better than an on-policy algorithm, although for a different reason. As an off-policy algorithm, G-learning does learn the value of the update policy, which prefers trajectories far from the cliff in the early stages of learning. This occurs because near the cliff, avoiding the cost of falling requires ruling out downward moves, which has a high information cost. On the other hand, trajectories far from the cliff, while paying a higher cost in overall distance to the goal, enjoy lower information cost because acting randomly is not costly for them.

As shown in the bottom row of Figure 2, by using a greedy policy for $G$ as the basis of the $\epsilon$-greedy exploration, we enjoy the benefits of being aware of the value of the exploration policy during the learning stage. At the same time, G-learning converges faster than either Q-learning or Expected-SARSA to the correct value function. In this case the "noise" that G-learning mitigates is related to the variability associated with the exploration.

## 7 CONCLUSIONS

The algorithm we have introduced successfully mitigates the slow learning problem of early stage Q-learning in noisy environments, that is caused by the bias generated by the hard optimization of the policy.

Although we have focused on Q-learning as a baseline, we believe that early-stage information penalties can also be applied to advantage in more sophisticated model-free settings, such as TD($\lambda$), and combined with other incremental learning techniques, such as function approximation, experience replay and actor-critic methods.

G-learning takes a Frequentist approach to estimating the optimal $Q$ function. This is in contrast to Bayesian Q-learning [38], which explicitly models the uncertainty about the $Q$ function as a posterior distribution. It would be interesting to study the bias that hard optimization causes in the mean of this posterior, and to consider its reduction using methods similar to G-learning.

An important next step is to apply G-learning to more challenging domains, where an approximation of the $G$ function is necessary. The simplicity of our linear $\beta$ schedule (26) should facilitate such extensions, and allow G-learning to be combined with other schemes and algorithms. Further study should also address the optimal schedule for $\beta$. We leave these important questions for future work.

### Acknowledgments

AP is supported by ONR grant N00014-14-1-0243 and IARPA via DoI/IBC contract number D16PC00003. RF and NT are supported by the DARPA MSEE Program, the Gatsby Charitable Foundation, the Israel Science Foundation and the Intel ICRI-CI Institute.

# Chapter 5

# Discussion

In this thesis we studied bounded agents that operate in dynamical systems under intrinsic informational constraints. This setting can be modeled as a sequential rate-distortion problem, and solved with a forward-backward algorithm. We investigated the convergence properties of the algorithm, exploited the structure of the special LQG case, and simplified the setting to be usable for learning. In this section we summarize and discuss some of the insights gained in our work.

**A Principle for the Tradeoff of Informational Resources and Costs**

There are various ways to model bounded agents with limited information-processing resources. Our approach, introduced in Section 2.1, is to identify distinct components within the agent, such as sensors, memory and actuators, and consider the information rates on the communication channels between these components.

The reduction of extrinsic costs is usually taken as the optimization target, with extrinsic dynamical constraints, to which we add intrinsic constraints on the rates at which information can be communicated between sensors, memory and actuators. In the Lagrangian form of this optimization problem, the latter become intrinsic informational costs, which are traded off with extrinsic expectational costs.

As an alternative formulation of the optimization problem, we can set an upper constraint on the extrinsic cost, and seek the simplest agent that achieves this cost level. With a fixed prior behavior, simplicity of a policy can be measured as the Kullback-Leibler (KL) divergence of the solution policy from the prior. On an evolutionary timescale, the prior itself is adaptive, and the optimization target becomes the information rate.

Interpreting informational constraints as costs is insightful, in that it allows trading off the informational costs of various channels among themselves. For example, when memory resources are scarce, it may be easier to extract some information from observations again and again, rather than remember it. On the other hand, when memory has a high enough capacity, it can help process the sensory input by generating a good prediction of the observation, and only attending to some surprises.

To illustrate this point further, consider the optimal policy for a fully observable MDP. Without informational constraints, there is always an optimal policy which is reactive (memoryless), and simply depends on the current state [2]. However, if attending the state spends precious informational-processing resources, and thus incurs informational costs, the setting becomes *partially attendable*, even though it remains *fully observable*. If we allow a memory channel from past internal agent states to future ones that is cheaper than the sensory channel, it may be beneficial to make up for unattended sensory input using remembered information.

**Periodicity and Instability of the Optimization Principle**

The algorithm presented in Section 2.1 is applicable to general POMDPs. However, it is only demonstrated there on passive POMDPs, where actions incur costs but do not affect the state of the world. Experiments with other types of examples exhibited poor convergence that seemed to be the result of periodicity or instability of the solution under the update operator, particularly after phase transitions that increase the support of the agent policy.

We have thus taken the first steps in the study of the bifurcation structure of the learning dynamics around critical values of the tradeoff param-eter $\beta$. Section 2.2 analyzes examples that illustrate this structure. The phenomenology of planning in partial observability under informational constraints includes period doubling through supercritical pitchfork bifurcations. The optimal solution at these critical points becomes periodic, requiring the agent to start paying attention to a clock signal. The optimal stationary (aperiodic) solution remains a fixed point, but loses its stability to perturbations.

The conclusion is that for reinforcement learning algorithms to converge in partially observable domains (or, indeed, under partial attendability or approximate inference), they must allow for periodicity of the solution policy. This holds true for value iteration and gradient methods alike. We also note that the periodicity itself is a channel from the clock to the controller, and may be subject to information constraints.

**The Linear-Quadratic-Gaussian Case**

The general algorithm presented in Section 2.1 is polynomial in the sizes of the world state, memory state, observation and action spaces involved. When these spaces are very large or continuous, we can no longer apply the algorithm in this tabular form. Instead, the solution must be parameterized in a tractable manner, and the gradient must be taken with respect to these parameters.

A particularly important and insightful parametric family, studied in Chapter 3, is the Gaussian distributions (for $\bar{p}$), the linear-Gaussian conditional distributions (for $p$, $\sigma$, $q$ and $\pi$) and the quadratic functions (for $c$ and $\nu$). This family has special properties when considering unbounded agents. It is self-conjugate, meaning that under linear-Gaussian dynamics, a Gaussian marginal remains Gaussian, and a quadratic cost-to-go function remains quadratic when the cost rate is quadratic. There is also separation of the forward inference process and the backward control process.

It comes as no surprise that this case is also special when considering bounded agents. Although the cost-to-go is not quadratic when considering informational costs, the solution method only involves its second-order expansion. The forward and backward processes are coupled, but many local optima of the type that plagues the discrete case are avoided by the tools available to treat second-order systems.

In particular, the sequential rate-distortion problem can be formulated in the LQG case as a sequential semidefinite program. Its solution provides not only first-order necessary conditions for a solution to be optimal, i.e. having gradient 0, but also higher-order necessary conditions. This prevents some local optima where the inference and control policies are optimal given each other, but jointly suboptimal as a pair.

**Learning and $\beta$ Scheduling**

Learning is the process of gaining useful information about the world through interaction. An agent in interaction with an environment whose state is partially observable has to perform learning, whether or not it has a model of the dynamics. The setting where no such model is available is particularly interesting, since it illustrates how the tradeoff between cost and simplicity changes as the algorithm progresses, as shown in Section 4.1.

The maximum relative entropy principle states that a solution should minimize the KL divergence to a simple prior, under the constraint that it fits any additional information we have about the solution. In the MDP learning setting, this additional information is represented by the value function, which is iteratively improved by sample-based updates. An imperfect value function cannot generally be used to select an optimal policy, and we must settle for a suboptimal value guarantee. The policy used in each update should thus be the simplest one, in terms of KL divergence, under the constraint that this value guarantee level is achieved. As learning progresses, the value function becomes more accurate, the guarantee can be improved, and hence the tradeoff coefficient $\beta$ is increased to reflect a larger emphasis on the extrinsic cost.

The principle that $\beta$ should generally increase as the learned parameters improve is not unique to sample-based methods. The goal of any computation is to reduce the uncertainty about its output, and iterative algorithms generally reduce this uncertainty gradually. If the partially optimized solution is used to obtain an improved solution, it may be beneficial to consider soft-optimization, by taking the simplest solution under the constraint that a gradually increasing guarantee level is achieved. For example, $\beta$ scheduling can also be used in this manner to improve convergence in value iteration, and many other algorithms.

# Glossary

**Action, Control signal** Input to the POMDP state transition. Output of the agent control policy.
**Notation**: $a_t$.

**Backward process** Computation of the cost-to-go function through the application of the Bellman operator backward in time.

**Bayesian network** Graphical model of a distribution as a directed acyclic graph with a variable in each node. The joint distribution of all variable is given by the product of the distributions of each variable given its parents.
**Notation**: $p(x_1, \ldots, x_n) = \prod_i p(x_i|\text{parents}(x_i))$.

**Belief** Probability distribution over world states that is represented in the agent memory state.
**Notation**: objective: $\mathbb{P}(w|m)$; subjective: $b_m(w)$.

**Capacity-cost problem** Optimization of the tradeoff between the capacity $\mathbb{I}[x;y]$ for information on the channel and the expected cost $\mathbb{E}[c(x)]$.
**Notation**: inputs: $p_{Y|X}(y|x)$, $c(x)$; output: $q(x)$.

**Channel** Stochastic mapping of the channel input $x$ to the channel output $y$. A cost $c(x)$ on the channel input is sometimes also considered part of the definition of the channel.
**Notation**: $p_{Y|X}(y|x)$.

**Channel coding** Encoding of an input signal $s$ into the channel input $x$ and decoding of the channel output $y$ as a reconstruction signal $\hat{s}$.
**Notation**: encoder: $x = g(s)$; decoder: $\hat{s} = h(y)$.

**Control policy** Probability distribution of the agent's action given its memory state.
**Notation**: $\pi(a_t|m_t)$.

**Cost** Real function of the system's state, usually the world state and the action, whose expectation is used as the minimization target.
**Notation**: $c(w_t, a_t)$.

**Exploitation** Agent behavior aimed at achieving good value, based on known aspects of the world.

**Exploration** Agent behavior aimed at learning unknown aspects of the world.

**Finite horizon** Minimization target that considers the total cost of the process. Meaningful when the process has a finite expected termination time, such as in the episodic, discounted and fixed-horizon settings.
**Notation**: $\sum_{t=0}^{\infty} \mathbb{E}[c(w_t, a_t)]$.

**Forward process** Computation of the marginal state distribution through the application of the dynamics operator forward in time.

**Full controllability** Complete determination of the next state by the input action.
**Notation**: $p(w_{t+1}|w_t, a_t) = \delta_{w_{t+1}=a_t}$.

**Full observability** Complete revelation of the state as an observation.
**Notation**: $\sigma(o_t|w_t) = \delta_{o_t=w_t}$.

**History-based policy** Most general form of an agent policy, where its output action depends arbitrarily on its past inputs, the observable history.
**Notation**: $\pi(a_t|o_{\leqslant t})$.

**Inference policy** Probability distribution of the agent's next memory state given its current memory state and the new observation.
**Notation**: $q(m_{t+1}|m_t, o_{t+1})$.

**Infinite horizon** Minimization target that considers the long-term average cost of the process. Meaningful when the process is stationary.
**Notation**: $\limsup_{T\to\infty} \frac{1}{T} \sum_{t=0}^{T-1} \mathbb{E}[c(w_t, a_t)]$.

**Marginal distribution** Distribution induced by a stochastic process on a subset of its random variables, often a state.
**Notation**: $\bar{p}(s_t)$, $\bar{\pi}(a_t)$.

**Model-based learning** Learning based on update equations that involve a model of the world dynamics.

**Objectively consistent inference** Inference policy that induces subjective beliefs which are consistent with the objective beliefs. **Notation**: $b_m(w) = \mathbb{P}(w|m)$.

**Observation** Output emitted by the POMDP depending on its state. Input to the agent inference policy.
**Notation**: $o_t$.

**Observation dynamics** Probability distribution of the observation given the state.
**Notation**: $\sigma(o_t|s_t)$.

**Partial attendability** Intrinsic limitation on the agent's ability to attend to its inputs when performing inference. For example, the constraint of a low mutual information $\mathbb{I}[m_t, o_{t+1}; m_{t+1}]$ between the inputs and the outputs of the inference step.

**Partial controllability** Extrinsic limitation on the agent's ability to determine the world state transition. In particular, the constraint that $\mathbb{P}(w_{t+1}|w_t, m_t)$ belongs to the convex hull of $\{p(w_{t+1}|w_t, a_t) : a_t \in \mathcal{A}\}$.

**Partial intendability** Intrinsic limitation on the agent's ability to intend its outputs when performing control. For example, the constraint of a low mutual information $\mathbb{I}[m_t; a_t]$ between the inputs and the outputs of the control step.

**Partial observability** Extrinsic limitation on the agent's ability to observe the world state. In particular, the constraint that $\mathbb{P}(m_{t+1}|m_t, w_{t+1})$ is induced by the mixture $\sigma(o_t|w_t)$, applied to an inference policy.

**Rate-distortion problem** Optimization of the tradeoff between the rate $\mathbb{I}[s; \hat{s}]$ of information that the reconstruction has on the source and the expected distortion $\mathbb{E}[d(s, \hat{s})]$.
**Notation**: inputs: $p_S(s)$, $d(s, \hat{s})$; output: $q(\hat{s}|s)$.

**Reactive, memoryless agent** Agent without an internal memory state, consisting of a memoryless policy.

**Reactive, memoryless policy** Probability distribution of the agent's action given its most recent observation.
**Notation**: $\pi(a_t|o_t)$.

**Retentive agent** Memory-utilizing agent, consisting of an inference policy and a control policy.

**Sample-based learning** Learning based on update equations that utilize samples of the world dynamics.

**Source** Probability distribution of a signal $s$. A distortion $d(s, \hat{s})$ between the signal and its reconstruction is sometimes also considered part of the definition of the source.
**Notation**: $p_S(s)$.

**Source coding** Stochastic encoding of a source signal into an intermediate representation and decoding its reconstruction.
**Notation**: encoder: $g(z|s)$; decoder: $\hat{s} = h(z)$.

**State** Time-dependent property of a system that separates the past and the future of the system.

**Notation**: closed system, joint world-agent system: $s_t$; world: $w_t$; agent memory: $m_t$.

**State dynamics, transition** Probability distribution of the next state given the current state and any inputs.

**Notation**: closed system: $p(s_{t+1}|s_t)$; open system: $p(w_{t+1}|w_t, a_t)$.

**Stationary distribution** Marginal distribution of the state that remains the same after a step of the dynamics.